\documentclass[a4paper,fleqn]{cas-dc}

\usepackage{booktabs}  
\usepackage{array}     
\usepackage{multirow} 
\usepackage{subcaption}
\usepackage{amsmath,amsfonts}
\usepackage{array}
\usepackage{textcomp}
\usepackage{stfloats}
\usepackage{url}
\usepackage{verbatim}
\usepackage{graphicx}
\usepackage{graphicx}%
\usepackage{multirow}%
\usepackage{amsmath,amssymb,amsfonts}%
\usepackage{amsthm}%
\usepackage{mathrsfs}%
\usepackage{xcolor}%
\usepackage{textcomp}%
\usepackage{manyfoot}%
\usepackage{booktabs}%
\usepackage{algpseudocode}%
\usepackage{listings}%
\usepackage{array}
\usepackage{rotating}
\usepackage{lineno}
\usepackage{booktabs}
\usepackage{makecell}
 
\usepackage[authoryear,longnamesfirst]{natbib}

\def\tsc#1{\csdef{#1}{\textsc{\lowercase{#1}}\xspace}}
\tsc{WGM}
\tsc{QE}

\begin{document}
\let\WriteBookmarks\relax
\def\floatpagepagefraction{1}
\def\textpagefraction{.001}

\shorttitle{}    

\shortauthors{}  

\title [mode = title]{Combining Hierarchical Cognitive Process with Process Supervision for Interpretable Scene Safety Understanding}  

\author{Zhiyun Jiang, Hanyong Wang, Binbin Liang, Yu Xie, Zhengjie Wang, Menglong Yang, Wei Li}

\begin{abstract}
Scene safety understanding plays a life-or-death role in situational awareness in various critical domains. Traditional methods that rely on learning direct mappings between scenes and safety levels often lack interpretability, limiting their reliability in critical applications. An effective approach to overcoming this challenge lies in interpreting human cognitive processes and equipping machine models with analogous cognitive capabilities. This work explores an effective way of integrating scene safety cognitive process modeling and process supervision. Specifically, we first construct a hierarchical cognitive safety structure, which motivates the development of a novel, high-quality scene safety understanding dataset based on multi-step reasoning with process labels. This dataset serves both as a benchmark and a resource to improve the safety reasoning capabilities of Large Language Models (LLMs), while also enabling a granular analysis of intermediate reasoning steps through information flow and saliency-based techniques. Building upon this foundation, we introduce a modular and flexible process supervision framework that reflects the hierarchical nature of human cognition. This framework leverages LLMs as the core architecture and incorporates Low-Rank 
Adaptation(LoRA) and Mixture-of-Experts (MoE) strategies to enable specialization and collaboration among expert modules, each tasked with specific sub-processes of the overall reasoning chain. Systematic experimental evaluations and analyses confirm that our framework exhibits superior interpretability and performance characteristics compared to traditional approaches.
\end{abstract}





\maketitle

\section{Introduction}
Safety constitutes a fundamental cognitive dimension in scene understanding frameworks, serving as a critical indicator of situational awareness in dynamic environments\cite{cognitive_goal}. The significance of scene safety understanding is paramount across diverse applications, including safety assessments on construction sites\cite{construction1} and public transportation scenarios\cite{traffic1}. The rapid advancement in machine intelligence technologies has led to a growing interest in the automatic understanding of scene safety. The primary objective of automatic scene safety understanding is to leverage machine learning techniques to extract features of interest from a given scene, analyze the relationships and behaviors of these features using computer vision and natural language processing methods, and ultimately infer the safety status of the scene.

Current machine learning models predominantly employ an outcome-supervised training paradigm, wherein the models are trained to reconstruct a holistic scene safety assessment\cite{risk_assessment_tradition_1}. However, scenes invariably encompass a variety of entities whose appearances and interactions exhibit significant stochasticity. The intricate relationships among these scene components extend beyond simple compositionality\cite{visual_composition}. For machine learning models, the inherent complexity and variability of real-world scenes pose substantial challenges in information extraction and integration, relational reasoning, and logical induction and deduction. Furthermore, the heavy reliance of machine learning technologies on statistical learning as their core framework\cite{statistical_learning} frequently leads to issues such as latency and lack of interpretability in cognition, assessment, and decision-making, thereby complicating the assurance of the reliability of machine intelligence.

To achieve accurate scene safety understanding, it is essential that machine intelligence incorporates human cognitive capabilities. Humans possess extensive visual experience, common sense knowledge, and semantic understanding\cite{human_visual_experience}, which enable them to perform in-depth analysis and relational reasoning about complex scenes. Reliable and effective human-machine collaborative cognition requires the seamless integration of human cognitive information into machine intelligence systems, coupled with the provision of timely feedback from the system\cite{brain_to_machine1}. However, effectively articulating human cognitive processes and designing interaction and reasoning mechanisms that are both observable and controllable within the context of human-machine collaborative scene safety understanding remain critical challenges. The global workspace theory suggests that the brain solves complex problems through the collaboration of multiple submodules that perform specific tasks\cite{gwt1}. If machine learning systems can be decomposed into modular subcomponents that map to different analytical levels within the human cognitive hierarchy, this facilitates targeted injection of human knowledge into relevant stages of the model.

 In this study, we construct a step-wise modular machine cognition model, with each component focusing on different aspects of scene safety understanding analogous to stages in the human cognitive process. We introduce a cognitive hierarchy model that mimics the human cognitive process as a supervisory signal to perform process supervision on the machine cognition model. Unlike conventional outcome supervision approaches, process supervision models receive feedback from each stage of the cognitive hierarchy, allowing for targeted debugging and adjustment of relevant modules in the machine cognition model. Specifically, the machine cognition model can begin with the impact of lower-level object/instance cognition on scene safety understanding from the cognitive hierarchy, progressively acquiring relationship mining capabilities from the cognitive hierarchy, and ultimately achieving cognitive hierarchy-guided overall scene safety understanding. In other words, each step or stage of the machine cognition model can receive feedback from the cognitive hierarchy. Thus, under the guidance of the cognitive hierarchy model, scene safety understanding by the machine cognition model is accomplished progressively. The step-wise injection of human knowledge through the cognitive hierarchy enables interpretable debugging capabilities at each stage of the machine cognition model and promotes model evolution in accordance with human cognitive processes.

The motivation for this study lies in the design of scene safety understanding capabilities by emulating the hierarchical and sequential nature of human scene perception and safety assessment. This approach seeks to develop scene safety interpretation abilities via process-supervised learning, inspired by and modeled after human cognitive mechanisms. The specific contributions of this research can be summarized as follows:

\begin{enumerate}
    \item We propose a hierarchical cognitive structure for scene safety and further create a scene safety understanding dataset based on multi-step reasoning with process labels (SSUPL), which reflects this cognitive mechanism.
    \item We demonstrate that SSUPL-based scene safety understanding in LLMs exhibits human-like hierarchical cognitive mechanisms from an information flow perspective, and further analyze the key factors affecting the final safety prediction in LLMs.
    \item We propose a process-supervised training framework that reflects the cognitive hierarchy by combining Mixture-of-Experts (MoE) and Low-Rank Adaptation (LoRA). We conduct a large number of experiments based on this framework on different LLMs, and the results show that our approach outperforms the traditional paradigm.
\end{enumerate}

\section{Related Works}
\subsection{Scene Safety Understanding}
Traditional machine learning-based approaches to scene safety typically begin by having an expert define a collection of categories for some common pitfalls, hazards, or anomalies in a vertical domain. Then, these predefined categories in a specific scene are identified or detected using discriminative models of computer vision to determine whether the scene is safe or not based on their presence or absence\cite{risk_identification_tradition_2}. In addition, some studies consider the relationships between entities based on entities perception and introduce scene graphs to enhance the model reasoning capability, thus improving the model performance\cite{risk_identification_tradition_4}.  Benchmarks for road traffic safety\cite{risk_identification_tradition_1} and construction site safety\cite{risk_identification_tradition_7} have emerged. Overall, these approaches focus on improving the model's recognition or detection performance, and are unable to understand the safety of the scene from a macroscopic perspective. The model in this case learns only the mapping relationship between objects and categories in the scene or between the scene and the safety judgment, and the interpretability of the final results obtained is poor.

Recently, under the development wave of generative artificial intelligence, LLMs, multi-modal large models(MLLMs), etc., have started to contribute to the development of scene safety understanding. Thanks to their extensive knowledge base, they are able to understand input scenes and perform causal reasoning for safety\cite{chatgpt_road_safety_overview}. Currently, the vast majority of research addresses the field of road traffic safety\cite{scene_graph_vlm_road}. Overall, most of the studies are currently at the stage of proposing generative model benchmarks for vertical domains. Their methods of constructing datasets require specialised vertical domain knowledge and are not universal. They do provide safety reasoning processes in textual form to interpret the results, however, they lack the perspective of interpreting from within the model.

\subsection{Process Supervision} 
Process supervision refers to continuous monitoring, guidance and adjustment by providing feedback on the various stages of training a model to accomplish a particular task. Broadly speaking, autoregressive models for language generation tasks embody the idea of process supervision during the training process. However, this is mainly attributed to the directional and process properties of the language and the \textit{next token prediction} of the autoregressive model. 
Narrow process supervision encompasses three key points: task decomposition, intermediate oversight, and transparency and error correction, the last of which is not present in the broad process supervision mentioned earlier.
To achieve transparency and error correction, it is common to train a validator-like model with individual inference steps\cite{process_verifier2}, giving it the ability to evaluate the quality of the inference steps, and then collaborates with the main model to enhance its inference. Process supervision has been shown to have more powerful capabilities in the field of mathematical reasoning compared to outcome supervision\cite{process_supervision,process_supervision3}. As a result, it has also been successfully applied to OpenAI's o1 model. However, research on the paradigm is only at a preliminary stage, and its application is limited to the mathematical domain and relies on reward models. Its potential in other domains has not been fully explored.

\section{Hierarchical Safety Cognitive Structure}
\label{cognition_structures_section}

\begin{figure*}
    \centering
    \includegraphics[width=0.9\linewidth]{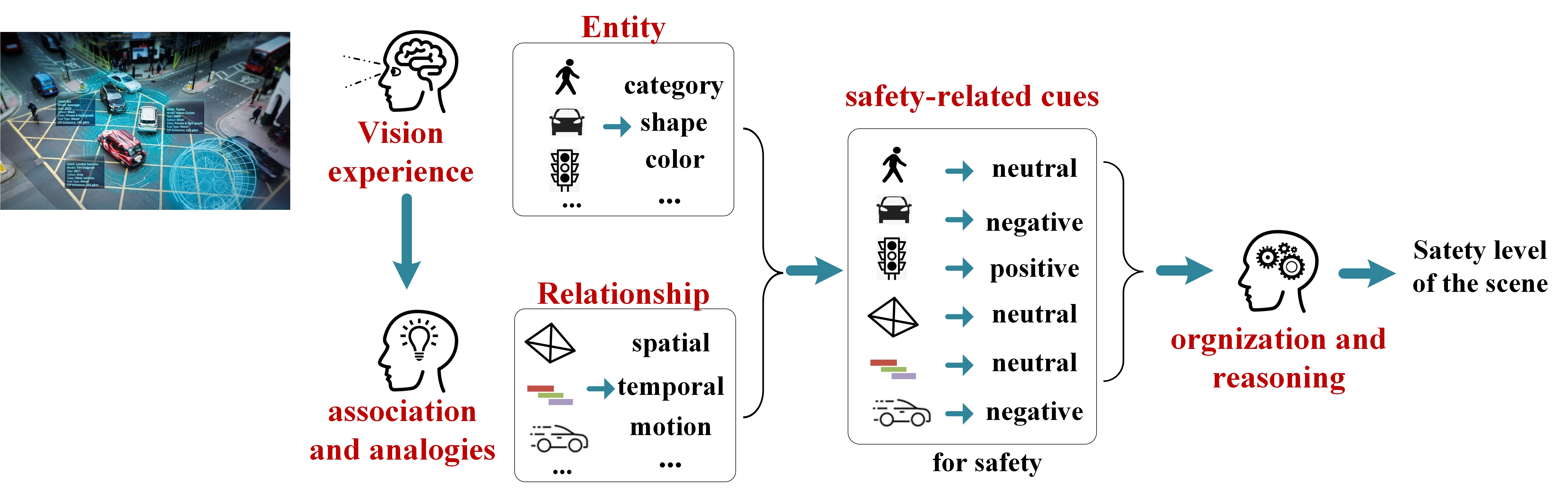}
    \caption{Hierarchical scene safety cognitive structure}
    \label{cognition_structure}
\end{figure*}

A comprehensive understanding of human knowledge structures and cognitive processes is crucial for identifying key mechanisms and information flows within human cognition. This understanding provides the foundation for supervising and guiding machine cognitive processes, thereby enabling more efficient and accurate scene safety understanding. In principle, the human cognitive process begins with bottom-up perception of sensory data, progressing to comprehension, and culminating in reasoning\cite{hierarchical_science}. Human sensory perception serves as the primary mechanism for acquiring information from environmental scenes. This perceptual process encompasses the detection and recognition of diverse entity within the environment. Sensory perception forms the foundational basis for scene comprehension, providing the requisite initial data for higher-order cognitive processes.

Inspired by this progression, our study attempts to delineate this process into distinct stages, proposing the concept of \textit{cognitive hierarchy} to describe the human cognitive process in scene safety understanding, as shown in Figure \ref{cognition_structure}. Specifically, the cognitive hierarchy models the bottom-up process from perception to scene comprehension and reasoning. The lower levels of the hierarchical structure represent the extraction of salient features and patterns from raw perceptual inputs. Intermediate levels capture the formation of conceptual representations and semantic relationships between scene elements based on perceptual features. Finally, the higher levels of the hierarchy build upon useful information obtained from the lower stages, leveraging the underlying implicit knowledge or intuition derived from this information to perform logical reasoning about the overall safety of the scene.

\section{SSUPL Dataset Information}
\label{dataset}
\begin{figure*}[!t]
	\centering
	\includegraphics[width=0.8\linewidth]{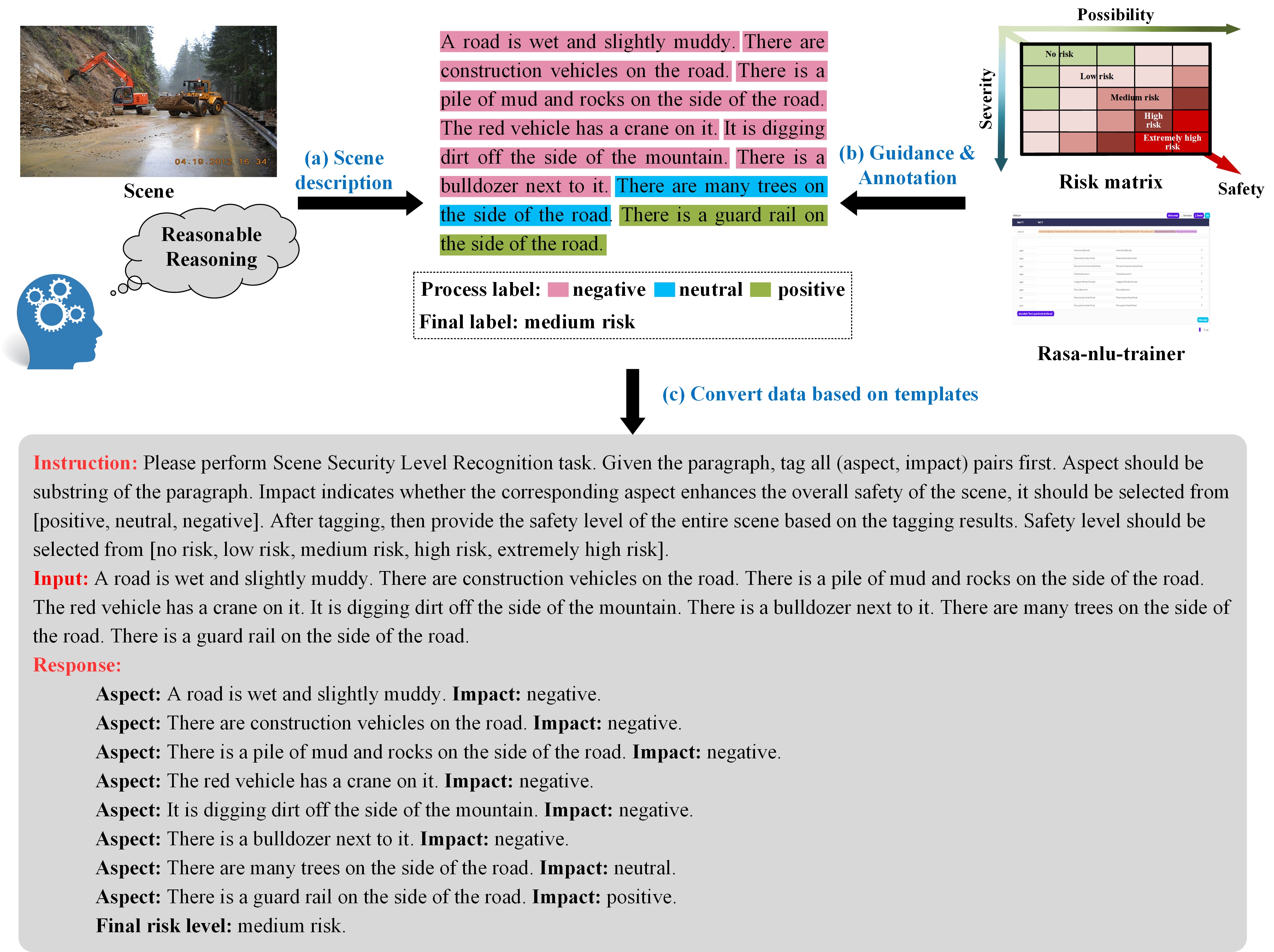}
	\caption{An illustration of the data construction process in the SSUPL dataset. The proposed SSUPL dataset comprises 3,105 scenarios with uniformly distributed final labels across five risk levels, along with 17,464 process labels. The training, test, and validation sets contain 2,100, 705, and 300 samples respectively. The distribution of final labels in the test set is uniform.}
	\label{fig: 1}
\end{figure*}

Based on the above principles, we have constructed a dataset SSUPL for scene safety understanding that simulates human cognitive processes. The dataset is designed in an intuitive and structured manner to facilitate efficient assimilation by machine intelligence. Each sample of the SSUPL consists of four elements: the scene, clauses describing specific entities or relationships in the scene, the safety implications of the entities or relationships (i.e., process labels), and the holistic scene safety level (i.e., final label), as shown in Figure \ref{fig: 1}. The annotator's operational process of constructing the remaining three elements based on the scene reflects the different levels in the cognitive hierarchy, respectively. 

\subsection{Data Construction Process}
We manually collect images with the theme of unexpected moments online, and then utilize the caption anything model\cite{caption_anything} and the rasa-nlu-trainer tool\footnote{https://rasahq.github.io/rasa-nlu-train} to annotate text information. 

\paragraph{Getting descriptions} There are currently three main approaches to obtaining scene description data: utilizing the generative power of MLLMs, manual annotation, and based on existing image caption task datasets. However, text passages generated by MLLMs often exhibit hallucinations and contain numerous subjective descriptions. Manual annotation methods are inefficient. Most of the datasets involved in the image caption task are either sentence-level datasets that roughly describe the scene or dense captioning based datasets that do not have a coherent overall description. Even if there are additional paragraph level Stanford image-paragraph dataset\cite{stanford_caption}, the scenes covered in the dataset are almost always low or no risk. SFT for LLMs requires high-quality of data. We therefore employ a human-machine interaction approach based on the caption anything model\cite{caption_anything} to construct high-quality scene description texts. In this study, we primarily focus on identifying various visual attributes, including size, color, and shape.

\paragraph{Risk matrix} After labeling all the clauses in the paragraph, a risk matrix\cite{risk_matrix} is constructed, which is then used to guide the annotator in labeling the overall safety level of the scene. Scene safety level is a prediction of the severity of incidents that may occur in the scene at some time in future, therefore, the risk matrix classifies the safety level of the scene based on two dimensions: the severity and the possibility of an accident occurring. This paper considers 5 levels of safety: \textit{no risk}, \textit{low risk}, \textit{medium risk}, \textit{high risk}, and \textit{extremely high risk}.

\paragraph{Principles of labeling} 
When labeling the data, we follow the following 4 principles to align cognitive as well as human values: 
\begin{itemize}
	\item \textbf{Human-centered}: Measure the safety of a scene in terms of the physical safety of the humans in the scene.
	\item \textbf{Perspectives on Justice}: If the characters in the scene have different positions, the scene safety level is measured from the righteous side of the scene.
	\item \textbf{No over-reasoning}: In both temporal evolution and spatial extension, do not consider entities and relationships that do not appear in the paragraphs unless their presence is clearly suggested.
	\item \textbf{Maximum risk level}: Humans can understand a scene in multiple different chains of thought, resulting in different risk levels based on the risk matrix. We label the data with the maximum of these risk levels.
\end{itemize}

Moreover, given a scene, we assign it to one of the categories of pre-accident scene, during-accident scene, and post-accident scene, depending on the stage in the hazardous accident cycle.
\begin{itemize}
	\item \textbf{Pre-accident scene}: There is currently no direct harm to humans in the scene, but based on the scene clues, the scene has the potential to cause harm to humans after a period of time. We determine the safety level through a complete risk matrix.
	\item \textbf{During-accident scene}: The scene has a certain accident that happened some time ago. The accident has not yet ended and continues to occur, already with some consequences. Based on the risk matrix, we determine the safety level of the scene based on the severity of the cumulative consequences that may result when the accident is over.
	\item \textbf{Post-accident scene}: An accident of some kind occurred in the scene some time ago, and the accident has ended with some consequences. We categorize the consequences into reversible consequences (e.g., minor and serious injuries) and irreversible consequences (e.g., death), and then determine the safety level of the scene based on the likelihood of a reversible consequence developing into an irreversible consequence and on the rows of the constructed risk matrix whose severity exceeds the reversibility threshold.
\end{itemize}

\subsection{Dataset Statistics}

\begin{figure*}
	\centering
	\subfloat[]{\includegraphics[width=1.7in]{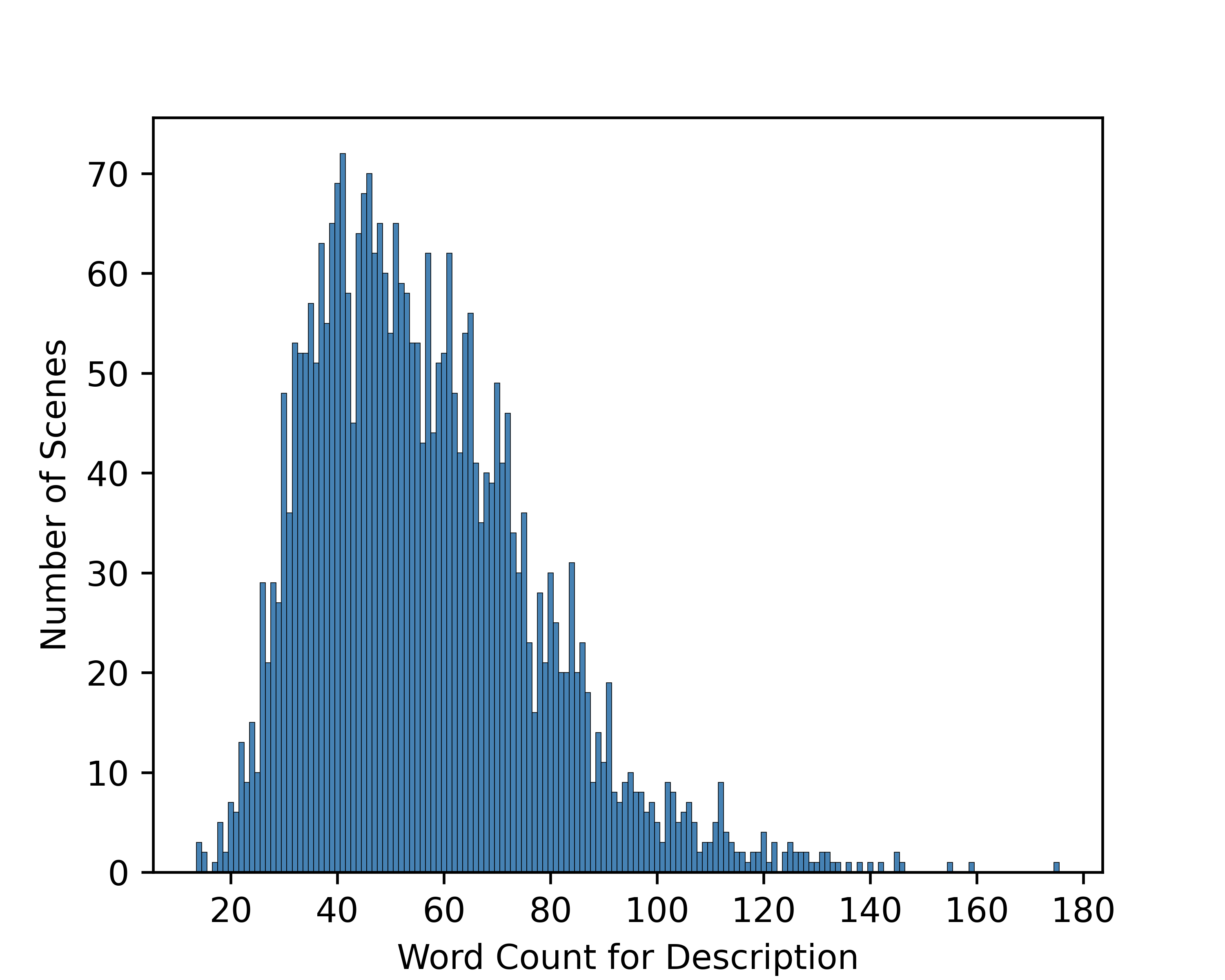}%
		\label{ds1}}
	\hfil
	\subfloat[]{\includegraphics[width=1.7in]{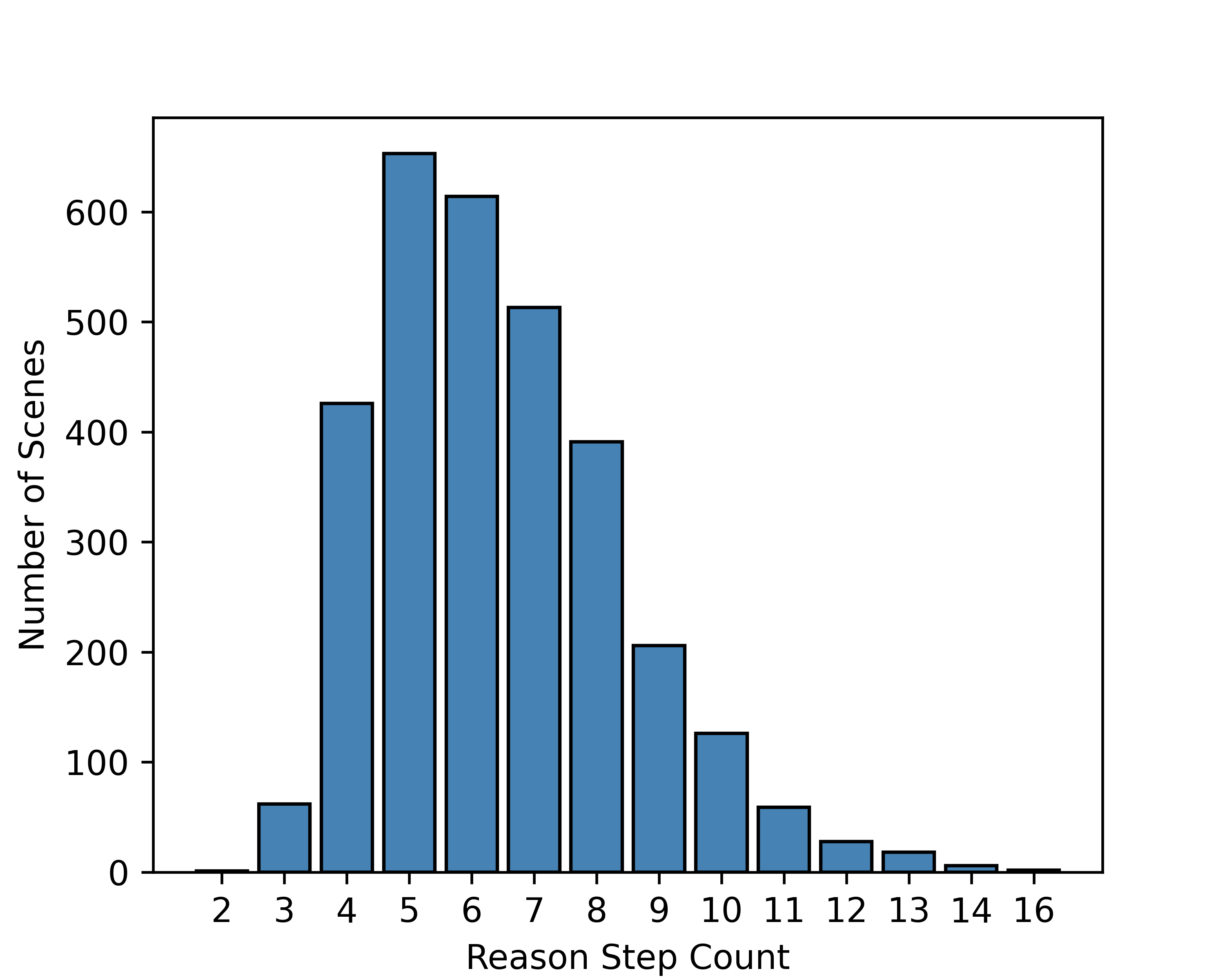}%
		\label{ds2}}
	\hfil
	\subfloat[]{\includegraphics[width=1.7in]{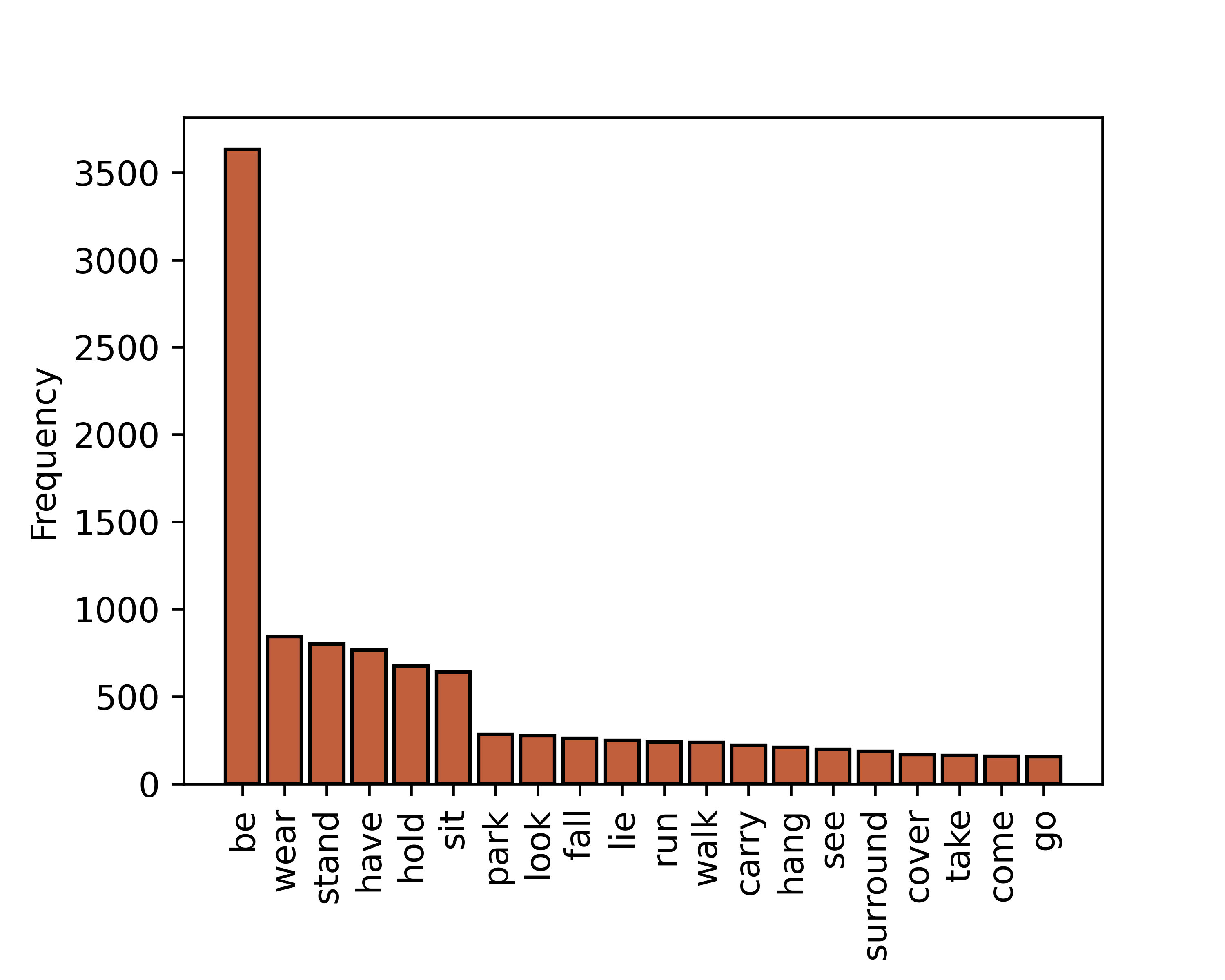}%
		\label{ds3}}
	\hfil
	\subfloat[]{\includegraphics[width=1.7in]{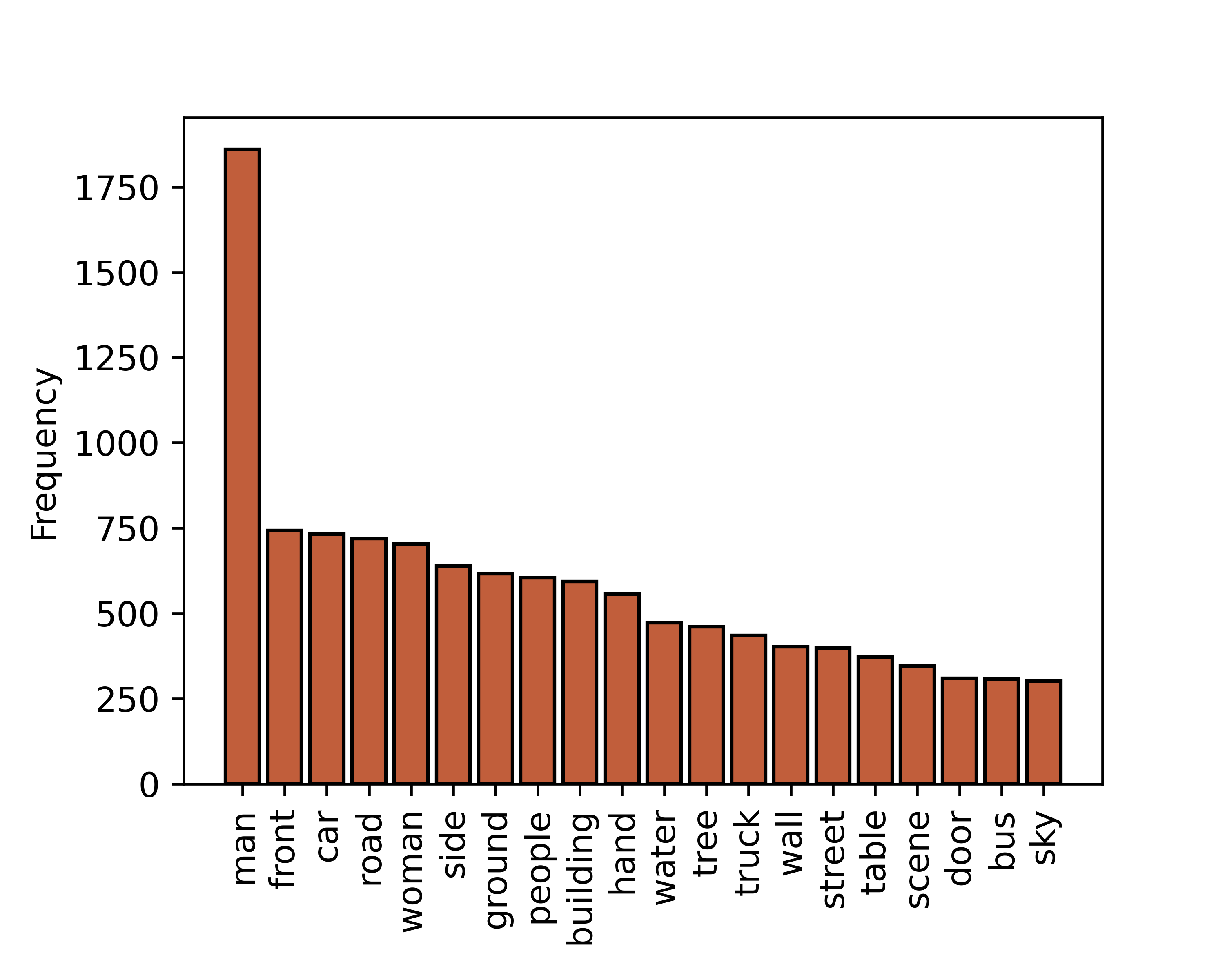}%
		\label{ds4}}
	\caption{Statistical analysis of SSUPL dataset. (a) Word count distribution. (b) Step count distribution. (c) Top 20 verbs distribution. (d) Top 20 nouns distribution.}
	\label{fig: 2}
\end{figure*}

For scene description paragraphs of the SSUPL dataset, the word count ranges from 14 words to 175 words, as detailed in Figure \ref{ds1}. For the safety reasoning process, the count of steps ranges from 2 steps to 16 steps, as detailed in Figure \ref{ds2}. The dataset contains 9608 unique words, including 1035 unique verbs and 2938 unique nouns, ensuring a diversity of entities and relationships. The top 20 verbs and top 20 nouns with the highest frequency of occurrence are shown in Figure \ref{ds3} and Figure \ref{ds4}, respectively. 

In addition, we explore the sentence diversity of the dataset. Specifically, we calculate the average CIDEr-D\cite{cider} similarity between individual sentences in the descriptive paragraph of each scene and subtract the average CIDEr-D similarity from 100 as the final diversity score. The diversity score on the SSUPL dataset reaches 78.63, which is better than the Stanford image-paragraph dataset with a diversity score of 74.22. This indicates that the SSUPL dataset has more diverse utterances, and the sentences in the paragraphs provide more information about the scene.

\begin{table}
	\caption{Distribution of labels in the SSUPL dataset}
	\centering
	\begin{tabular}{lc}
		\hline
		\textbf{Label} & \textbf{Number} \\
		\hline
		no risk & {621}   \\
		low risk & {621}   \\
		medium risk & {621}   \\
		high risk & {621}   \\
		extremely high risk & {621}   \\
		\hline
		negative & {8183}   \\
		neutral & {6517}   \\
		positive & {2164}   \\
		\hline
	\end{tabular}
	\label{label_dis}
\end{table}

\section{Inference Process of Machine Intelligence According to Safety Cognitive Hierarchy}
\label{interpretability_analysis}

In this section, we begin by outlining the token output process of SFT LLM in safety reasoning. Following this, we conduct a detailed analysis of information transfer within the SFT LLM from the perspective of information flow, illustrating how the underlying mechanisms align with the hierarchical cognitive process structure. Moreover, we examine the critical factors that shape the final safety prediction, providing insights into their respective influences and interactions.

It is necessary to note that despite the predominance of visual information in human perception, machine systems face fundamental challenges in directly comprehending visual experiences: significant disparities exist between machine and human visual perception mechanisms, and machines struggle to capture the subjectivity and diversity inherent in visual experiences. Converting visual information into textual representations offers structured semantic data that bridges the gap between low-level visual features and high-level semantic concepts, while facilitating more effective integration of human prior knowledge, thereby enhancing model generalization capabilities and interpretability. Therefore, this study adopts a language-mediated modal decoupling paradigm\cite{image_to_text1}, utilizing
a mature image captioning model to convert images into descriptive text, and then employing the LLM to perform scene safety inference based on this text. The research focuses on enhancing the reasoning capabilities of LLMs.

\subsection{Problem Formulation}

The SFT LLM is capable of performing scene safety understanding in a structured format. Given a scene $i$, its textual description $S_i$ is provided by an image caption model and then fed into the SFT LLM. The SFT LLM then autoregressively generates $k_i$ judgments regarding the influence of entities or relationships on scene safety, forming a reasoning process represented as:
\begin{equation}
    R_i=\{(x_{i,j},y_{i,j})|1\leq j \leq k_i, j \in \mathbb{Z}, \sum_{j=1}^{k_i} x_{i,j}=S_i ,y_{i,j} \in \mathbb{P} \}
\end{equation}
where $x_{i,j}$ represents the words in the $j$-th clause describing an entity or entity relationship, and $y_{i,j}$ represents the corresponding process label words for $x_{i,j}$. The process label space $\mathbb{P}$ is defined as:
\begin{equation}
\mathbb{P}=\{negative, neutral, positive\}
\end{equation}
To predict the final safety level of scene $i$, we define $W_i$ and $F_i$ as the SFT LLM prompts and the final prefix words that guide the prediction, respectively. The input to the model is formulated as:
\begin{equation}
    C_i=W_i\oplus S_i \oplus T(R_i) \oplus F_i
\end{equation}
where the $T(\cdot)$ is a function that appends process prefix words, and $\oplus$ denotes concatenation. The final safety level of scene $i$ predicted by the SFT LLM based on greedy decoding strategy is:
\begin{equation}
    \hat{r_i}=\arg\max_{r_i \in \mathbb{D}}\mathcal{P}(r_i|C_i; \theta)
\end{equation}
where $\theta$ represents the parameters of the SFT LLM, $\mathcal{P}$ denotes the probability distribution produced by the model for safety level prediction, and $\mathbb{D}$ is the final label space, defined as:
\begin{equation}
    \mathbb{D}=\{no\;risk,\dots, extremely\;high\;risk\}
\end{equation}

\subsection{Information Flow Analysis}
\label{info}

\begin{figure*}
    \centering
    \includegraphics[width=12cm]{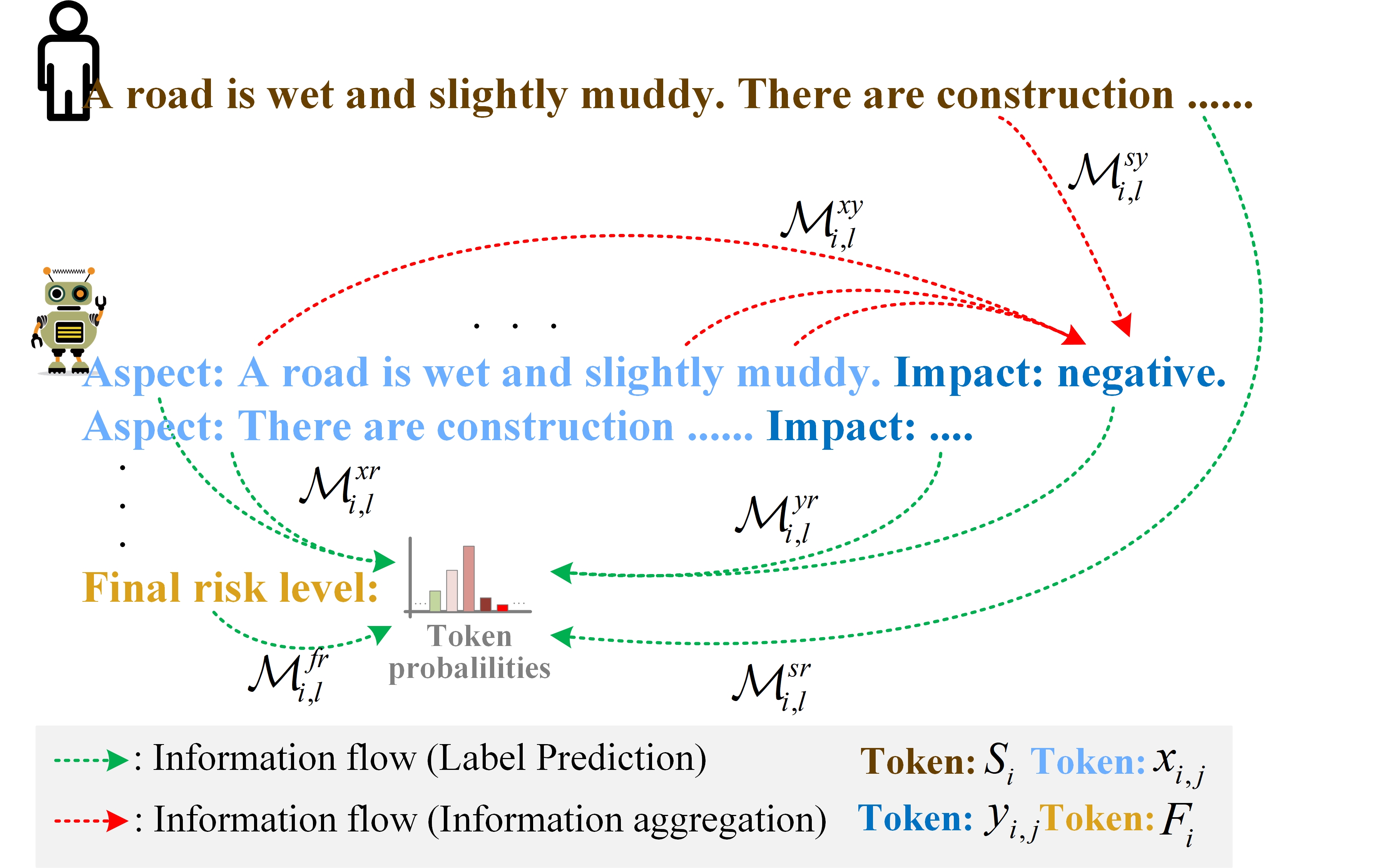}
    \caption{Schematic of the information flow for SFT LLM to predict the safety level of a test sample.}
    \label{information_flow}
\end{figure*}

Building upon the prior works\cite{label_anchor}, this study investigates the mechanisms underlying information transfer during the reasoning process within SFT LLMs. Specifically, we analyze the key factors influencing the prediction of the final safety level from the perspective of information flow. To quantify information flow within the model, we employ saliency scores between tokens, which serve as a measure of the significance of information transfer. In classification tasks, these scores reflect the contribution of individual components to the final classification outcome. For a given scene $i$, the saliency score matrix $I_{l}^i$ at the $l$-th layer of the SFT LLM is computed as follows:
\begin{equation}
    I_{l}^i=\left|\sum_{h} A_{l, h}^i \odot \frac{\partial \mathcal{L}(x)}{\partial A_{l, h}^i }\right| .
\end{equation}
where $A_{l, h}^i$ denotes the attention weight matrix associated with the $h$-th head in the $l$-th layer, and $\mathcal{L}(x)$ represents the cross-entropy loss function. The resulting $I_l^i$ is a two-dimensional lower triangular matrix, where each element $I_l^i(m,n)$ characterizes the significance of the information flow from the $n$-th token to the $m$-th token.

\subsubsection{Information Flow from Process Labels}
Given the critical role of process labels, we first compute the mean importance of information flow from tokens in $x_{i,j}$ and $S_i$ to tokens in $y_{i,j}$, denoted as $\mathcal{M}_{i,l}^{xy}$ and $\mathcal{M}_{i,l}^{sy}$, respectively. Furthermore, to establish a baseline for evaluating information flow intensity, we analyze the mean importance of information flow among normal tokens, denoted as $\mathcal{M}_{i,j}^{xx}$. They are formulated as:

\begin{equation}
    \left\{
    \begin{array}{l}
            \mathcal{M}_{i,l}^{xy}=\frac{\sum_{(m,n)\in \mathbb{C}^{xy}_i} I_l^i(m,n)}{\left|\mathbb{C}^{xy}_i\right|} ,\\
           \mathcal{M}_{i,l}^{sy}=\frac{\sum_{(m,n)\in \mathbb{C}^{sy}_i} I_l^i(m,n)}{\left|\mathbb{C}^{sy}_i\right|} ,\\
           \mathcal{M}_{i,l}^{xx}=\frac{\sum_{(m,n)\in \mathbb{C}^{xx}_i} I_l^i(m,n)}{\left|\mathbb{C}^{xx}_i\right|} .
    \end{array}
    \right.
\end{equation}
where $\mathbb{C}^{xy}_i=\{(q_m,q_n)|m \in y_{i,j}, n \in x_{i,j}, 1\leq j \leq k_i, j \in \mathbb{Z}\}$, $\mathbb{C}^{sy}_i=\{(q_m,q_n)|n \in S_i, m \in y_{i,j}, 1\leq j \leq k_i, j \in \mathbb{Z}\}$, $\mathbb{C}^{xx}_i=\{(q_m,q_n)|m, n \in x_{i,j}, q_n < q_m, 1\leq j \leq k_i, j \in \mathbb{Z}\}$. Here, $q_n$ represents the position of token $n$. 

\subsubsection{Information Flow for Final Safety Level Prediction}
To analyze the information flow contributing to the final safety level prediction, we consider four sets of tokens: $S_i$, $x_{i,j}$, $y_{i,j}$, and $F_i$, representing different components of the SFT LLM input and inference process (as illustrated in Figure \ref{information_flow}). The mean importance of the information flow from these sets to the final safety level prediction is denoted as $\mathcal{M}_{i, l}^{sr}$, $\mathcal{M}_{i, l}^{xr}$, $\mathcal{M}_{i, l}^{yr}$, $\mathcal{M}_{i, l}^{fr}$, respectively. These are defined as follows:

\begin{equation}
    \left\{
    \begin{array}{l}
    \mathcal{M}_{i, l}^{sr}=\frac{\sum_{(m,n)\in \mathbb{C}^{sr}_i} I_l^i(m,n)}{\left|\mathbb{C}^{sr}_i\right|} ,\\
    \mathcal{M}_{i, l}^{xr}=\frac{\sum_{(m,n)\in \mathbb{C}^{xr}_i} I_l^i(m,n)}{\left|\mathbb{C}^{xr}_i\right|} ,\\
    \mathcal{M}_{i, l}^{yr}=\frac{\sum_{(m,n)\in \mathbb{C}^{yr}_i} I_l^i(m,n)}{\left|\mathbb{C}^{yr}_i\right|} ,\\
        \mathcal{M}_{i, l}^{fr}=\frac{\sum_{(m,n)\in \mathbb{C}^{fr}_i} I_l^i(m,n)}{\left|\mathbb{C}^{fr}_i\right|}
    \end{array}
    \right.
\end{equation}
where $\mathbb{C}^{sr}_i=\{(q_{\hat{r_i}},q_n)|n \in S_i \}$, $\mathbb{C}^{xr}_i=\{(q_{\hat{r_i}},q_n)|n \in x_{i,j}, 1\leq j \leq k_i, j \in \mathbb{Z}  \}$, $\mathbb{C}^{yr}_i=\{(q_{\hat{r_i}},q_n)|n \in y_{i,j}, 1\leq j \leq k_i, j \in \mathbb{Z}  \}$, $\mathbb{C}^{fr}_i=\{(q_{\hat{r_i}},q_n)|n \in F_i\}$.

\subsubsection{Experimental Analysis of Information Flow Mechanism on SFT LLM}
\label{get_sft_model}
To investigate the internal information flow mechanisms and their impact on final safety reasoning ability, we selected five widely used open-source models: LLama2-7B, which features 32 hidden layers from the LLama family\cite{llama2}; GPT-J, consisting of 28 layers from the GPT family\cite{gptj}; Gemma2-9B, which includes 42 layers from the Gemma family\cite{gemma}; Intern2.5-7B, incorporating 32 layers from the Intern family\cite{intern}; and Baichuan2-7B, which comprises 32 layers from the Baichuan family \cite{baichuan}.

We first perform SFT on the selected LLMs using the training set. The SFT LLMs are subsequently employed to conduct information flow experiments on the test set. Given hardware constraints, we utilize Low-Rank Adaptation (LoRA)\cite{lora} for SFT and integrate the resulting adapter into the base model. The LoRA rank is set to 8, the number of training epochs is set to 5, the initial learning rate is set to $5\times 10^{-5}$, and the warmup ratio is set to 0.1. All experiments are conducted using two NVIDIA A6000 GPUs (48GB memory each) and are implemented via LLama-factory\cite{llamafactory} with our modifications. For Llama2-7B-chat, Gemma2-9B-it, and GPT-J, we designate $q\_proj$ and $v\_proj$ as LoRA target modules. For Baichuan2-7B-chat and Intern2.5-7B-chat, LoRA is applied to $all$ modules.

\begin{figure}
    \centering
    \subfloat[]{\includegraphics[width=2.5in]{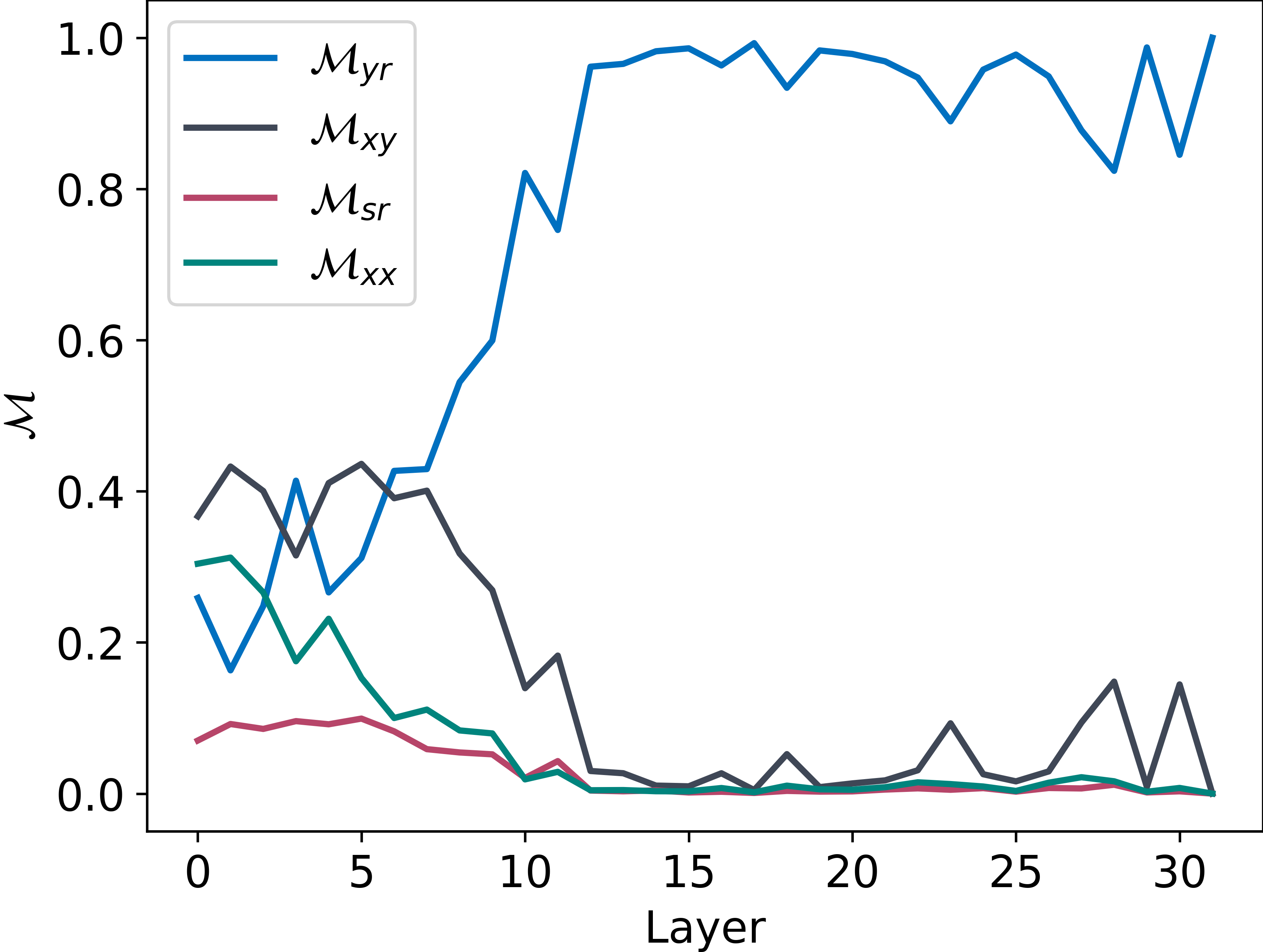}%
    \label{saliency_diff_part_for_process}}
     \hfill
    \subfloat[]{\includegraphics[width=2.5in]{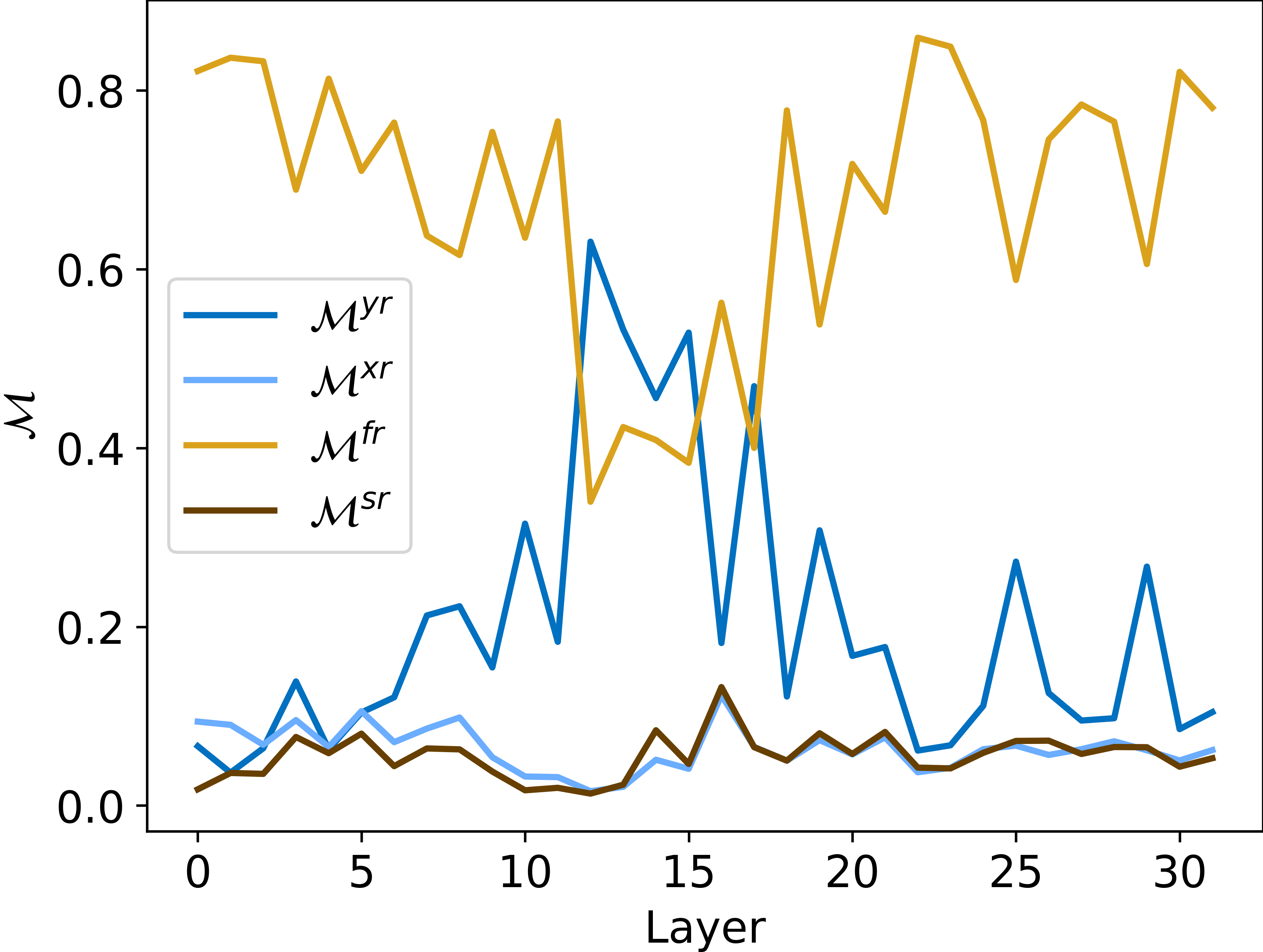}%
    \label{saliency_diff_part}}
    \caption{Information flow. (a) The information flow reflecting hierarchical cognitive structure in Llama2-7b-chat-sft. (b) Importance of information flow for each layer in Llama2-7b-chat-sft for the prediction of final scene safety level. Results for other models can be found in Appendix \ref{information_results_one_other_llms} and Appendix \ref{information_results_two_other_llms}.}
\end{figure}

Figure \ref{saliency_diff_part_for_process} illustrates the role of information flow associated with process labels across different layers of the SFT LLMs, as well as the significance of information flow among normal tokens. It can be seen that $\mathcal{M}_{xy}$ dominates in the shallow layer, accounting for about 40\%, while $\mathcal{M}_{xx}$, $\mathcal{M}_{yr}$ and $\mathcal{M}_{sr}$ account for about 30\%, 20\% and 10\%, respectively. As the layer deepens, $\mathcal{M}_{yr}$ approximates monotonically increasing and converges to 1. In contrast, the other information flows approximately monotonically decrease and converge to 0. These phenomena indicate that process labels serve as semantic anchors throughout the reasoning process. In the shallow layers, they facilitate the aggregation of information from relevant clauses and model inputs, contributing to the formation of semantic representations. In the deeper layers, the SFT LLMs extract information from these process labels, which are rich in contextual cues, to generate the final prediction. This hierarchical mechanism aligns with the hierarchical safety cognitive structure outlined in Section \ref{cognition_structures_section}.

Figure \ref{saliency_diff_part} further examines the influence of input words and distinct components of the inference process on the final prediction across all layers of the SFT LLMs. It can be seen that the change of $\mathcal{M}_{fr}$ shows a pattern of decreasing and then increasing as the layers deepen, with a stable leadership position in the shallow and deep layers, accounting for between 60\% and 80\% of the total. The change of $\mathcal{M}_{yr}$ is completely opposite, exceeding 40\% in the middle layers, thus replacing $\mathcal{M}_{fr}$ to occupy the leadership position. In addition, the share of $\mathcal{M}_{sr}$ and $\mathcal{M}_{xr}$ is consistently low. In summary, the key findings are as follows:

(1) The clause component of the inference process ($x_{i,j}$) exhibits a functional similarity to the model input ($S_i$), both exerting a relatively indirect influence on the final prediction.

(2) In both shallow and deep layers, prefix words that introduce the final prediction (e.g., "Final risk level:") exert the most significant impact. This pattern is analogous to the phenomenon in ICL\cite{reason_step_impact}. By applying the information flow blocking method\cite{label_anchor}, we observe that words outside the five predefined safety levels occasionally appear in the final prediction. This finding suggests that these prefix words function as category constraints, effectively narrowing the classification space and ensuring that the five safety levels consistently rank among the most probable words in the vocabulary.

(3) Compared to the normal textual elements, process labels play a particularly critical role during inference, especially in the middle layers of the model. This observation further reinforces their function as semantic anchors, as demonstrated in Figure \ref{saliency_diff_part_for_process}.

\subsection{Distribution Correlation Analysis}
\label{correlation_analysis}

From a cognitive perspective, the cumulative distribution of hazards within a given scene is closely correlated with the overall distribution of safety. Specifically, a scene characterized by a significant number of hazardous objects or a particularly dangerous object is likely to present an extremely high risk level. This subsection intends to translate this cognitive understanding into a reliable framework of prior knowledge within the context of machine intelligence, thereby informing and guiding subsequent research.

\subsubsection{Distribution Formulation}

In the context of this study, the intensity of a particular entity/relationship with a specific process label can be represented by the attention given to it by the final safety level. We use $A_l^i$ to represent $\frac{1}{H}\sum_{h} A_{l, h}^i $, where $H$ is the number of heads. $\mathbb{S}^{1}_{i,l}$ and $\mathbb{S}^{2}_{i,l}$ denote the attention distributions on process label tokens and normal text tokens guided by corresponding process labels when predicting the final safety level, respectively. $P_i$ represents the probability distribution of the final prediction result. $\mathbb{S}^{1}_{i,l}$, $\mathbb{S}^{2}_{i,l}$ and $P_i$ are shown as follows:

\begin{equation}
\begin{aligned}
    \mathbb{S}^{1}_{i,l}\negthickspace=\negthickspace &\left[ \sum_{j=1}^{k_i}\negthickspace \! \sum\limits_{\substack{ t \in y_{i,j}\\ y_{i,j}=\mathbb{P}_1}} \negthickspace \! \! \!  A_l^i(q_{\hat{r_i}},\!q_t),\!\! \sum_{j=1}^{k_i} \negthickspace \!  \sum\limits_{\substack{ t \in y_{i,j}\\ y_{i,j}=\mathbb{P}_2}} \negthickspace  \! \!\!  A_l^i(q_{\hat{r_i}},\!q_t), \!\!\sum_{j=1}^{k_i} \negthickspace \!  \sum\limits_{\substack{ t \in y_{i,j}\\ y_{i,j}=\mathbb{P}_3}} \negthickspace  \! \! \!A_l^i(q_{\hat{r_i}},\!q_t) \right]
\end{aligned}
\end{equation}

\begin{equation}
\begin{aligned}
\mathbb{S}^{2}_{i,l}\negthickspace=\negthickspace&\left[ \sum_{j=1}^{k_i} \negthickspace \!  \sum\limits_{\substack{t \in x_{i,j}\\y_{i,j}=\mathbb{P}_1}} \negthickspace \! \! \!  A_l^i(q_{\hat{r_i}},\! q_t), \!\!\sum_{j=1}^{k_i} \negthickspace \!\sum\limits_{\substack{t \in x_{i,j}\\y_{i,j}=\mathbb{P}_2}} \negthickspace \! \! \! A_l^i(q_{\hat{r_i}},\!q_t), \!\!\sum_{j=1}^{k_i} \negthickspace \! \sum\limits_{\substack{t \in x_{i,j}\\y_{i,j}=\mathbb{P}_3}} \negthickspace \! \! \! A_l^i(q_{\hat{r_i}},\!q_t) \right]
\end{aligned}
\end{equation}

\begin{equation}
\begin{aligned}
P_i\negthickspace=\negthickspace\left[\mathcal{P}(r_i=\mathbb{D}_1 | C_i;\theta), \mathcal{P}(r_i=\mathbb{D}_2 | C_i;\theta),\dots, \mathcal{P}(r_i=\mathbb{D}_5 | C_i;\theta) \right]\\
\end{aligned}
\end{equation}

\subsubsection{Correlation Indicator}

We adopt the AUC-ROC score to measure the correlation between attention distributions and model predictions of safety levels. Specifically, the AUC-ROC score $D_{i,l}^n$ is calculated for $\mathbb{S}^{n}_{i,l}\cdot\mathcal{T}_n+\mathcal{H}_n,n\in\left\{1,2\right\}$ and $\mathcal{P}_i$, where  $n \in \{1,2\}$. $\mathcal{T}_n \in \mathbb{R}^{\left| \mathbb{P} \right| \times \left| \mathbb{D} \right|}, \mathcal{H}_n \in \mathbb{R}^{ \left| \mathbb{D} \right|}$ are the global transformation matrices, which can be additionally learned by the training samples of SFT. Further, we introduce the metric $\mathcal{C}_{i,l}^n$ to measure the cumulative contribution of the first $l$ layers of the model to the prediction of the final safety level\cite{label_anchor}. It can be expressed as:

\begin{equation}
    \mathcal{C}_{i,l}^n=\frac{\sum_{j=1}^{l}(D_{i,j}^n-0.5)}{\sum_{i=1}^{L}(D_{i,j}^n-0.5)},n\in\left\{1,2\right\}
\end{equation}
where $L$ is the number of layers of the SFT LLM.

\subsubsection{Experimental Analysis of Correlation on SFT LLM}
\label{results_correlation}

Here, we perform experiments on the SFT LLMs mentioned in Section \ref{get_sft_model}. We define simple neural networks and train $\mathcal{T}_n$ and $\mathcal{H}_n$ with training samples from SFT, and finally test the results on the test set of SFT. A learning rate of 0.01 is used, and the number of training epochs is set to 1000.

\begin{figure}
    \centering
    \subfloat[]{\includegraphics[width=2.5in]{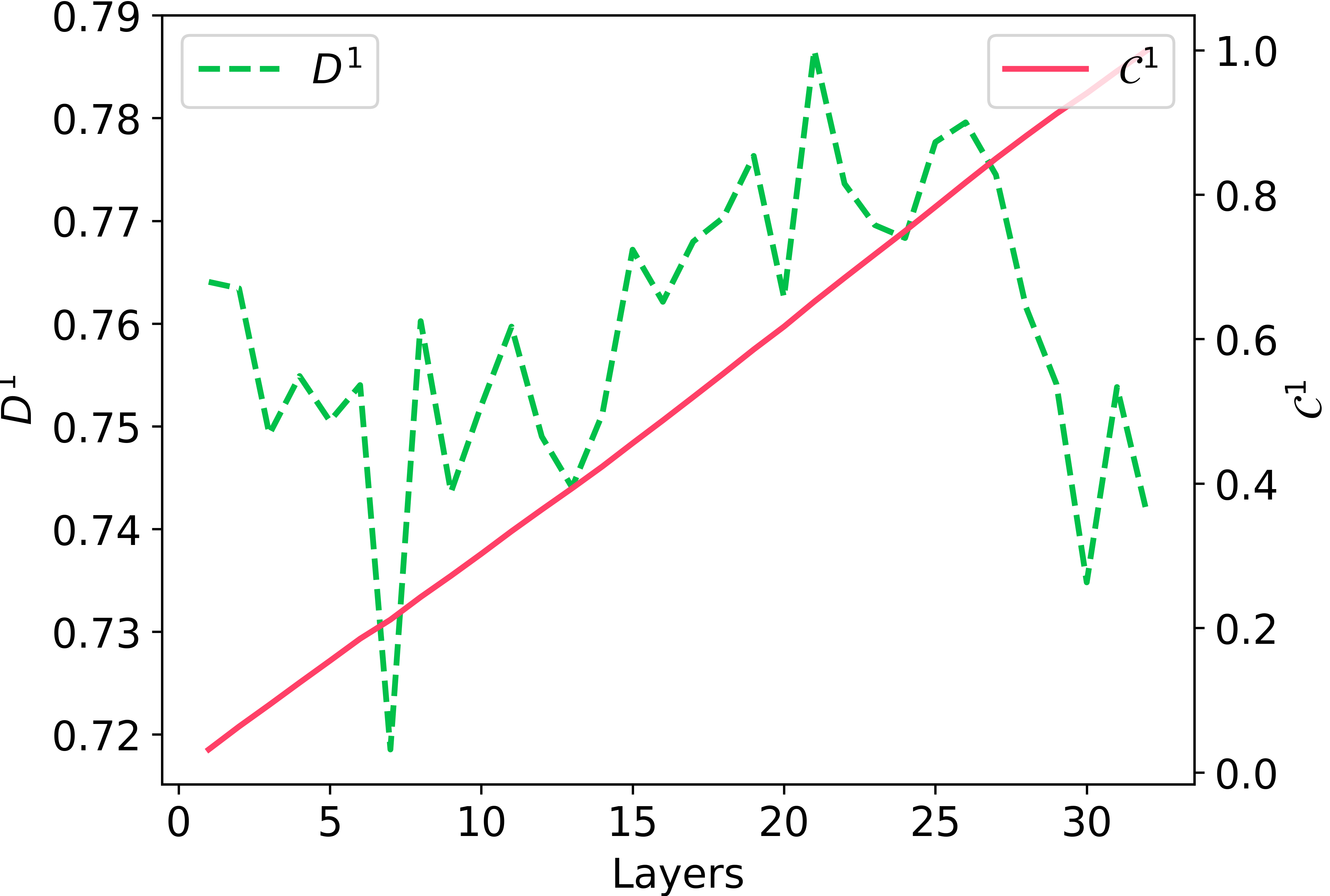}%
    \label{key_factor_final_p}}
    \hfill
    \subfloat[]{\includegraphics[width=2.5in]{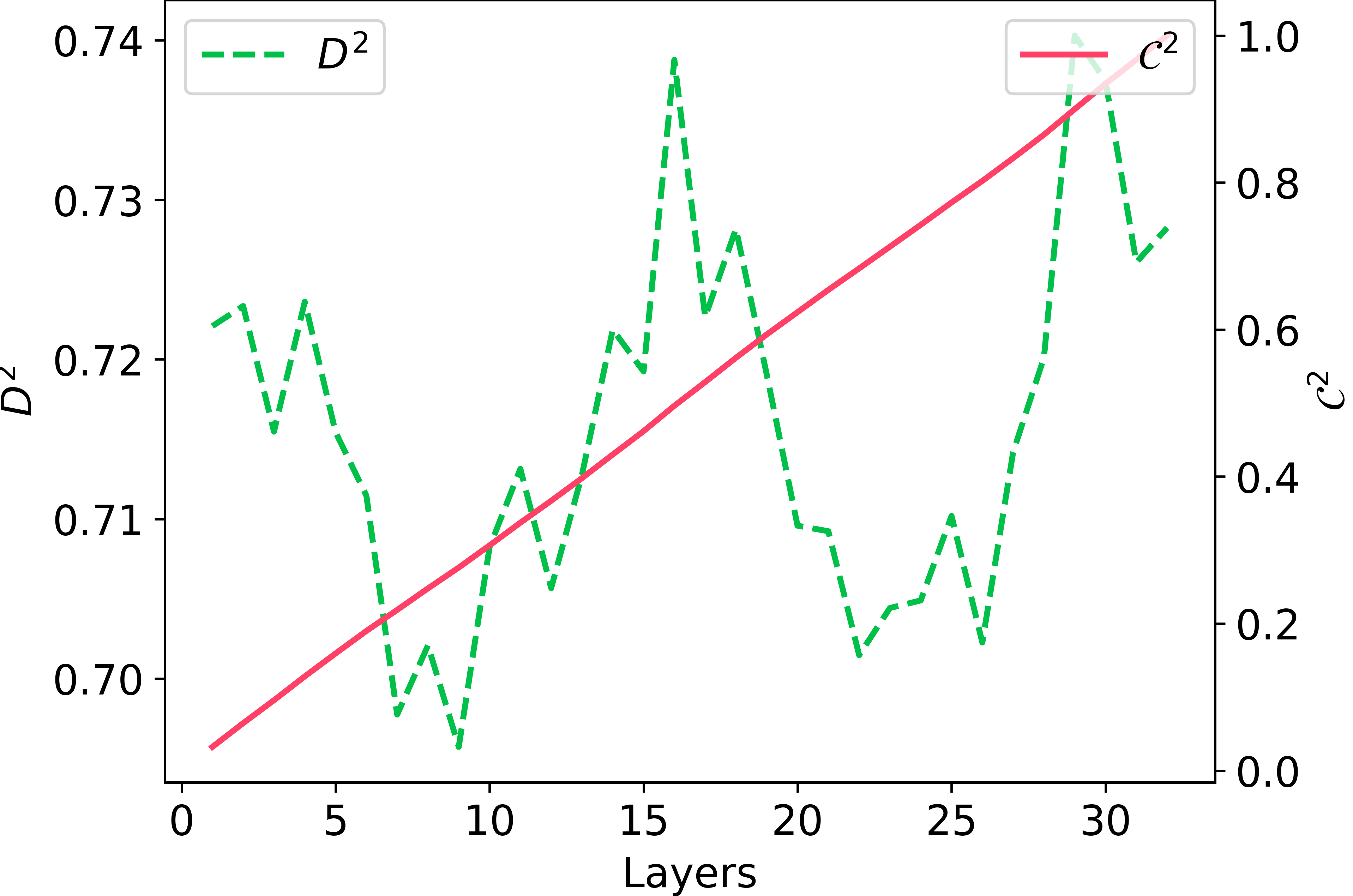}%
    \label{key_factor_final_n}}
    \caption{Correlation analysis results. (a) $\mathcal{C}^1_{l}$ and $D^1_l$ of different layers in Llama2-7b-chat-sft. (b) $\mathcal{C}^2_{l}$ and $D^2_l$ of different layers in Llama2-7b-chat-sft.  Results for other models can be found in Appendix \ref{correlation_results_one_other_llms} and Appendix \ref{correlation_results_two_other_llms}.}
\end{figure}

Figures \ref{key_factor_final_p} and \ref{key_factor_final_n} depict the correlation metric and cumulative contribution in each layer of the SFT LLMs involving the process labels and the normal text, respectively. We find that the values of $D^1$ and $D^2$ are higher than 0.7 in most layers of the SFT LLMs, indicating that both the attention distributions on process label tokens and normal text tokens are strongly correlated with the final predictions of the models\cite{label_anchor}. Moreover, based on the values of $\mathcal{C}^1$ and $\mathcal{C}^2$, we find that each layer in the SFT LLMs can make a positive contribution to the final prediction. In summary, it suggests that machine intelligence for scene safety understanding under LLM is highly aligned with human cognitive intuition, which can be regarded as a reliable prior knowledge.

\section{Process Supervision Model based on Hierarchical Cognitive Structure}

Subject to the autoregressive properties of LLMs and the process properties of the linguistic modality, the direct implementation of SFT for LLMs using SSUPL in Section \ref{interpretability_analysis} constitutes a formal-level process supervision approach. However, this formal-level process supervision method only supervises at the formal level, emphasizing the matching of process labels without deeply modeling cognitive-level reasoning mechanisms. This means that the models may have learned process patterns superficially but lack understanding of the underlying logic and reasoning pathways. Additionally, its training approach uses end-to-end global parameter optimization for overall adjustment, making the model difficult to decompose during the inference process and consequently challenging to adapt to complex reasoning tasks.

The objective of this study is to construct a progressively modular machine model that facilitates the development of hierarchical cognitive structures through process-supervised training of modular components. This methodology is designed to enhance the machine model with human-like reasoning and cognitive capabilities. Two fundamental challenges must be addressed to achieve this objective: (1) functional modularity: structuring the model into specialized functional components to enhance interpretability and efficiency. (2) distributed collaboration: ensuring seamless cooperation between different modules to optimize performance. Fortunately, the LoRA framework\cite{lora} and the MoE paradigm\cite{moe} are well-aligned with the technical architecture of LLMs: LoRA facilitates the efficient generation of adapters, enabling functional modularity, and MoE provides a flexible mechanism for dynamic multi-module collaboration. Through functional modularity and distributed collaboration, combined with LoRA and MoE technologies, this study aims to construct a model architecture that better simulates human hierarchical cognitive structures, thereby enhancing safety assessment and decision reliability in complex environments.

\subsection{Main Architecture of the Model}

\begin{figure*}
    \centering
    \includegraphics[width=0.9\linewidth]{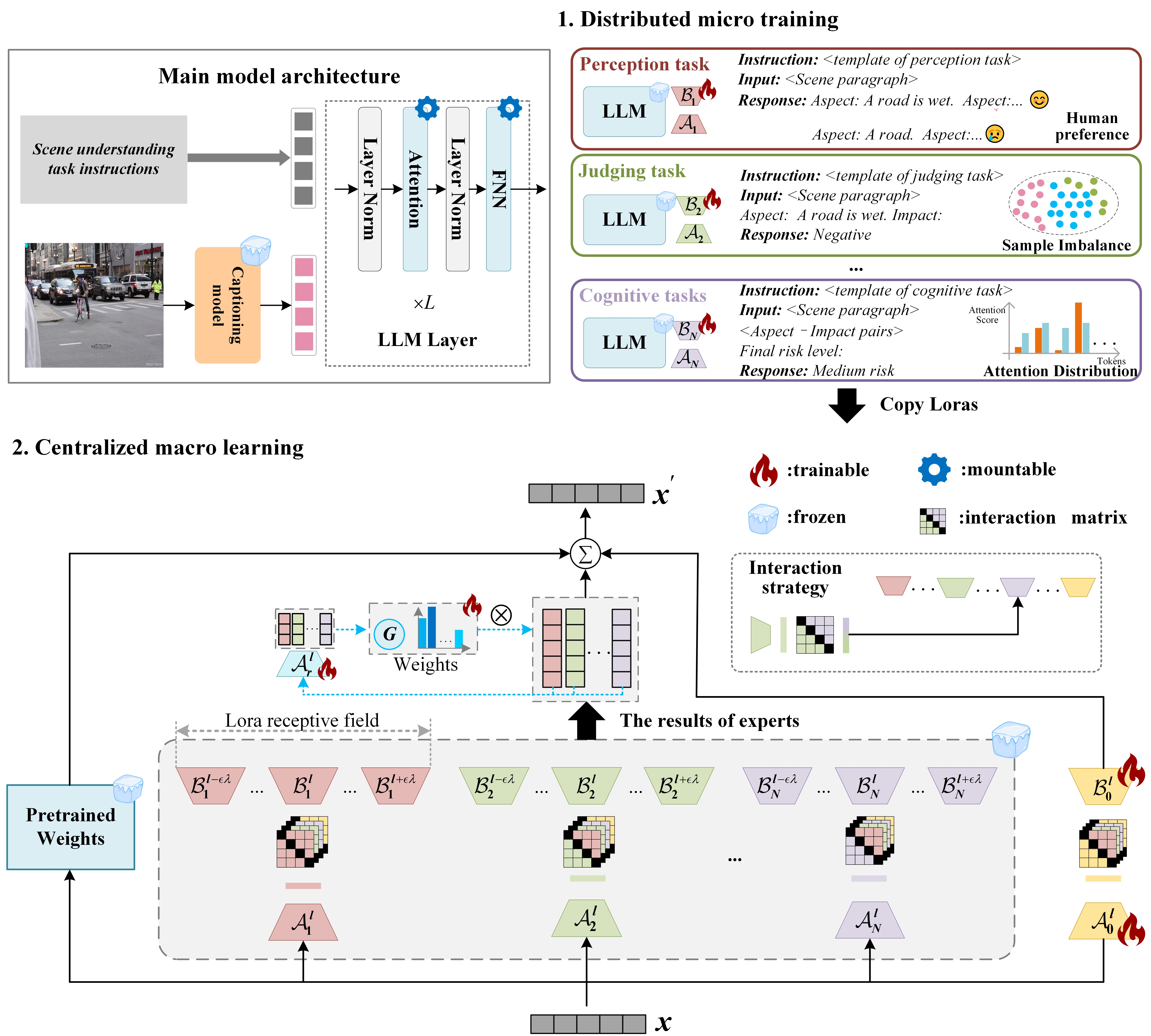}
    \caption{Our proposed LoRA and MoE-based process-supervised training framework for scene safety understanding.}
    \label{our_mehod}
\end{figure*}

A typical model architecture for scene safety understanding and the process-supervised training framework reflecting the cognitive hierarchy model is depicted in Figure \ref{our_mehod}. The framework utilizes an existing image caption model to convert
scene images into textual captions, focusing on optimizing and utilizing the reasoning capabilities of the LLM to achieve safety assessment. This process-supervised training framework comprises two critical training phases: distributed micro training and centralized macro learning.

\paragraph{Distributed micro training} In this phase, the objective is to train a set of LoRA modules using data from single-level tasks, ensuring that each LoRA specializes in a specific cognitive-level task. These trained modules will subsequently serve as experts in the next stage. The training data for a single-level task incorporates the results of the preceding task as conditional inputs, as illustrated in Figure \ref{our_mehod}. 

For a pre-trained module at layer $l$, denoted by the weight matrix $\mathcal{W}_0^l \in \mathbb{R}^{d_1 \times d_2}$, the LoRA technique updates the module weights by means of $\mathcal{W}_0^l+\Delta \mathcal{W}^l=\mathcal{W}_0^l+\mathcal{A}^l\mathcal{B}^l$, where $\mathcal{A}^l \in \mathbb{R}^{d_1 \times \xi}, \mathcal{B}^l \in \mathbb{R}^{\xi \times d_2},\xi \ll min(d_1,d_2)$. During training, the original weights $\mathcal{W}_0^l$ remain frozen, and only $\mathcal{A}^l$ and $\mathcal{B}^l$ are trained, enabling efficient parameter fine-tuning with minimal computational overhead. Upon completion of this stage, a LoRA set $\Omega=\{(\mathcal{A}_i,\mathcal{B}_i)\}_{i=1}^{N}$ is obtained, where each pair $(\mathcal{A}_i,\mathcal{B}_i)$ is responsible for a specific low-level cognitive task, $\mathcal{A}_i=\{\mathcal{A}_i^1,\mathcal{A}_i^2,...,\mathcal{A}_i^L\}$, $\mathcal{B}_i=\{\mathcal{B}_i^1,\mathcal{B}_i^2,...,\mathcal{B}_i^L\}$, and $N$ is the number of tasks. In this study, based on the proposed cognitive hierarchical model, the scene safety understanding task is decomposed into three distinct cognitive tasks: perception, judgment, and cognition, corresponding to $N=3$. For detailed LoRA training templates for these tasks, refer to Appendix \ref{templates_icl_cot}.

\paragraph{Centralized macro learning} This stage aims to learn a macro-expert that can perform cognitive tasks from the lowest to the highest level of cognition and the way it collaborates with the micro-expert. Denote the macro-expert by $(\mathcal{A}_0,\mathcal{B}_0)$. Based on the idea of MoE, in this phase, the micro-experts in the $\Omega$ are first copied into the pre-trained module of the corresponding layer and their parameters are frozen, as shown in Figure \ref{our_mehod}. The input $x \in \mathbb{R}^{d_1}$ outputs $x^{\prime} \in \mathbb{R}^{ d_2}$ after the experts' operations of downscaling, interaction, upscaling and collaboration, which can be flexibly designed under this framework.

\textbf{Lora receiptive field}: The two modules $\mathcal{A}^l$ and $\mathcal{B}^l$ of Lora contain implications for feature extraction and extra knowledge\cite{lora_feature2}, respectively. Individual layers of a LoRA exhibit unique features, and all layers accumulate to define the overall characteristics of the LoRA\cite{lora_feature2}. Therefore, the architecture of this paper introduces cross-layer micro-expert modules and proposes the concept of the \textit{lora receiptive field}, so as to break the limitation of same-layer correspondences and mix the learning features of each layer to enhance the capture of relevant information. Lora receptive field represents the range of the same type of Lora in other layers that a given Lora in a layer can perceive. It can be controlled collaboratively by two hyperparameters: \textit{lora radius} and \textit{lora step length}, using $\epsilon$ and $\lambda$ to represent them respectively. 

\textbf{Interaction strategy}: In the cognitive hierarchy, there are channels for interaction between different levels\cite{level_channel}. Inspired by this, we introduce interaction matrices in the latent space of experts to realize the interaction between modules of different experts. We use $\mathcal{I}_a^b\in \mathbb{R}^{\xi \times \xi}$ to denote the interaction matrix between $a$ and $b$. The set of outputs of the features in the latent space after going through the interaction matrix and upscaling is:

\begin{equation}
    \mathcal{V}_l=\{x a \mathcal{I}_a^b b | a \in \{\mathcal{A}_0^l,...,\mathcal{A}_N^l\},b\in \mathcal{R}^l \cup \{\mathcal{B}_0^l\} \}
\end{equation}

In particular, $\mathcal{I}_a^b$ is a zero matrix if there is no interaction between $a$ and $b$, and a unit matrix if a vanilla interaction strategy is applied. $\mathcal{R}^l$ is a set controlled by the lora receptive field, denoted as:
\begin{equation}
    \mathcal{R}^l=\{\mathcal{B}^j_i |i \in \{1,...,N\},j\in \{l-\epsilon \lambda,... ,l,...,l+\epsilon \lambda\} \}
\end{equation}

\textbf{Routing strategy}: 
After obtaining $\mathcal{V}_l$, a routing strategy is needed to assign weights to the elements in the set. In this paper, we design a routing strategy based on gating network. The macro-expert is the main goal of this phase, so the result coming exclusively from it does not need to be multiplied by a special weight. The set of weights that need to be assigned by the gating network is $\mathcal{V}_l \setminus\{ x \mathcal{A}^l_0 \mathcal{I}_{\mathcal{A}^l_0}^{\mathcal{B}^l_0} \mathcal{B}^l_0 \}$, and we use $\mathcal{V}_l^*$ to represent it. Dimensionality reduction, concatenation, normalization, and flattening need to be performed sequentially before feeding the results into the gating network. Dimensionality reduction utilises a learnable matrix $\mathcal{A}_r^l \in \mathbb{R}^{d_2 \times \mu},\mu \ll d_2$ to reduce the dimensionality of each result, thus avoiding an excessive training burden. Normalization aims to enhance training stability\cite{lora_feature2}. The above operation process can be described as follows:

\begin{equation}
    g=f_{F}(f_{N}(\bigoplus_{v\in \mathcal{V}_l^*}v\mathcal{A}^l_r)), g \in \mathbb{R}^{\left|\mathcal{V}_l^*\right| \mu}
\end{equation}
where $f_{N}(.)$ and $f_{F}(.)$ correspond to the normalization and flattening operations, respectively. $g$ is fed into the gating network, and then the dimension is reduced to $\left|\mathcal{V}_l^*\right|$ by the learnable parameter $e \in \mathbb{R}^{\left|\mathcal{V}_l^*\right|\mu \times \left|\mathcal{V}_l^*\right|}$. 

\begin{equation}
    \Tilde{x}=g^{T}\cdot e, \Tilde{x} \in \mathbb{R}^{\left|\mathcal{V}_l^*\right|}
\end{equation}

The weight of the $i^{th}$ element in $\mathcal{V}_l^*$ is calculated by:
\begin{equation}
    \mathcal{G}(\Tilde{x}_i)=\frac{exp(\Tilde{x}_i/\tau)}{\sum_{j=1}^{\left|\mathcal{V}_l^*\right|}exp(\Tilde{x}_j/\tau)}
\end{equation}
where $\tau$ is a temperature scalar that can be learned\cite{lora_feature2}. Eventually, the output $x^{\prime}$ of the current block is:
\begin{equation}
    x^{\prime}=x\left[ \mathcal{W}_0^l+\mathcal{A}^l_0 \mathcal{I}_{\mathcal{A}^l_0}^{\mathcal{B}^l_0} \mathcal{B}^l_0 \right] +\sum_{v \in \mathcal{V}_l^*}\mathcal{G}(\Tilde{x}_i)v
\end{equation}

In the inference phase, the model can automatically capture how each expert collaborates based on the inputs and generate the output of the current step. Both during centralized macro learning and in the inference phase, micro-experts can play the role of verifier in process supervision\cite{process_supervision}, guiding training and providing interpretability.

\subsection{Training Object}
Under the proposed framework, the scene safety understanding task in this paper is decomposed into multiple subtasks and realized by multi-stage training. The quality of the training in each stage affects the final complete task, and therefore, training needs to be tailored to the task characteristics and data attributes.

\paragraph{Perception task} For the perception task in this paper, it is essentially the segmentation of a scene description paragraph into several distinct clauses. This can be easily achieved by vanilla SFT. However, there is no single correct segmentation result for this task, and the differences between various segmentation methods are qualitative. High-quality segmentation has a reasonably fine granularity, and each clause in the result usually contains complete semantic information, which can be helpful for the judgment task. In contrast, the semantic information in low-quality results is fragmented, and the judgment task has a high probability of arriving at \textit{neutral} results based on these clauses. Therefore, the essence of this task is human preference learning in a fixed format. To achieve this, we combine the respective advantages of SFT and DPO\cite{dpo} by adding the losses of both as the loss function for this task. It is calculated as follows:


\begin{equation}
\begin{aligned}
    \mathcal{L}_{P}^i\negthickspace=&\mathcal{L}\negthickspace-\negthickspace\mathbb{E}_{\mathcal{X}_i\sim \mathcal{D}}\left[log\sigma  (\beta log\frac{\pi_{\theta}(X_i^w|S_i)}{\pi_{ref}(X_i^w|S_i)}\negthickspace-\negthickspace\beta log\frac{\pi_{\theta}(X_i^l|S_i)}{\pi_{ref}(X_i^l|S_i)})\right]
\end{aligned}    
\end{equation}
where $\mathcal{X}_i=(S_i,X_i^w,X_i^l)$, $X_i^w$ and $X_i^l$ are the preferred and dispreferred responses, respectively, for $S_i$. $\sigma$ is the sigmoid function. $\beta$ is a hyperparameter. $\pi_{\theta}$ is the policy network we need to train. $\pi_{ref}$ is the reference network. $\mathcal{D}$ is the preference dataset. We construct $\mathcal{D}$ for this task based on the SSUPL, using the human annotations in the SSUPL as high-quality responses and the results of random segmentation or segmentation by nouns as low-quality responses.

\paragraph{Judging task}  The task conditions the scene description paragraph and a clause to give a judgment of the effect of the clause on the overall scene safety. However, with the constructed SSUPL dataset, one faces uneven distribution of process labels.
In order to avoid model overfitting to a class and model bias, we use an entropy-based regularization technique to design the loss function for this task\cite{attention_entropy}. The loss function for sample $i$ is:

\begin{equation}
    \mathcal{L}_{J}^i=\mathcal{L}-\sum_{l=1}^L \sum\limits_{\substack{ t \in x_{i}\\ n \in y_{i}}} \phi(y_i) \Tilde{A}_{l}^i(q_n, q_t) \log(\Tilde{A}_{l}^i(q_n, q_t) )
\end{equation}
where $\phi(y_i)$ is a weight function of the unity formula, equal to -1 when $y_j=\mathbb{P}_2$ and 1 otherwise. $\Tilde{A}_{l}^i=\psi(A_l^i)$, $\psi(.)$ is the row-wise softmax function. The training data for this task is obtained from the SSUPL.

\paragraph{Cognitive task} This task predicts the safety level of the entire scene conditional on the results of the previous two tasks. When humans measure the scene safety, they usually focus their attention mainly on the elements that have an impact on the overall safety. In the context of this paper, from an information theoretic perspective, entities/relationships with category \textit{positive} or \textit{negative} deserve to be more informative than those with category \textit{neutral}. In addition, we know from Section \ref{results_correlation} that the prediction of the final safety level is strongly correlated with each of the two attention distributions. Inspired by the prior knowledge, we design a loss function based on the entropy of attention distribution for this task\cite{attention_entropy}. Considering the ability of process labels to aggregate information from their corresponding clauses, and computational efficiency, the loss function is designed here based on the prior from process labels as follows:

\begin{equation}
    \mathcal{L}_{C}^i=\mathcal{L}-\sum_{l=1}^L\sum_{j=1}^{k_i} \sum_{t \in y_{i,j}} \phi(y_{i,j}) \Tilde{A}_{l}^i(q_{\hat{r_i}},q_t)\log(\Tilde{A}_{l}^i(q_{\hat{r_i}},q_t) )
\end{equation}

The loss function guides the model to adjust the attention distribution during training to achieve the largest possible entropy of the attention distribution for the tokens with the process labeled \textit{neutral}, and the smallest possible entropy of the attention distribution for the text with the process labeled \textit{positive} or \textit{negative}. The training data for this task are also obtained from the SSUPL.

\paragraph{Centralized macro learning} This phase supervised fine-tuning of the model with the complete scene safety understanding data, and the loss function takes the vanilla cross-entropy loss.

\subsection{Performance Evaluation}
\subsubsection{Comparison with State-of-the-art Methods}
\begin{table*}
\caption{Comparison of performance under different methods.}
\centering
\begin{tabular}{c|ccccc}
\hline
\toprule
\multirow{2}{*}{\textbf{Method}}& \multicolumn{5}{c}{\textbf{Base Model}(Accuracy\% | Maximum Trainable
Param.\%)} \\
&GPT-J&Llama2-7b-chat&Baichuan2-7b-chat &Gemma2-9b-it&Intern2.5-7b-chat\\
\midrule
ICL\citeyear{icl} & 12.78 $\vert- -$ &33.90 $\vert- -$ &33.05 $\vert- -$&38.16 $\vert- -$&41.84 $\vert- -$\\
CoT\citeyear{Cot} & 5.32 $\vert- -$&35.18 $\vert- -$ &30.64 $\vert- -$&43.12 $\vert- -$&36.74 $\vert- -$\\
LoRA-Final\citeyear{lora} & 68.79 $\vert$ 0.0606  &66.52 $\vert$ 0.0622 & 72.34 $\vert$ 0.2378 &72.62 $\vert$ 0.0484  &72.91 $\vert$ 0.2433\\
LoRA\citeyear{lora} & 65.11 $\vert$ 0.0606  &68.23 $\vert$ 0.0622 & 69.79 $\vert$ 0.2378 &72.48 $\vert$ 0.0484  &73.19 $\vert$ 0.2433\\
DoRA\citeyear{dora} & 64.37 $\vert$ 0.0644  &65.53 $\vert$ 0.0661 & 67.52 $\vert$ 0.2558  &72.65 $\vert$ 0.0512  &71.63 $\vert$ 0.2610 \\
MoSLoRA\citeyear{moslora} & 63.40 $\vert$ 0.0607  &70.21 $\vert$ 0.0623 & 67.94 $\vert$ 0.2378 &73.90 $\vert$ 0.0484  &72.00 $\vert$ 0.2435\\
LoRAMoE\citeyear{loramoe} & 61.56 $\vert$ 0.2571 & 63.40 $\vert$ 0.2638  & 67.52 $\vert$ 0.9903  &74.18 $\vert$ 0.2062 & 72.48 $\vert$ 1.0161\\
LoRA-Flow\citeyear{lora_flow} & 55.32 $\vert$ 0.0606 &61.42 $\vert$ 0.0622 & 66.95 $\vert$ 0.2378&70.21 $\vert$ 0.0484&70.35 $\vert$ 0.2433\\
LoRA-Hub\citeyear{lora_hub} & 49.15 $\vert$ 0.0606 &52.34 $\vert$ 0.0622& 67.37 $\vert$ 0.2378&70.76 $\vert$ 0.0484&69.50 $\vert$ 0.2433\\
LoRA-Fusion\citeyear{lora_fusion} & 51.35 $\vert$ 0.0606&49.79 $\vert$ 0.0622& 57.16 $\vert$ 0.2378&66.10 $\vert$ 0.0484&59.29 $\vert$ 0.2433\\
SFT+ORM\citeyear{process_supervision} & 65.82 $\vert$ 0.0606 & 69.50 $\vert$ 0.0622& 71.49 $\vert$ 0.2378&74.75 $\vert$ 0.0484&73.90 $\vert$ 0.2433\\
SFT+PRM\citeyear{process_supervision} & 60.28 $\vert$ 0.0606 &68.23 $\vert$ 0.0622& 70.07 $\vert$ 0.2378&72.62 $\vert$ 0.0484 &72.77  $\vert$ 0.2433\\

Ours & \textbf{70.92} $\vert$  0.0722 &\textbf{72.49} $\vert$  0.0740  & \textbf{73.76} $\vert$  0.2604  &\textbf{76.74} $\vert$  0.0598 &\textbf{74.61} $\vert$ 0.2651\\
\bottomrule
\end{tabular}
\label{final_comparative_results}
\end{table*}
To evaluate the performance of the proposed process-supervised training method, we selected two representative approaches for comparison: one that does not require fine-tuning and another that involves parameter fine-tuning.

 The methods that do not require finetuning include: (1) One-shot class vanilla In-Context Learning (ICL)\cite{icl}. For each safety level, a random sample is selected from the training data, concatenated with the corresponding safety level label, and subsequently incorporated into the prompt for the test samples. (2) The traditional chain-of-thought approach (CoT)\cite{Cot}. The phrase ‘\textit{Let's think step by step}’ is introduced into the prompt to assist the LLMs in reasoning about the safety level of the scene.  
        
Methods that require fine-tuning of parameters include: (1) LoRA-Final, which fine-tunes the base LLMs with data whose answers contain only the final safety level of the scene, i.e., ignoring intermediate processes in the SSUPL data. (2) LoRA, fine-tuning the base LLMs with the SSUPL dataset. (3) DoRA\cite{dora}, which mimics the learning ability of full fine-tuning, decomposes the pre-training weights into magnitude and direction components for fine-tuning; 4) MoSLoRA\cite{moslora}, which decomposes Lora into MoE-style subspaces and learns a mixer to fully blend the subspaces; (5) LoRAMoE\cite{loramoe}, which adds the most typical Moe-style LoRAs to the LLM and fine-tunes all LoRAs and a router simultaneously; (6) LoRA-Flow\cite{lora_flow}, which uses a fusion gate to dynamically adjust the impact of different LoRAs at each step of prediction; (7) LoRA-Hub\cite{lora_hub}, similar to LoRA-Flow, but only learns task-level fusion weights; (8) LoRA-Fusion\cite{lora_fusion}, mixing multiple related LoRAs by fixing constant weights; (9) SFT+ORM\cite{process_supervision}, which combines vanilla SFT model and outcome-supervised reward model for collaborative inference; (10) LoRA+PRM\cite{process_supervision}, which comes vanilla SFT model and process-supervised reward model for collaborative inference. Please refer to Appendix \ref{reward_methods_detail} for training details of the reward models.

\paragraph{Experimental settings} Similar to the dataset construction pipeline, we employ the captioning anything model\cite{caption_anything} within the framework to generate image description text. Accuracy is employed as the performance assessment metric. The rank of the LoRA involved in all methods is 8. In our approach, the parameter-frozen unit matrix is used as the interaction matrix within the interaction strategy. The descending matrix in the routing strategy is initialized using the Kaiming uniform distribution\cite{kaiming_initialize}. The values of $\epsilon$ and $\lambda$ for the LoRA receptive field are set to 1 and 3, respectively. The remaining training parameters are maintained as previously configured.

\paragraph{Results analysis} 
Table \ref{final_comparative_results} presents the experimental results across several approaches. From the results, we first observe that the two methods requiring no fine-tuning (ICL and CoT) perform the worst across all models, with accuracies generally below 50\%. Secondly, compared with the LoRA-Final approach, the vanilla LoRA method indeed exhibits superior performance on LLama2-7b-chat (68.23\% vs. 66.52\%) and Intern2.5-7b-chat (73.19\% vs. 72.91\%), while it slightly underperforms on other SFT LLMs such as GPT-J (65.11\% vs. 68.79\%), Baichuan2-7b-chat (69.79\% vs. 72.34\%), and Gemma2-9b-it (72.48\% vs. 72.62\%). This performance discrepancy is tentatively attributed to exposure bias \cite{exposure_bias}, a phenomenon that will be discussed in greater detail in the subsequent section. Notably, some methods in the table (e.g., LoRAMoE, DoRA) increase the number of trainable parameters yet fail to deliver corresponding performance improvements, often performing inferior to LoRA-Final on most models. These systematic observations strongly indicate that the target task in this study is inherently challenging, and that merely increasing the parameter count is insufficient.

The experimental results delineated in Table \ref{final_comparative_results} demonstrate the superior performance of our proposed method across all evaluated approaches. Our method consistently outperforms alternative fine-tuning strategies. Specifically, when compared against the second-best methods for each LLM, i.e., LoRA-Final for GPT-J and Baichuan2-7b-chat, MoSLoRA for Llama2-7b-chat, and SFT+ORM for both Gemma2-9b-it and Intern2.5-7b-chat, our framework exhibits performance improvements of 3.10\%, 1.96\%, 3.25\%, 2.66\%, and 0.96\%, respectively. This consistent pattern of enhancement across diverse model architectures provides substantial empirical evidence for the generalizability and efficacy of our approach in addressing complex cognitive tasks.

Furthermore, both our method and SFT+PRM utilize process supervision. The key distinction lies in the latter's construction of step-level preference labels (i.e., $(x_i,\mathbb{P}_1)\succ (x_i,\mathbb{P}_2)\succ (x_i,\mathbb{P}_3)$) to evaluate process labels for intermediate inference and train reward models. However, process labels inherently possess two-sided characteristics. For instance, a fire hydrant in a scene may enhance safety by enabling firefighting during a fire, yet its presence also signifies a potential fire hazard, indirectly compromising safety. Consequently, the method of constructing process label preferences, as outlined in Appendix \ref{reward_methods_detail}, inevitably introduces some noise, which results in the performance of SFT+PRM being inferior to both our method and SFT+ORM.

Regarding training parameters, LoRAMoE requires the simultaneous training of multiple experts, resulting in the highest number of training parameters. Due to the introduction of trainable routing strategies, our framework incorporates slightly more maximum training parameters than the other methods.

\subsubsection{Ablation Experiment}
\begin{figure}
    \centering
    \subfloat[]{\includegraphics[width=2.5in]{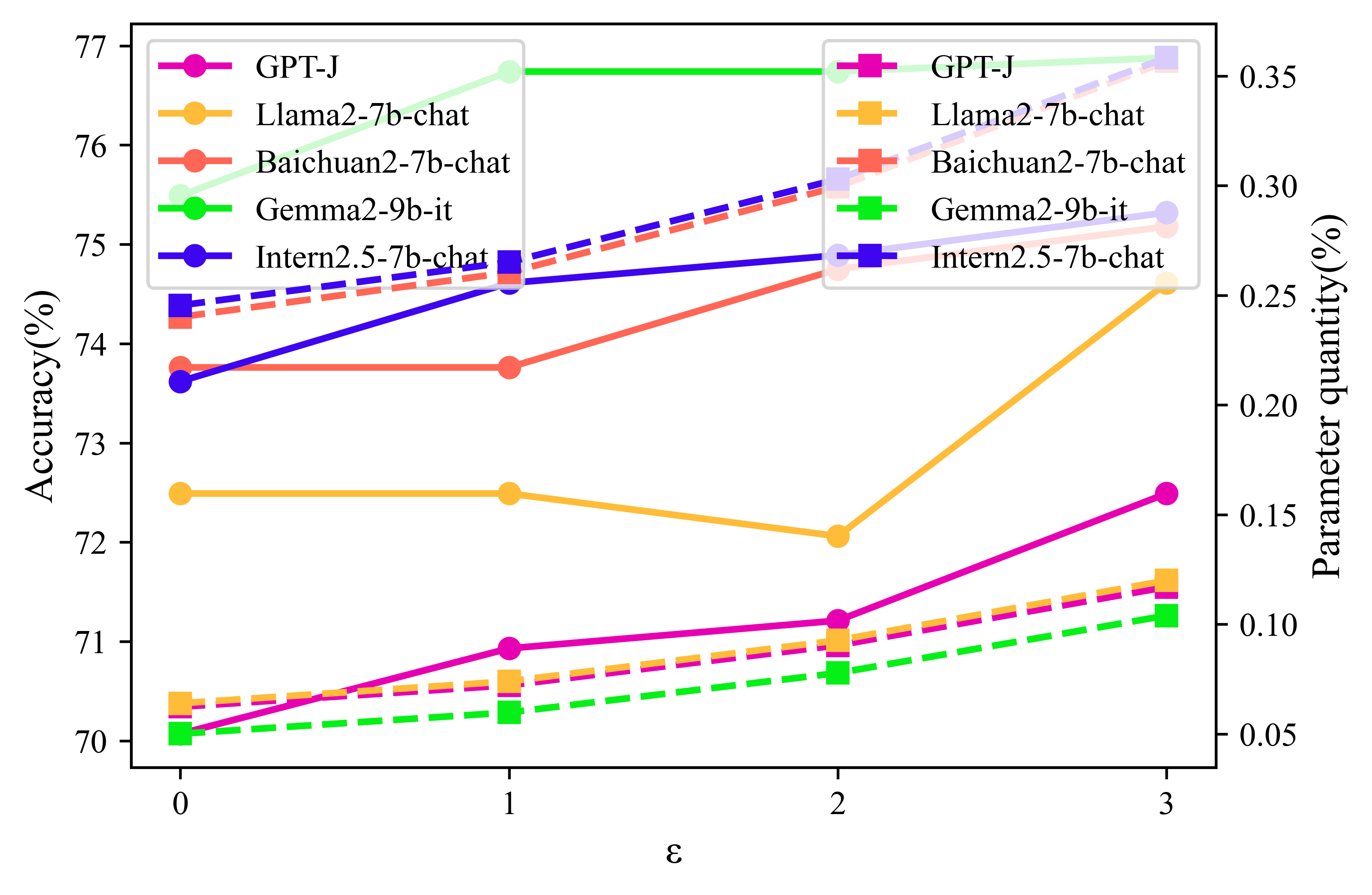}%
    \label{different_epsilon}}
    \hfill
    \subfloat[]{\includegraphics[width=2.5in]{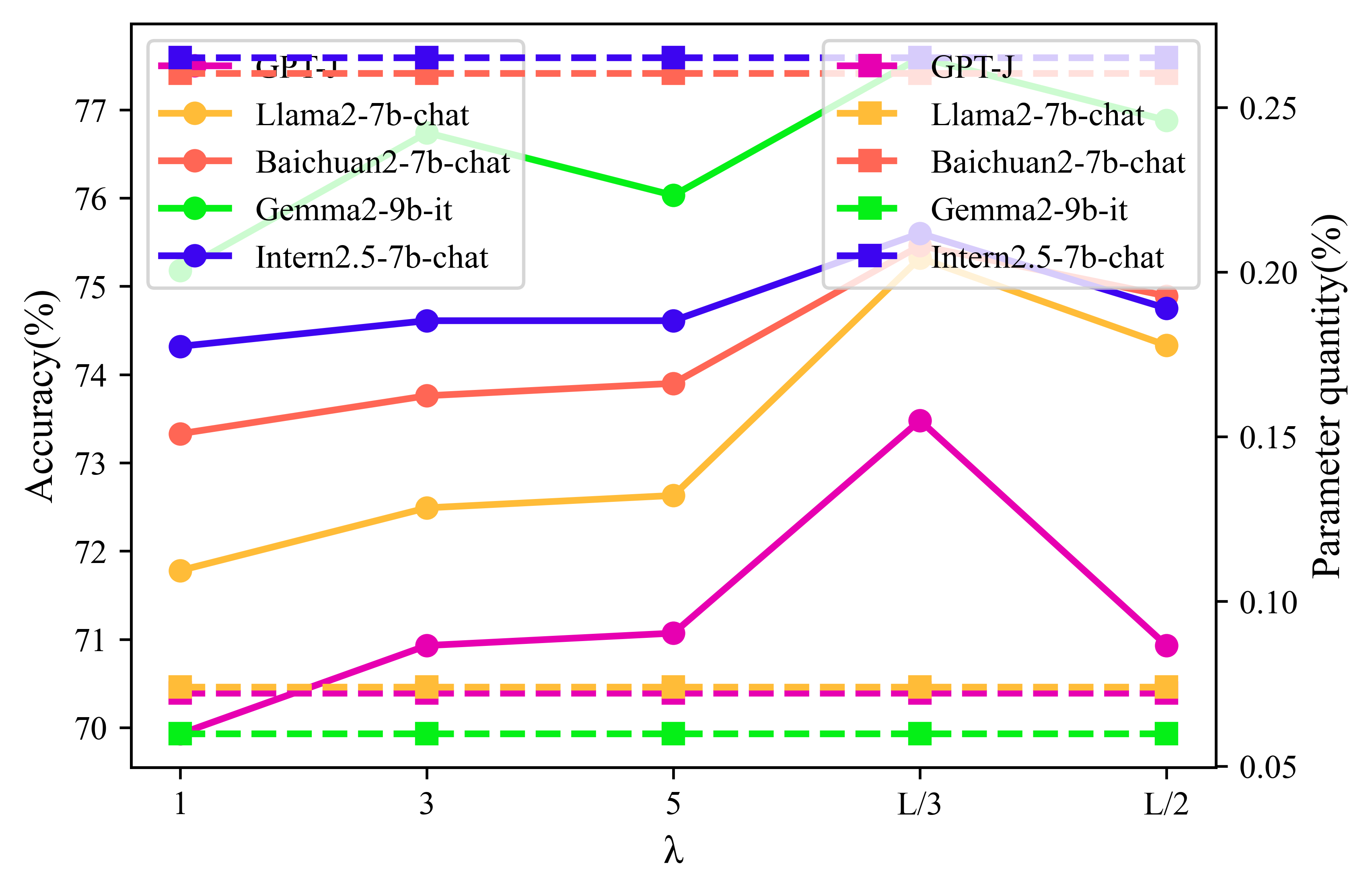}%
    \label{different_lambda}}
    \caption{Variation of performance with different parameters $\epsilon$ and $\lambda$. (a) Performance varies with $\epsilon$. (b) Performance varies with $\lambda$.}
    \label{lora_receiptive_field_comparsion}
\end{figure}

\paragraph{Effectiveness of lora receiptive fields}
Due to the flexibility and inclusiveness of this framework, we also compare the results under different detail settings. Specifically, we compare the performance of the LLMs in different settings $\epsilon$ and $\lambda$, respectively, and the results are shown in Figure \ref{lora_receiptive_field_comparsion}. It can be seen that changes in both parameters can lead to changes in performance, while the former will also lead to changes in the training parameters. Specifically, for a fixed $\lambda=3$, the performance and training parameters show an increasing trend as $\epsilon$ becomes larger. This is due to the gradual introduction of more experts at each layer, which increases the number of parameters that need to be trained for the routing strategy while absorbing more diverse knowledge. While in the case of fixing $\epsilon=1$ and increasing $\lambda$, the training parameters of the LLMs remain constant, while the performance shows a non-monotonic trend, with the optimal performance obtained at $\lambda=round(\frac{L}{3})$. Using the index difference of the layers as a measure of the difference between the knowledge of the different layers, then the discrepancy function obtained for the $\epsilon=1$ setting is $diff(\lambda)=-12\lambda^2+8L\lambda$. $\lambda=round(\frac{L}{3})$ allows for the widest range of LoRA receptive field within the LLMs for each layer on average, and thus allows for the uptake of the most differentiated knowledge. Furthermore, whereas on the left and right sides of $\lambda=round(\frac{L}{3})$, performance roughly tends to increase and decrease, respectively, as $L$ increases, highly consistent with the monotonicity of the discrepancy function. After $\lambda$ is greater than $\frac{L}{3}$, the relationship between $\lambda$ and performance enters the next cycle.

\begin{figure*}
    \centering
    \includegraphics[width=0.9\linewidth]{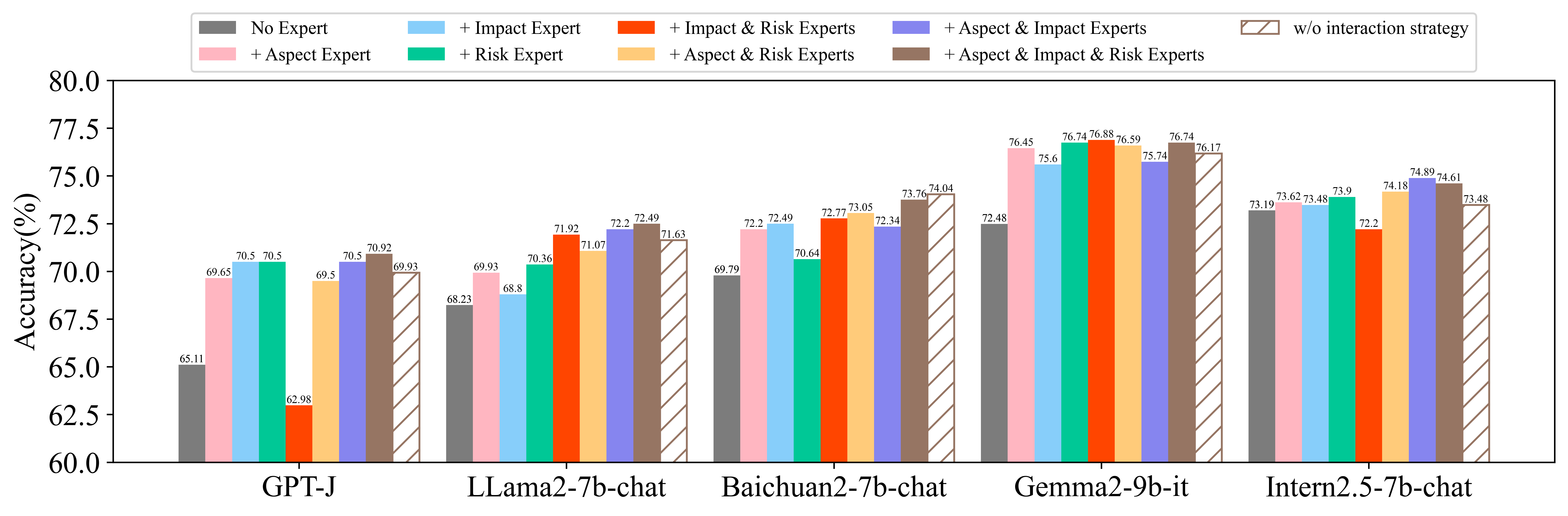}
    \caption{Ablation study}
    \label{expert_enhancement}
\end{figure*}

\paragraph{Effectiveness of each experts}
To evaluate the effectiveness of each expert, we gradually introduce them in the distributed micro training phase, and the results are shown in Figure \ref{expert_enhancement}. As can be seen from the figure, the introduction of additional micro-experts is likely to be beneficial to the model performance. Different kinds of experts have different gain effects in the same model, and the same kind of expert has different enhancement effects on different models. With the exception of Intern2.5-7b-chat and Gemma2-9b-it, the best performance is obtained by introducing experts for the three subtasks in the rest of the LLMs. In Intern2.5-7b-chat and Gemma2-9b-it, the introduction of all experts achieves the second best performance, only slightly worse than the settings without the Risk Expert and the Aspect Expert, respectively.

\paragraph{Effectiveness of interaction strategy}
The performance of the LLMs without the interaction strategy is also shown in Figure \ref{expert_enhancement}, as indicated by the columns filled with diagonal lines. It can be found that, except for Baichuan2-7b-chat, the interaction matrix between different experts is able to increase the performance of the LLMs, which indicates that the interaction between different experts learns useful knowledge.

\subsubsection{The Role of Process Labels in Inference}
Here, we explore the reasons for the performance degradation of the LoRA method based on SSUPL compared to the LoRA-Final method in the previous section, which reflects the specific role of process labels in the completion of the target task. Through Section \ref{info} and Section \ref{correlation_analysis}, we have learnt that process labels plays an important role in the prediction of final safety. Here we go a step further and roughly count the accuracy of the first process label and the last process label in the inference results of SFT LLMs on the test set, as shown in Table \ref{process_accuracy}. The accuracy of process label prediction is generally between 80\% and 90\%. 

Next, we control the process labels to explore the signals provided by them for predicting the final safety level in two components\cite{reason_step_impact}: (i) step-process label mapping, i.e., whether $x_{i,j}$ is mapped to the correct $y_{i,j}$; (ii) process label space, i.e., $\bigcup_{j=1}^{k_i} y_{i,j}$. We take a single-step-only inference approach based on manual process label flipping\cite{label_flip}, i.e., inputting $W_i\oplus S_i \oplus T(R_i) \oplus F_i$ into the SFT LLMs instead of $W_i\oplus S_i$. Of these, $R_i$ comes from human annotation in the data annotation stage, rather than model prediction. We manually flip the $y_{i,j}$ in the $R_i$.

\paragraph{Experimental settings}
We consider 5 types of process label flipping: (1) Ground truth, i.e., fully correct step-process label mapping; (2) Positive-negative flipping (PN flipping), i.e., changing negative to positive and positive to negative based on Ground truth, and in this way get some step-process label mappings with errors; (3) All negative, i.e., set all process labels to negative. (4) All positive, i.e., set all process labels to positive. (5)All neutral, i.e., set all process labels to neutral. Note that all reasoning is performed on the test set of SSUPL. Moreover, we use the normal case, i.e., inputting $W_i\oplus S_i$, as the baseline.

\begin{table}
\caption{Average prediction accuracy of the first process label and the last process label}
\centering
\begin{tabular}{ccc}
\hline
\multirow{2}{*}{\textbf{Model}} & \multicolumn{2}{c}{\textbf{Accuracy\%)}} \\
&First step&Last step\\
\hline
GPT-J& 79.01 & 81.13 \\
Llama2-7b-chat & 83.12 &  85.82\\
Baichuan2-7b-chat & 78.58& 87.23  \\
Gemma2-9b-it & 84.54 & 86.38 \\
Intern2.5-7b-chat & 80.71&  86.38 \\
\hline
\end{tabular}
\label{process_accuracy}
\end{table}

\paragraph{Results analysis}

\begin{table*}
\caption{Performance changes caused by manually flipping process labels}
\centering
\begin{tabular}{c|ccccc}
\hline
\toprule
\multirow{1}{*}{\textbf{Process Label}}& \multicolumn{5}{c}{\textbf{SFT Model}(Accuracy\%)} \\
\textbf{Setting}&GPT-J&Llama2-7b-chat&Baichuan2-7b-chat &Gemma2-9b-it&Intern2.5-7b-chat\\
\midrule
PN flipping &59.72  & 65.53   &68.22  &66.38 & 65.82 \\
All negative  &49.36  & 55.18   &56.45  &60.00 & 56.31  \\
All positive &52.77  & 58.44  &64.96  & 63.12 & 51.63  \\
All neutral &20.28  & 43.69   &50.78   &52.06 & 42.13 \\
Normal case  &65.11  & 68.23   &69.79   &72.48 & 73.19  \\
Ground-truth   &\textbf{71.91}  & \textbf{72.62}   &\textbf{76.31}   &\textbf{75.74} & \textbf{77.30}  \\
\bottomrule
\end{tabular}
\label{label_flip_results}
\end{table*}

\begin{figure*}
    \centering
    \subfloat[]{\includegraphics[width=1.75in]{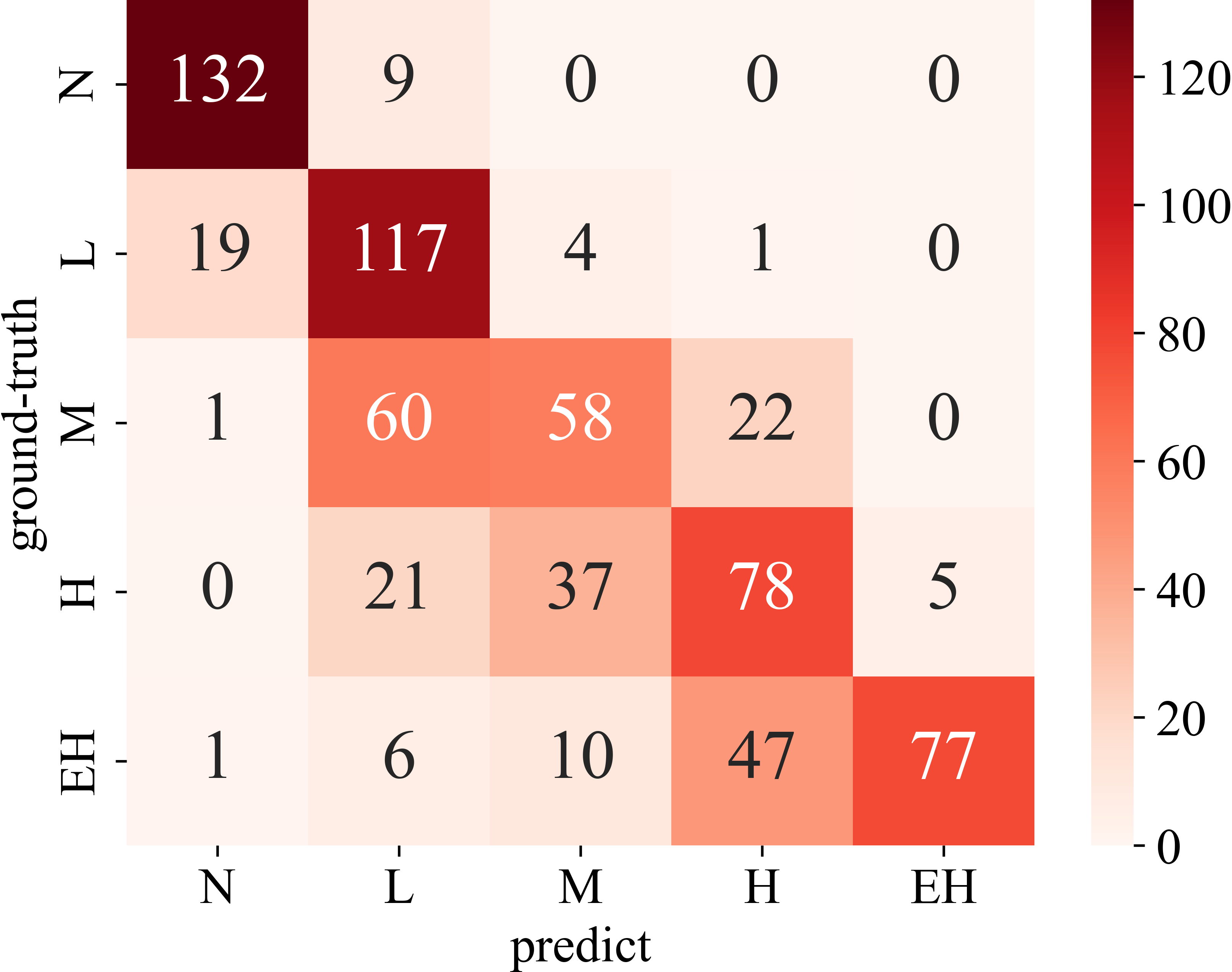}%
    \label{PN flipping}}
    \hfil
    \subfloat[]{\includegraphics[width=1.75in]{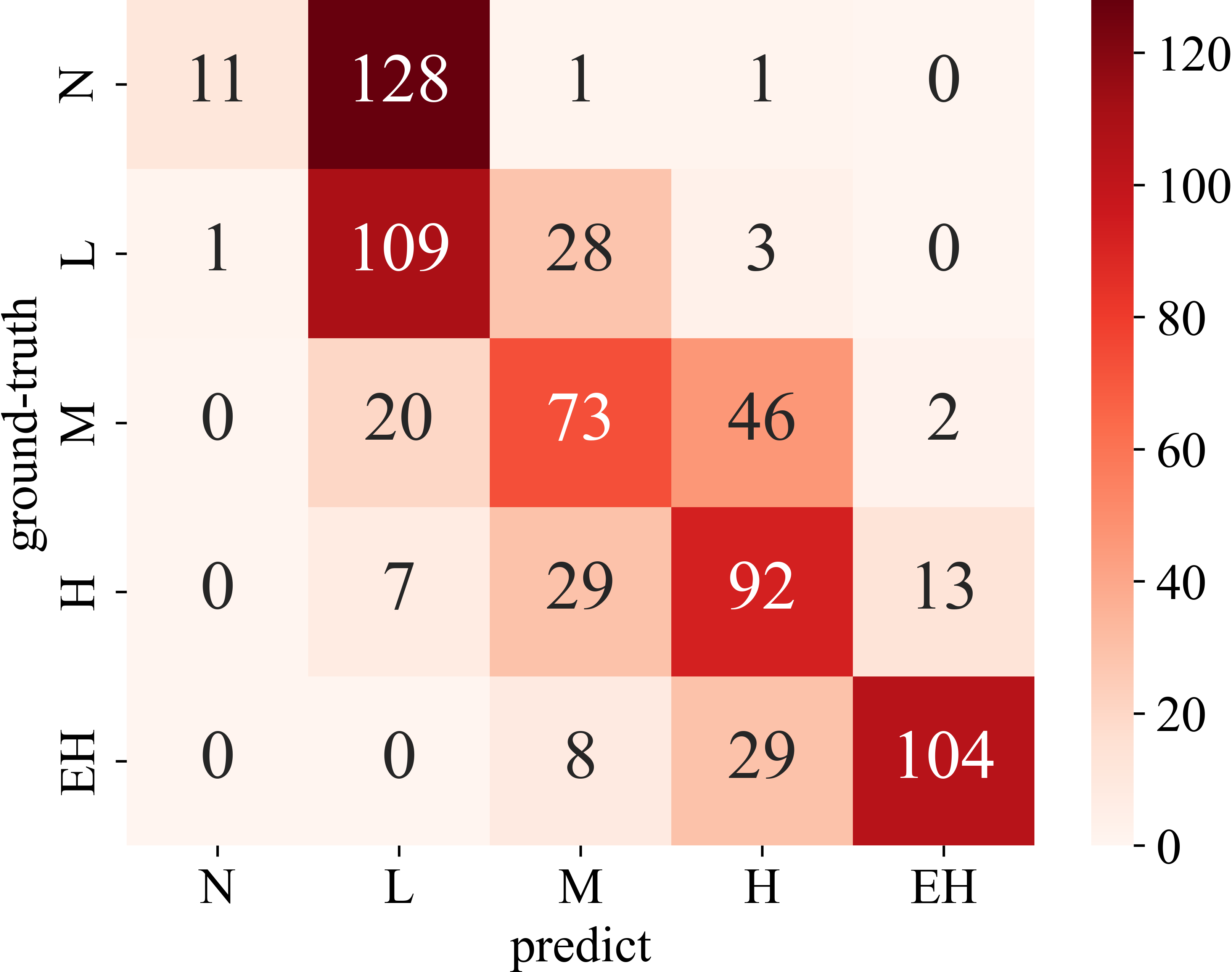}%
    \label{All negative}}
        \hfil
    \subfloat[]{\includegraphics[width=1.75in]{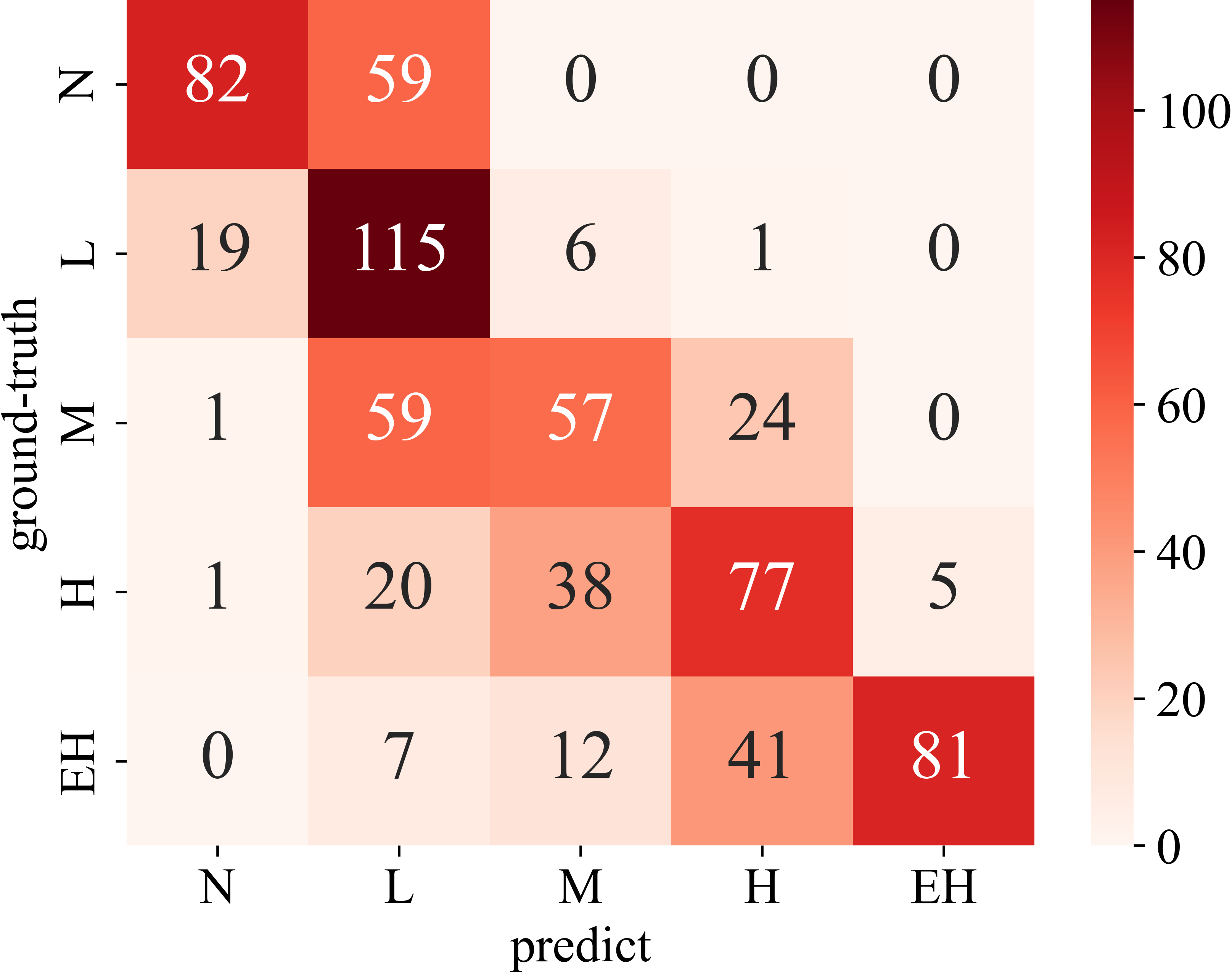}%
    \label{All positive}}
        \hfil
    \subfloat[]{\includegraphics[width=1.75in]{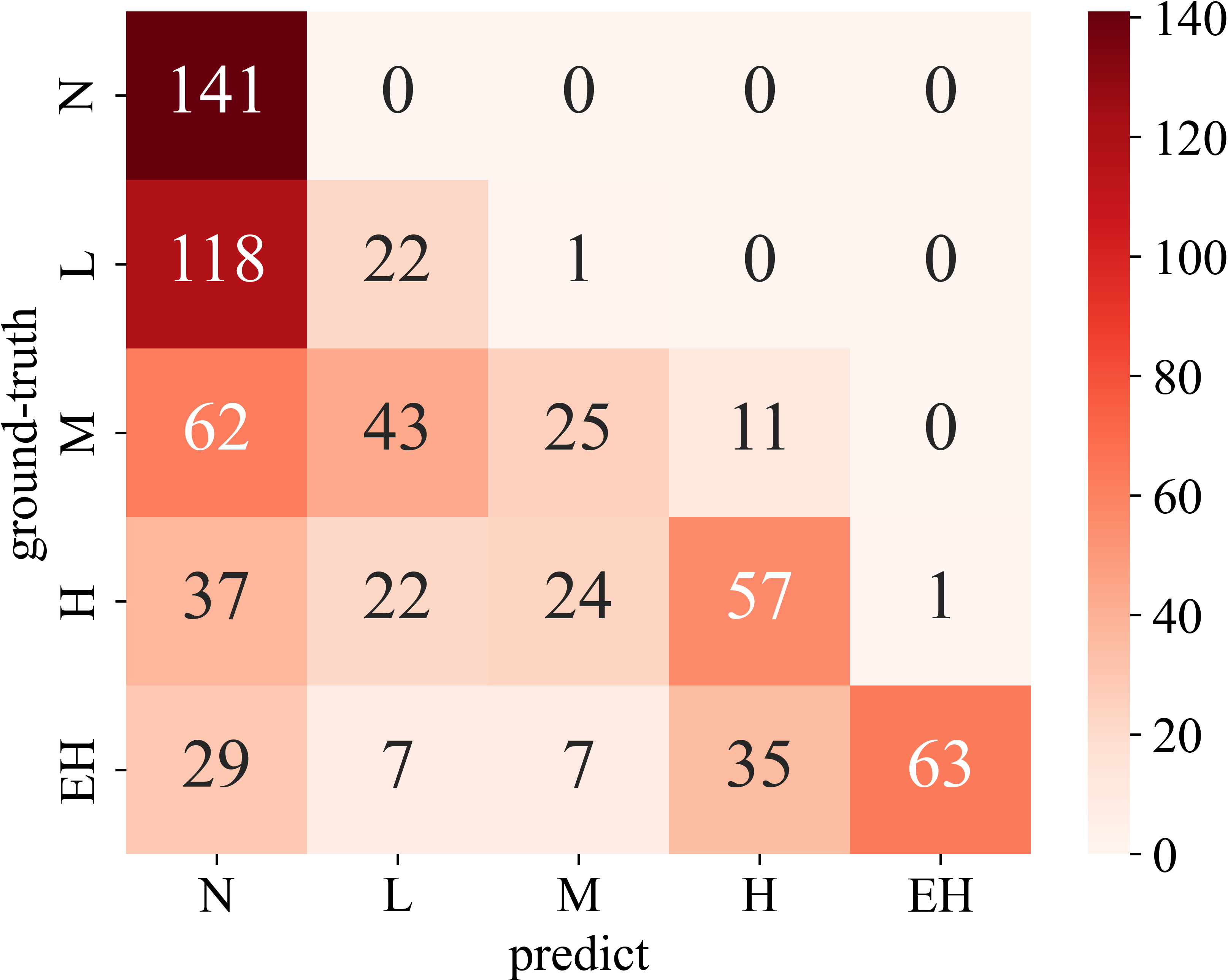}%
    \label{All neutral}}
        \hfil
    \subfloat[]{\includegraphics[width=1.75in]{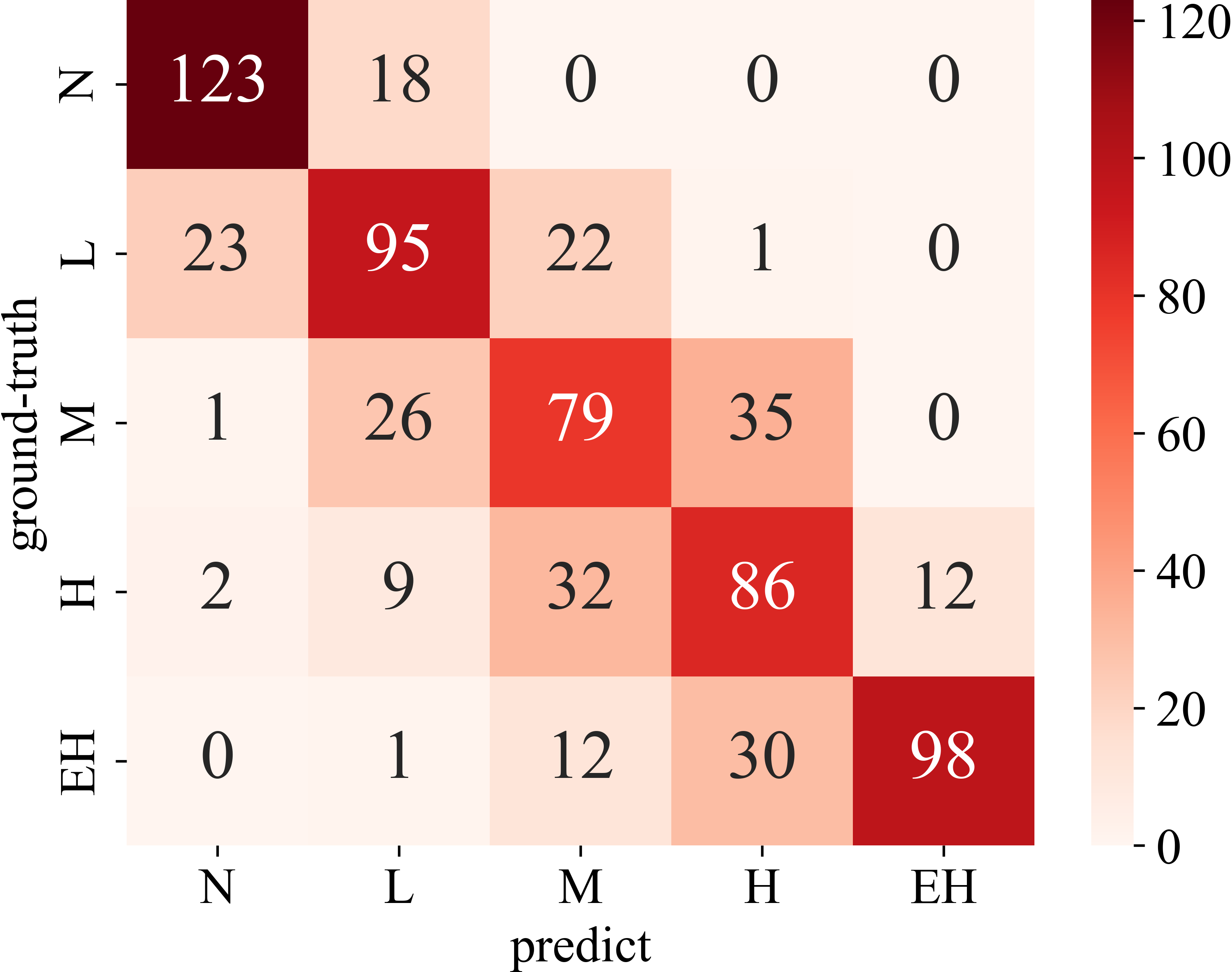}%
    \label{Normal case}}
        \hfil
    \subfloat[]{\includegraphics[width=1.75in]{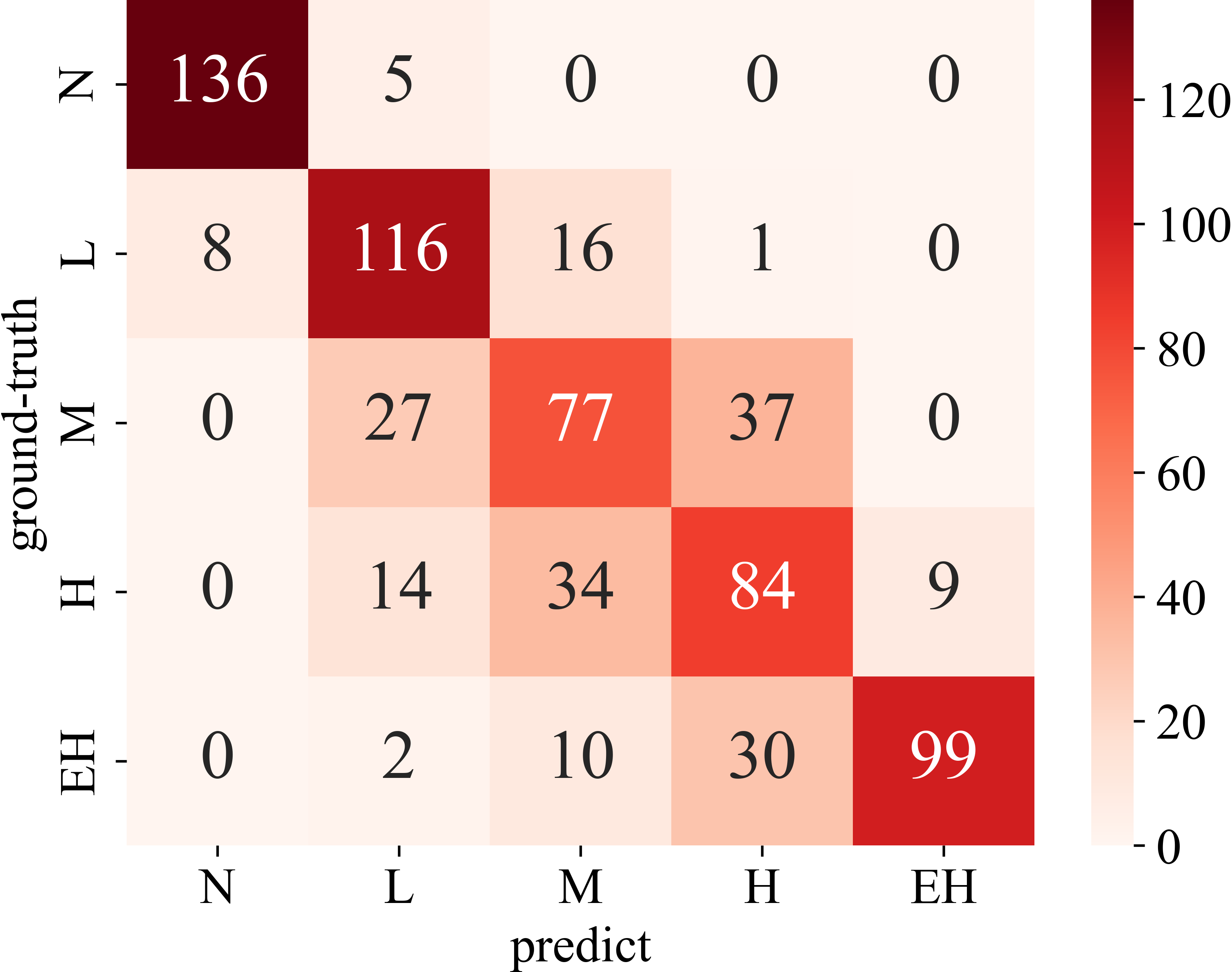}%
    \label{Ground truth}}
    \caption{Confusion matrix for the LLama2-7b-chat-sft with different process labels settings. 'N'='No risk', 'L'='Low risk', 'M'='Medium risk', 'H'='High risk' and 'EH'='Extremely high risk' in these images. (a) PN flipping. (b) All negative. (c) All positive. (d) All neutral. (e) Normal case. (f)Ground truth.}
    \label{confusion_matrix}
\end{figure*}

Table \ref{label_flip_results} shows the results obtained by flipping the process labels under different SFT LLMs. It first shows that the SFT LLMs with totally correct step-process label mapping has the best performance over the normal case where there is some misclassification of process labels. This suggests that improving the correctness of the process labels is beneficial for the prediction of the final safety level. Secondly, it can be observed that positive-negative flipping does not get an accuracy that would intuitively be low. We believe that there are three reasons for this. Firstly, the impact of certain entities or relationships on scene safety is two-sided, resulting in some labels still being correct after flipping. Second, there is a high percentage of \textit{neutral} in the reasoning process, making most of the process labels still correct. Finally, reasoning in this format possesses robustness, and the final safety level prediction does not depend entirely on the process labels, but also on the \textit{normal text part}, i.e., $x_i,j$, as described in Section \ref{correlation_analysis}. In addition, the performance of all SFT LLMs is degraded when only one type of process label is present in the process label space and is particularly pronounced in the \textit{all neutral} setting, suggesting that \textit{neutral} provides a small amount of useful information compared to the other two types of process labels. By looking at the confusion matrix shown in Figure \ref{confusion_matrix}, we further find that the SFT LLMs significantly underestimate the safety level of the scenes under the \textit{all neutral} setting, which indicates that the SFT LLMs are trying its best to align the prediction results with our view that ‘\textit{All neutral is safe}’. In addition, under the \textit{all negative} setting, the SFT LLMs hardly predict the safety level to be \textit{no risk}, which is highly consistent with our perception of no risk scenes.

Furthermore, the performance of SFT LLMs with fully correct step-process label mapping is clearly superior to the LoRA-Final method. This suggests that fully aligned inference process produces a negative alignment tax\cite{process_supervision}. However, in the normal case, the inference process is not fully aligned with human due to exposure bias, and errors gradually accumulate, leading to weaker performance than the Lora-Final method in the real case. On the other hand, the LoRA-Final method outputs only a few tokens representing the final safety level, and the probability of exposure bias generation is negligible.

\subsubsection{Interpretable Reasoning Process}
\begin{figure*}
	\centering
	\includegraphics[width=0.9\linewidth]{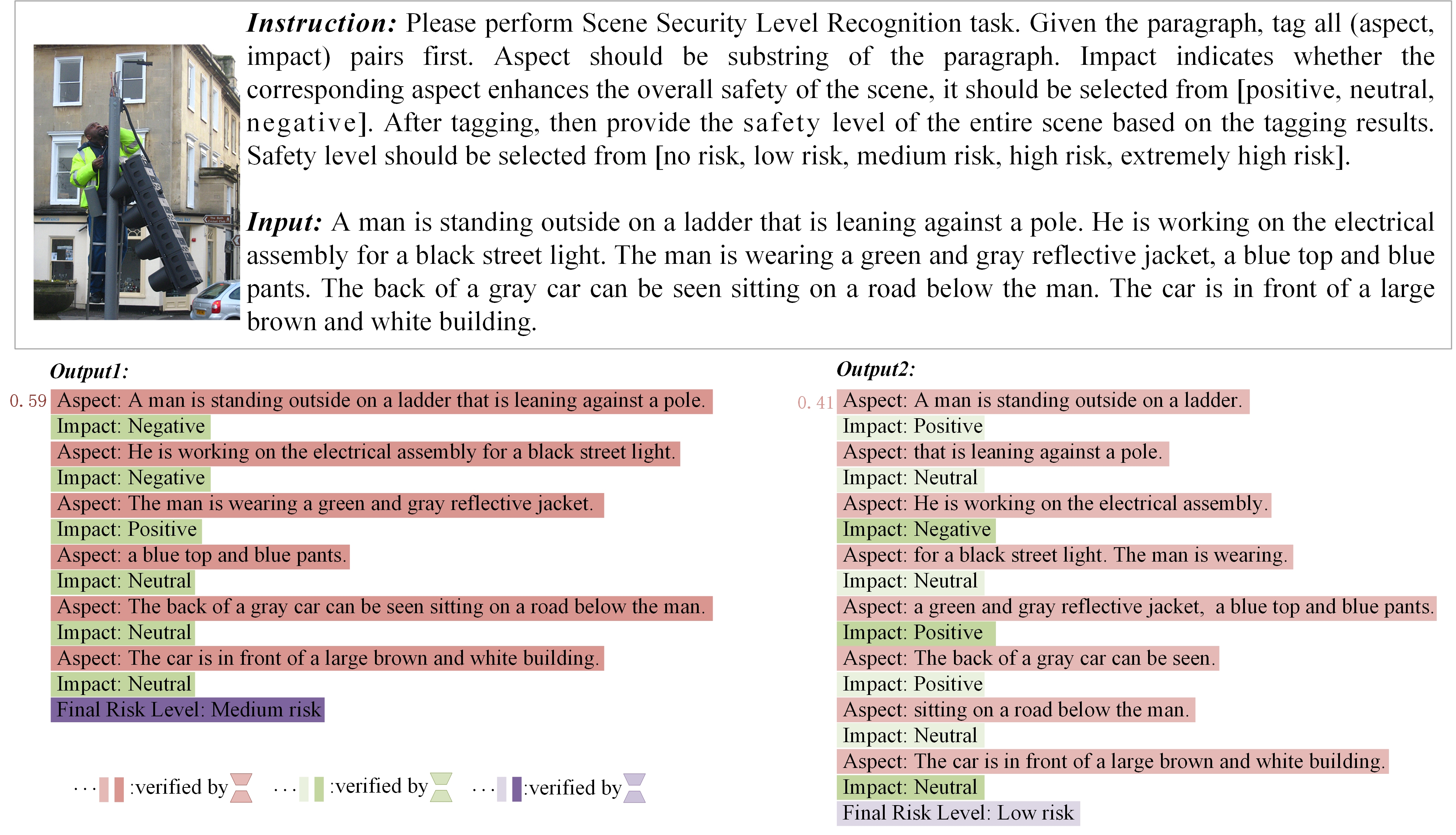}
	\caption{Two reasoning processes for the same scene.}
	\label{interpretable_reason_process}
\end{figure*}

Similar to PRM\cite{process_supervision}, this framework also provides interpretable reasoning processes. Figure \ref{interpretable_reason_process} illustrates an example under the Gemma2-9b-it model, where the output on the left is the correct output and the right is the incorrect output. The LoRAs obtained from distributed micro training phase act as verifiers to assess the quality or correctness of the reasoning process. Different LoRAs verify the results at different cognitive levels. Different colored steps are verified by different verifiers, the darker the color, the higher the quality. The micro-expert for the perception task normalizes the rewards of the two reasoned responses to obtain a validation score, e.g., the perceptual result on the left of the figure has a score of 0.59, while the one on the right has a score of 0.41. The micro-experts for the remaining two tasks judge the correctness of the corresponding results in terms of whether or not the results have the highest level of confidence. It can also be seen that all the steps in the output on the left are dark in color, while there are some steps in the output on the right that are light in color, which means that the verifiers have successfully identified the low quality or errors in the incorrect output.

\section{Conclusion}
In this work, we explored the possibility of integrating cognitive process modelling with process supervision by introducing, in turn, a hierarchical safety cognitive structure, the SSUPL dataset, and a process-supervised training framework that reflects the cognitive hierarchy. The training framework uses LLM as the reasoning backbone model and implements modular cognitive hierarchy functions and collaboration between different levels based on MoE and LoRA. The framework is flexible and scalable and outperforms various other types of approaches. The lora receptive field, interaction strategy, and routing strategy within this framework are worthy of deeper exploration in the future. This exploration represents a significant step towards bridging the gap between artificial and human intelligence in the domain of complex scene comprehension and safety assessment. We hope that the framework can be extended to other areas.

\clearpage
\newpage
\appendix

\section{Results for Other SFT LLMs}
\subsection{Results of $\mathcal{M}^{yr}$, $\mathcal{M}^{xr}$, $\mathcal{M}^{sr}$ and $\mathcal{M}^{xx}$}
\begin{figure*}
    \centering
    \subfloat[]{\includegraphics[width=1.5in]{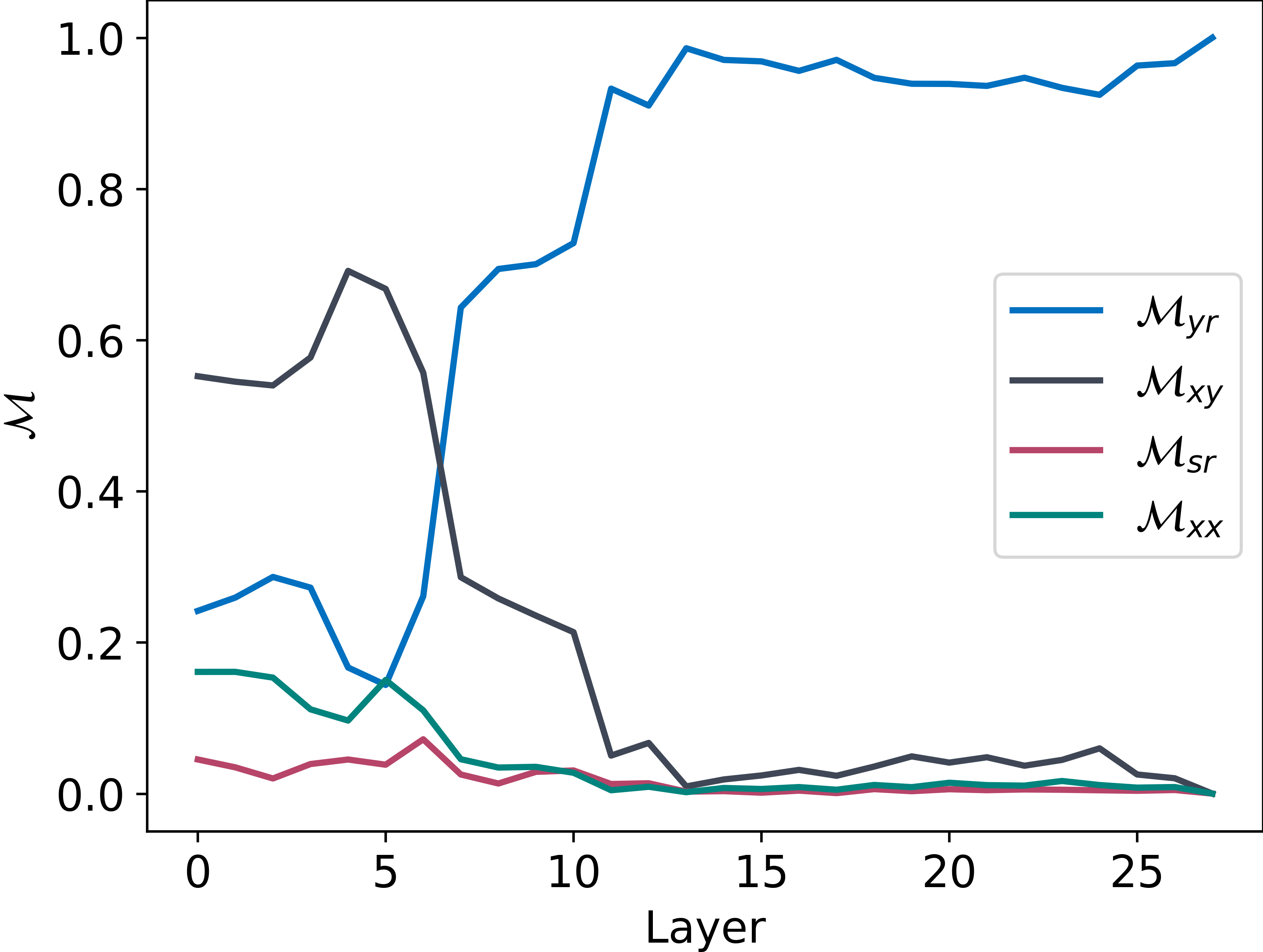}%
    \label{spp2}}
    \hfil
    \subfloat[]{\includegraphics[width=1.5in]{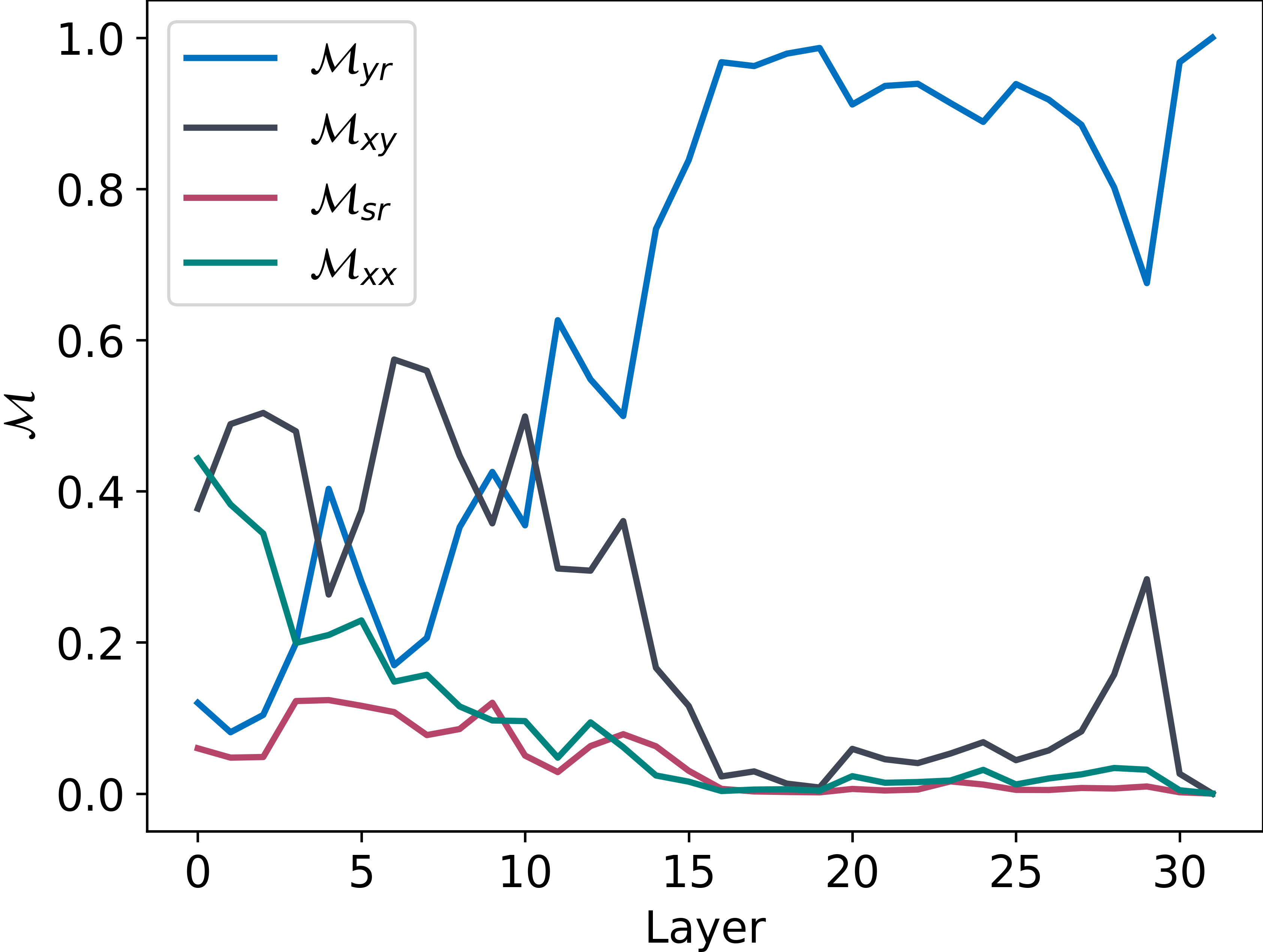}%
    \label{spp4}}
        \hfil
        \subfloat[]{\includegraphics[width=1.5in]{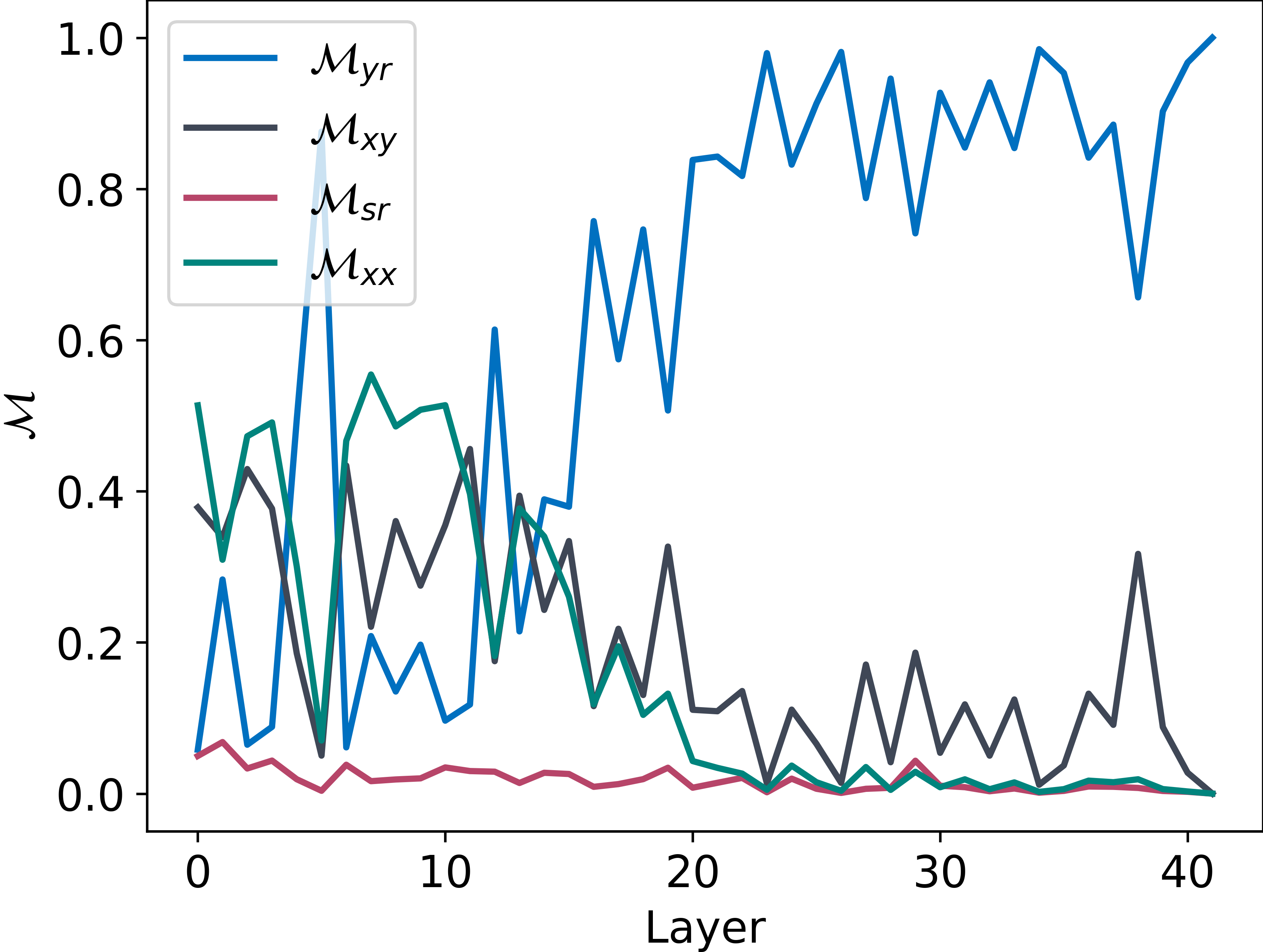}%
    \label{spp3}}
        \hfil
        \subfloat[]{\includegraphics[width=1.5in]{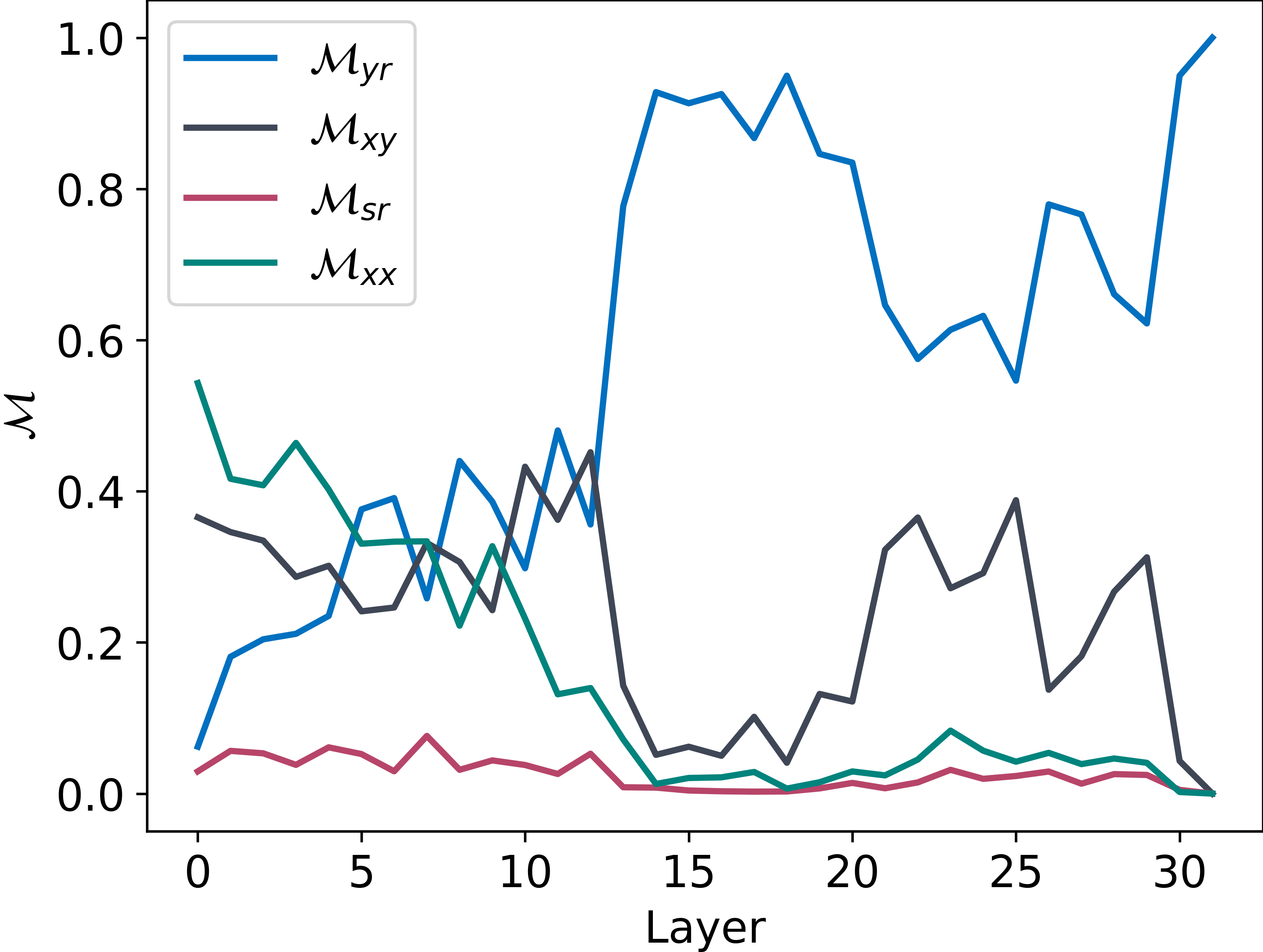}%
    \label{spp5}}
    \caption{Importance of information flow for each layer in the SFT LLMs involving process labels. (a) GPT-J. (b) Baichuan2-7b-chat. (c) Gemma2-9b-it. (d) Intern2.5-7b-chat.}
    \label{appendix_saliency_diff_part_for_process}
\end{figure*}

See Figure \ref{appendix_saliency_diff_part_for_process} for the $\mathcal{M}^{yr}$, $\mathcal{M}^{xr}$, $\mathcal{M}^{sr}$ and $\mathcal{M}^{xx}$ results of the remaining LLMs.
\label{information_results_one_other_llms}

\subsection{Results of $\mathcal{M}^{yr}$, $\mathcal{M}^{xy}$, $\mathcal{M}^{yr}$ and $\mathcal{M}^{yr}$}
\label{information_results_two_other_llms}

\begin{figure*}
    \centering
    \subfloat[]{\includegraphics[width=1.5in]{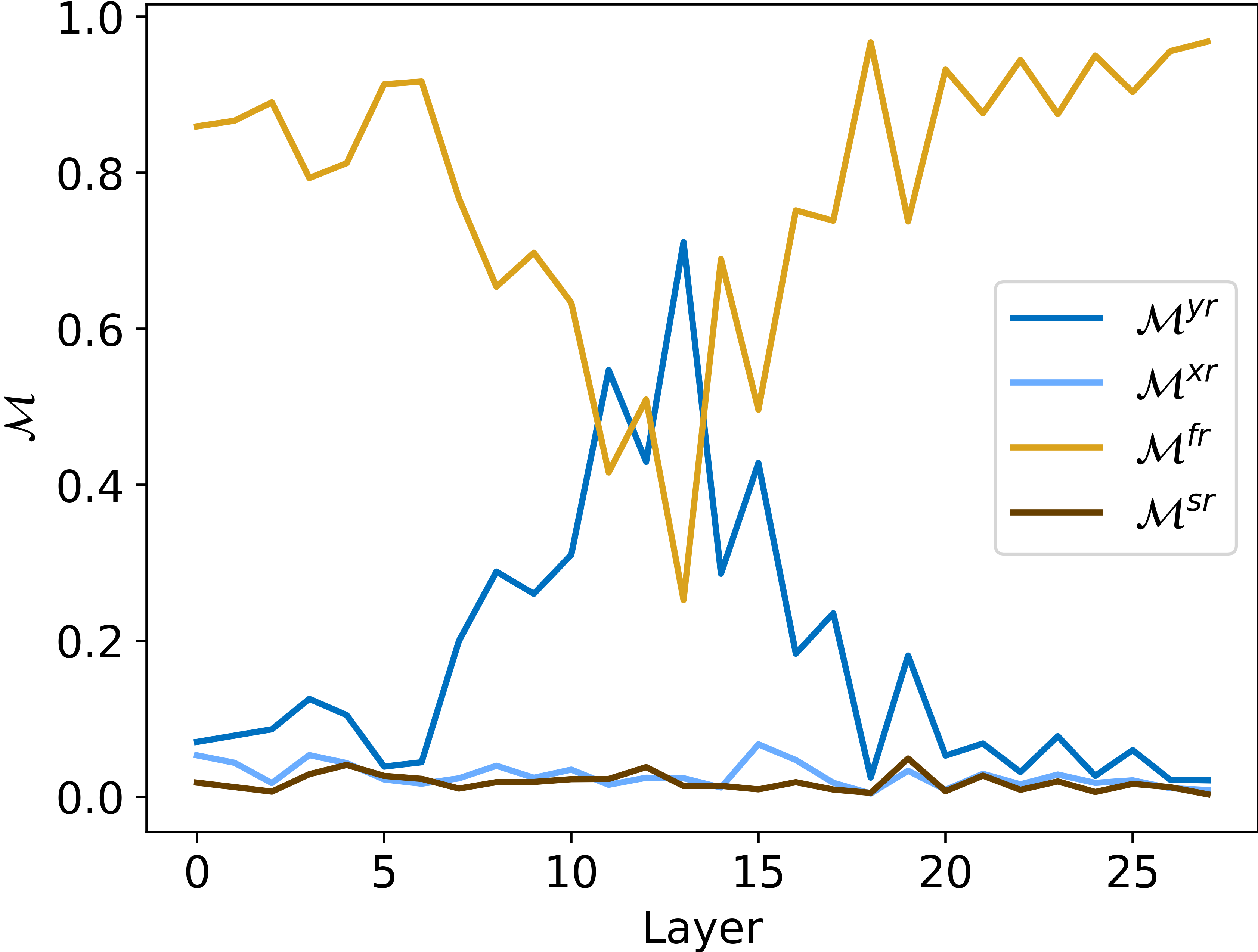}%
    \label{sp2}}
    \hfil
    \subfloat[]{\includegraphics[width=1.5in]{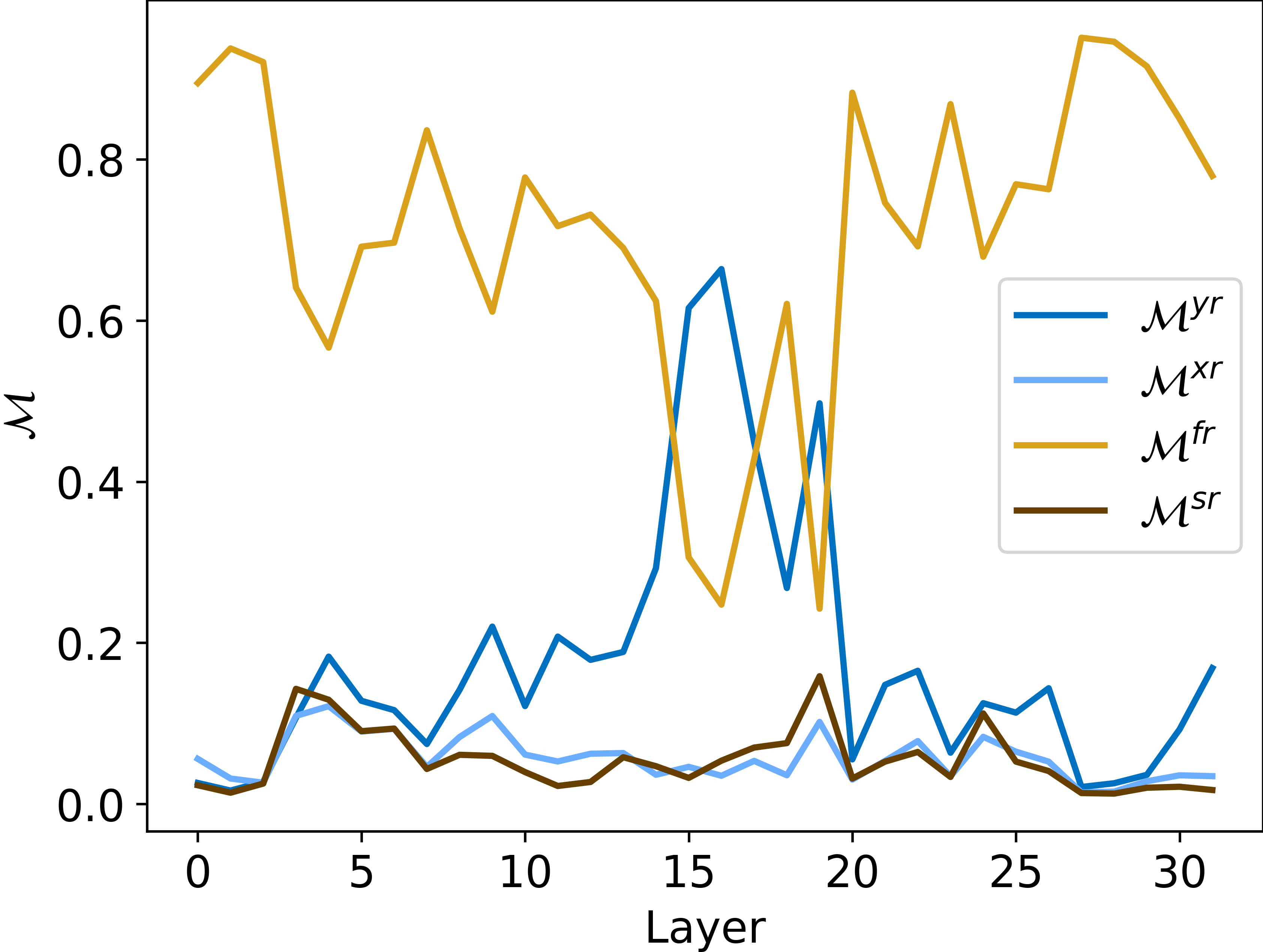}%
    \label{sp4}}
        \hfil
        \subfloat[]{\includegraphics[width=1.5in]{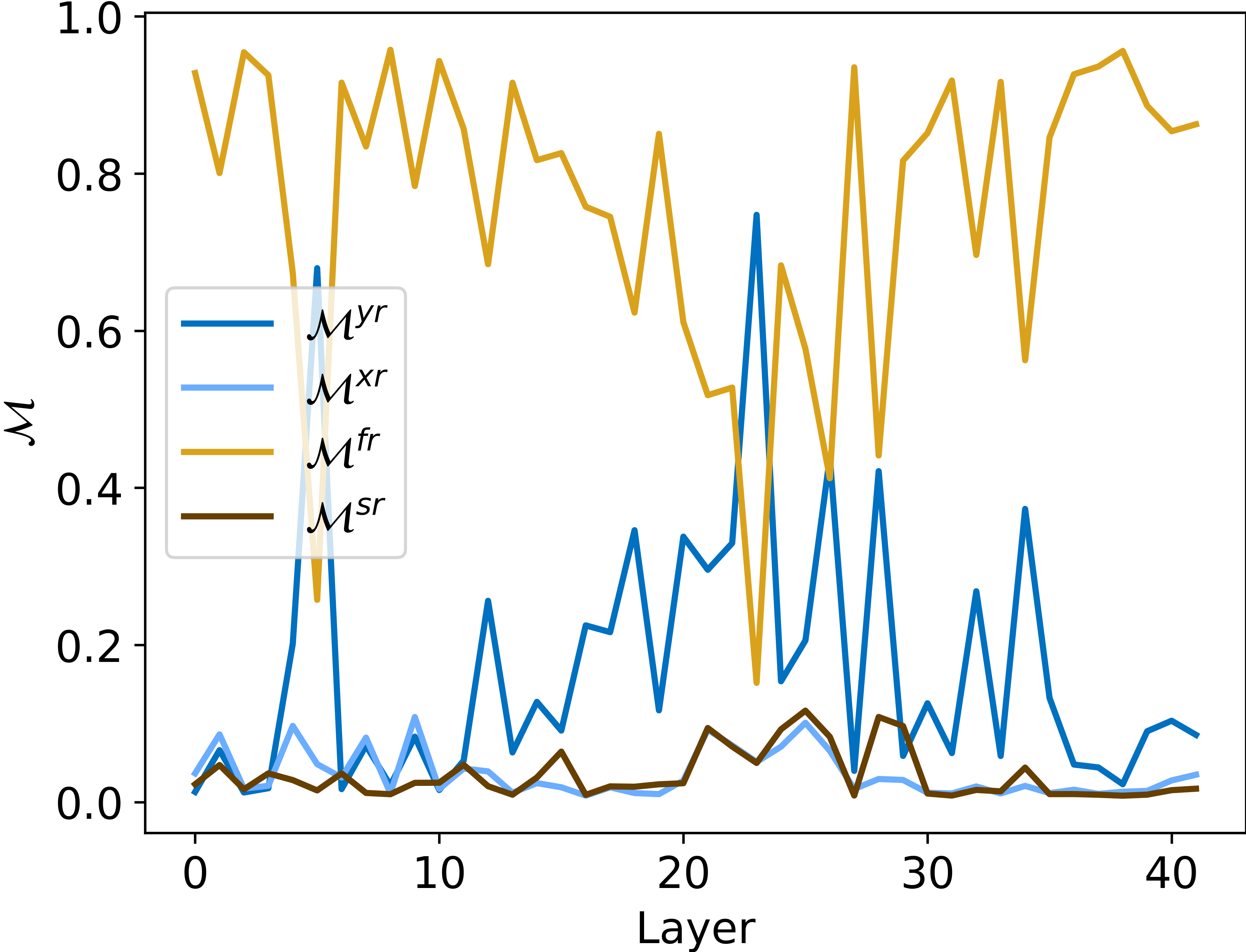}%
    \label{sp3}}
        \hfil
        \subfloat[]{\includegraphics[width=1.5in]{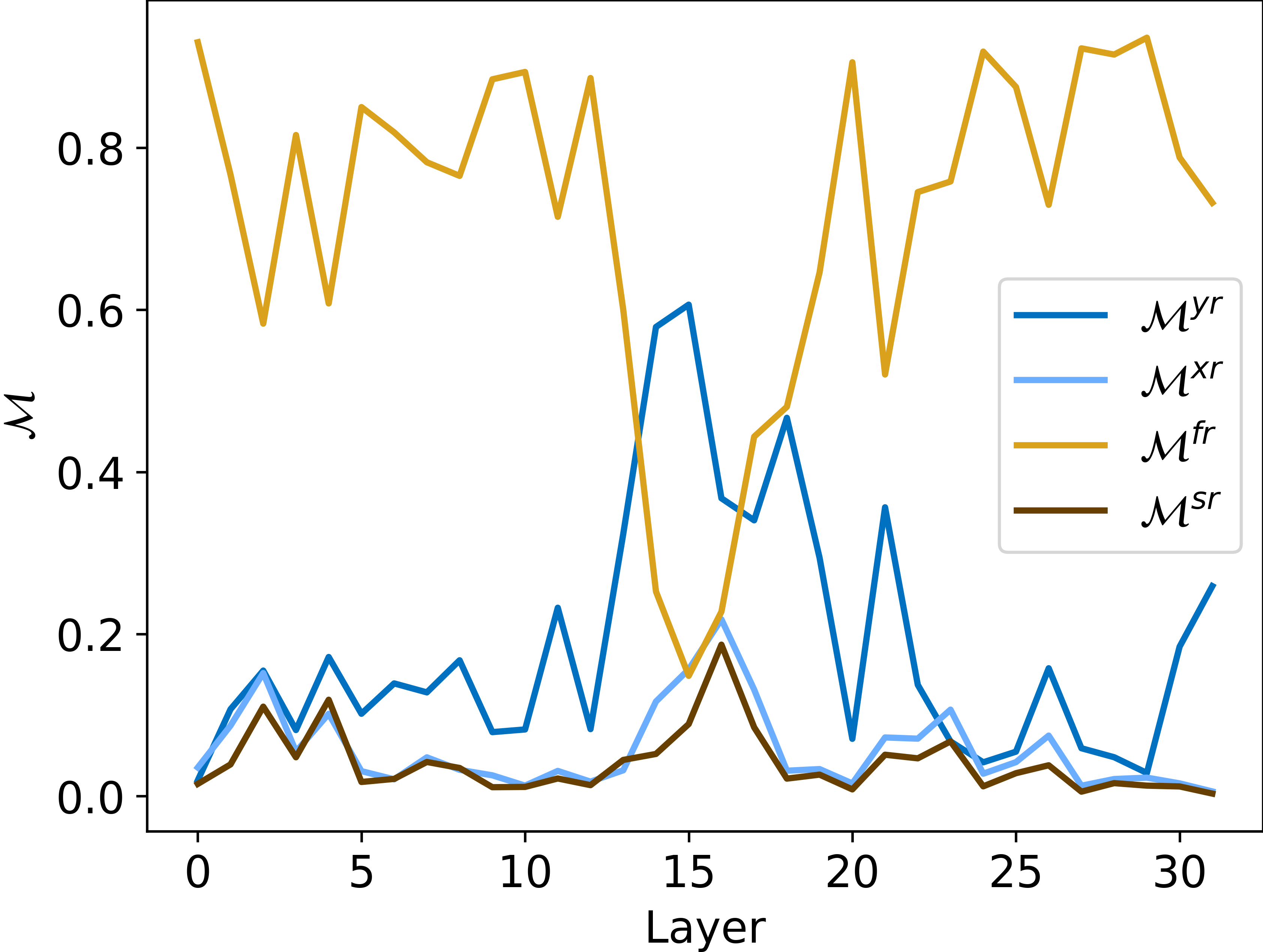}%
    \label{sp5}}
    \caption{Importance of information flow for each layer in the SFT LLMs for the prediction of final scene safety level. (a) GPT-J. (b) Baichuan2-7b-chat. (c) Gemma2-9b-it. (d) Intern2.5-7b-chat.}
    \label{appendix_saliency_diff_part}
\end{figure*}

See Figure \ref{appendix_saliency_diff_part} for the $\mathcal{M}^{yr}$, $\mathcal{M}^{xy}$, $\mathcal{M}^{yr}$ and $\mathcal{M}^{yr}$ results of the remaining LLMs.

\subsection{Results of $\mathcal{C}^1$ and $D^1$}
\label{correlation_results_one_other_llms}

\begin{figure*}
    \centering
    \subfloat[]{\includegraphics[width=1.5in]{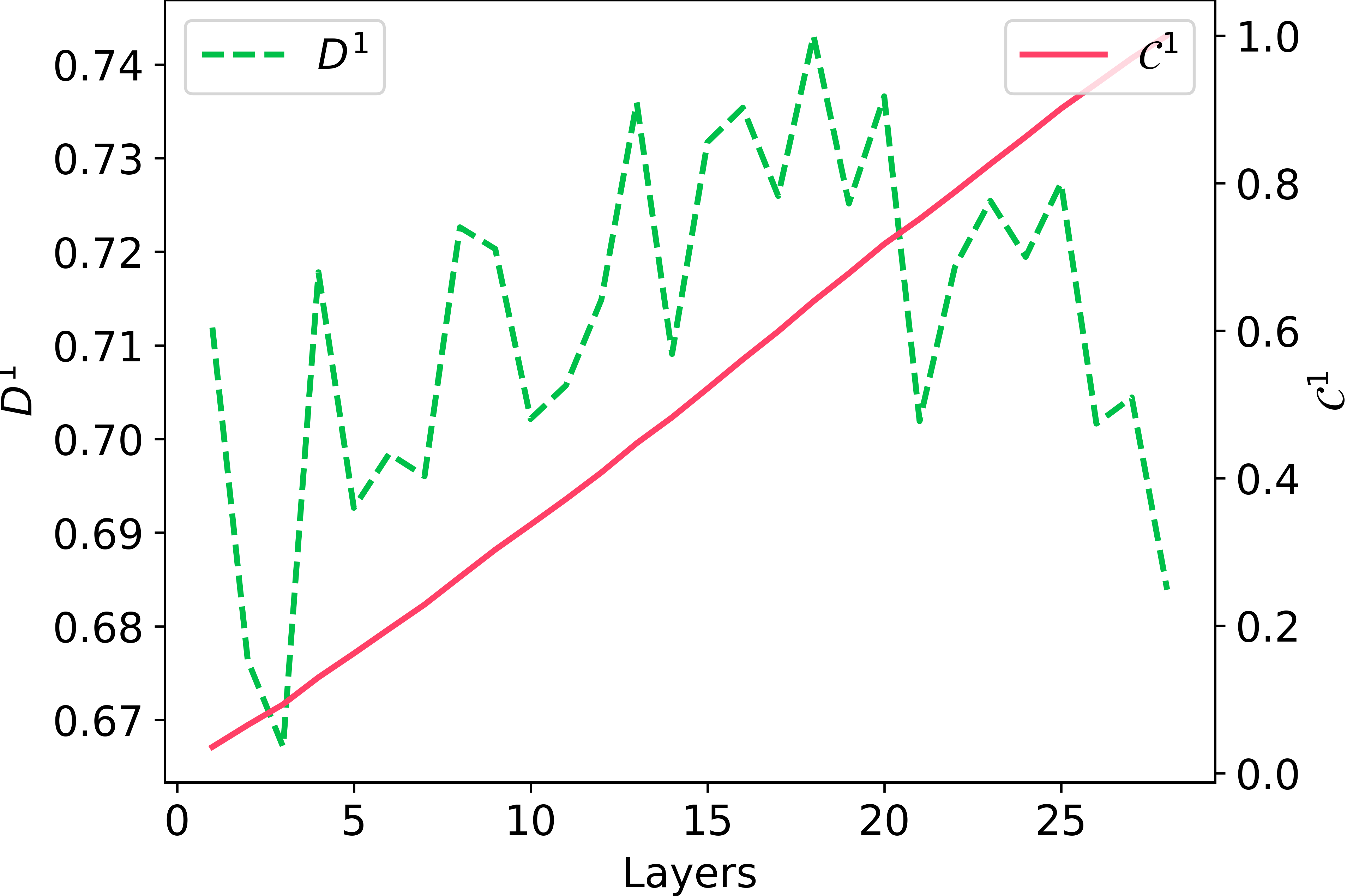}%
    \label{p2r2}}
    \hfil
    \subfloat[]{\includegraphics[width=1.5in]{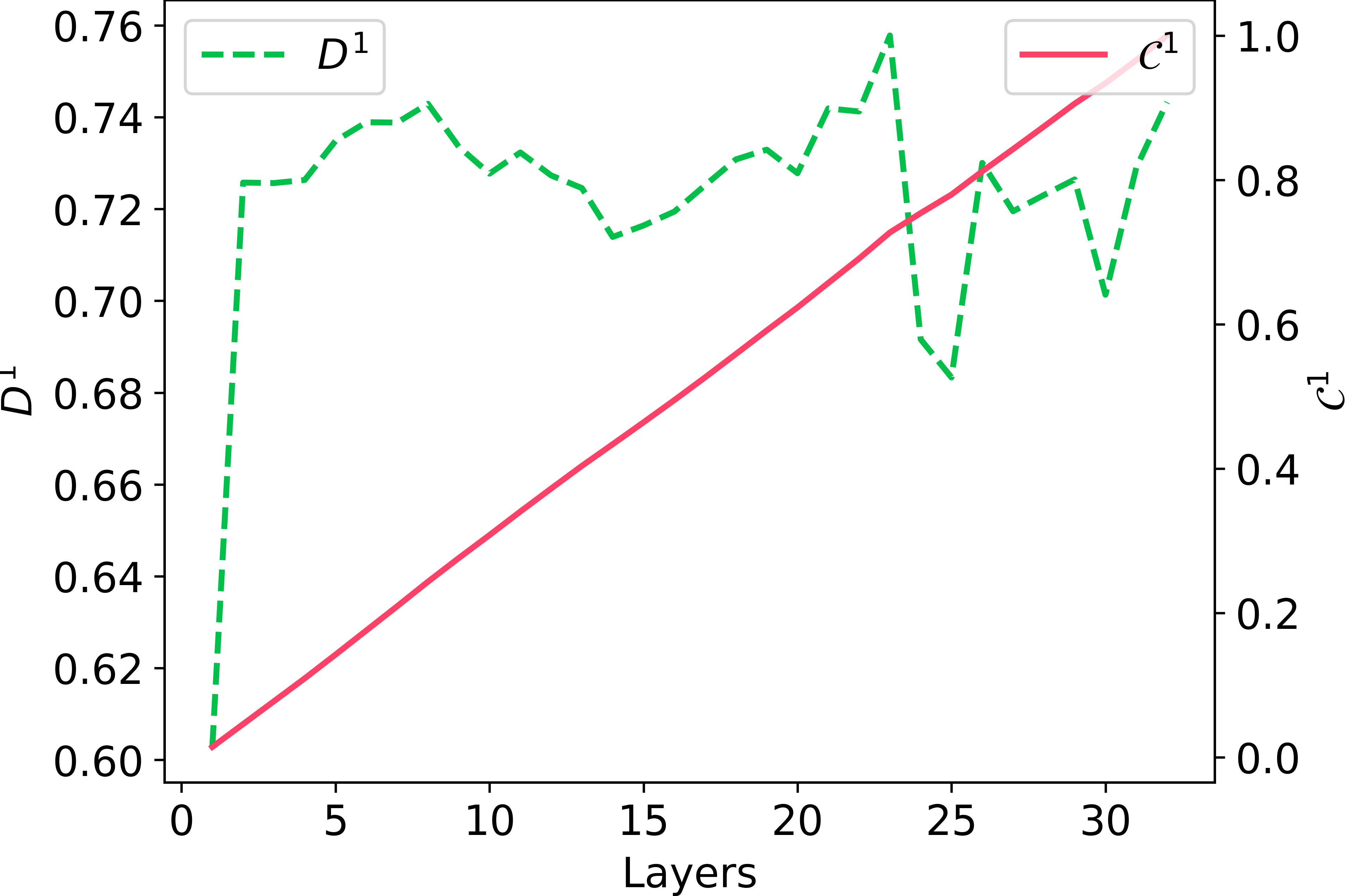}%
    \label{p2r4}}
        \hfil
        \subfloat[]{\includegraphics[width=1.5in]{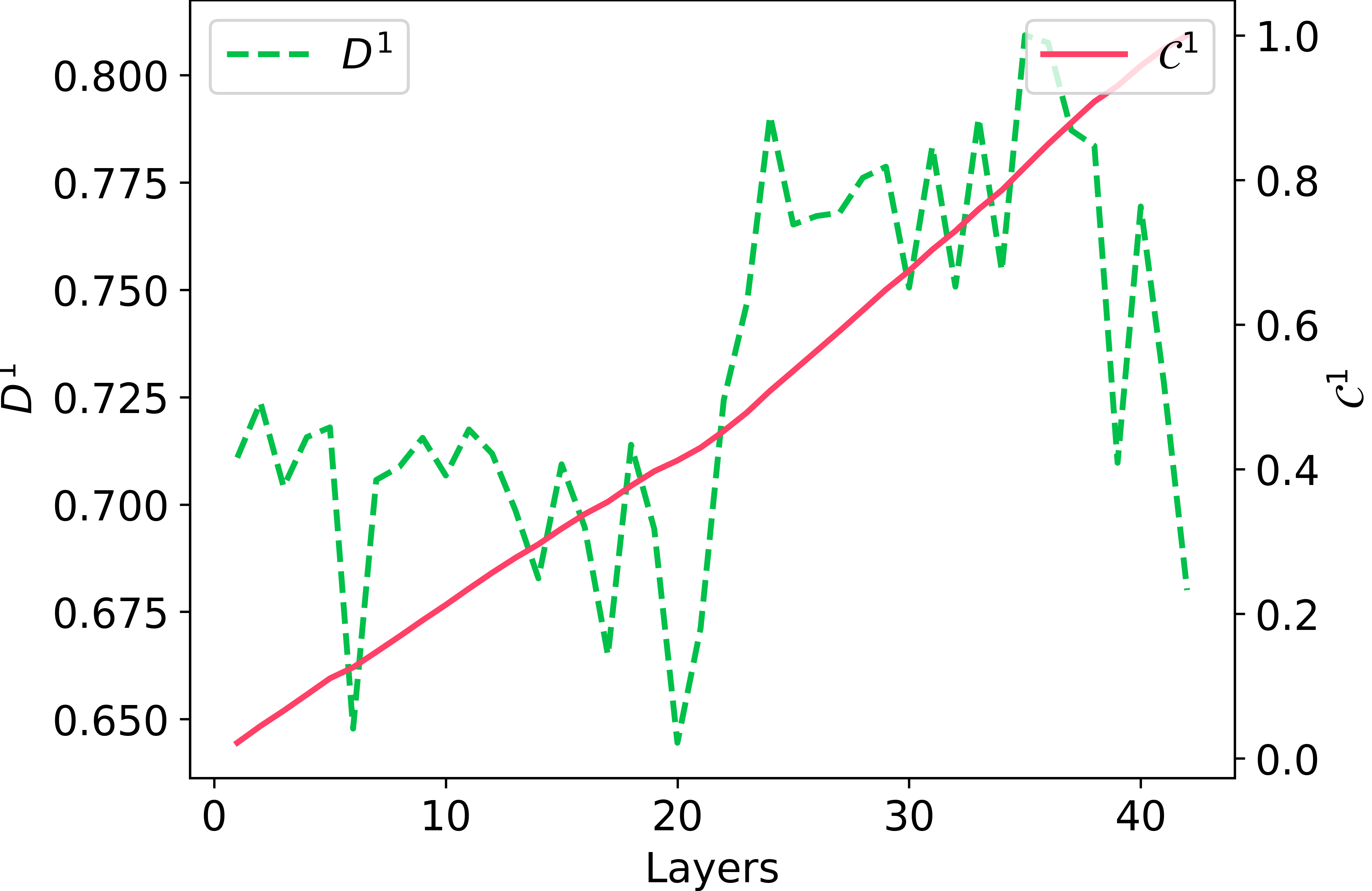}%
    \label{p2r3}}
        \hfil
        \subfloat[]{\includegraphics[width=1.5in]{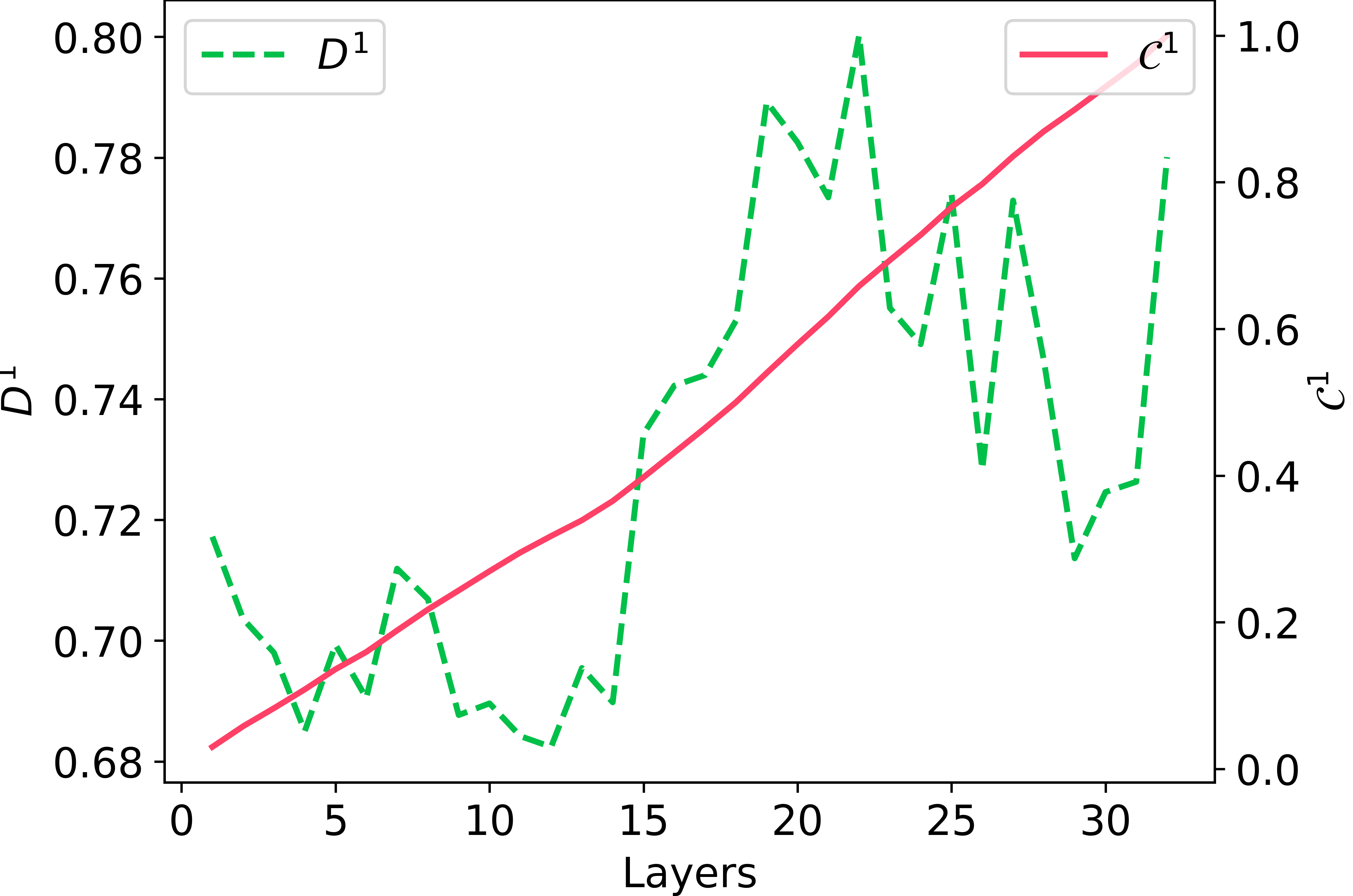}%
    \label{p2r5}}
    \caption{$\mathcal{C}^1_{l}$ and $D^1_l$ of different layers in LLMs.  (a) GPT-J. (b) Baichuan2-7b-chat. (c) Gemma2-9b-it. (d) Intern2.5-7b-chat.}
    \label{appendix_key_factor_final_p}
\end{figure*}

See Figure \ref{appendix_key_factor_final_p} for the $\mathcal{C}^1$ and $D^1$ results of the remaining LLMs.

\subsection{Results of $\mathcal{C}^2$ and $D^2$}
\label{correlation_results_two_other_llms}

\begin{figure*}
    \centering
    \subfloat[]{\includegraphics[width=1.5in]{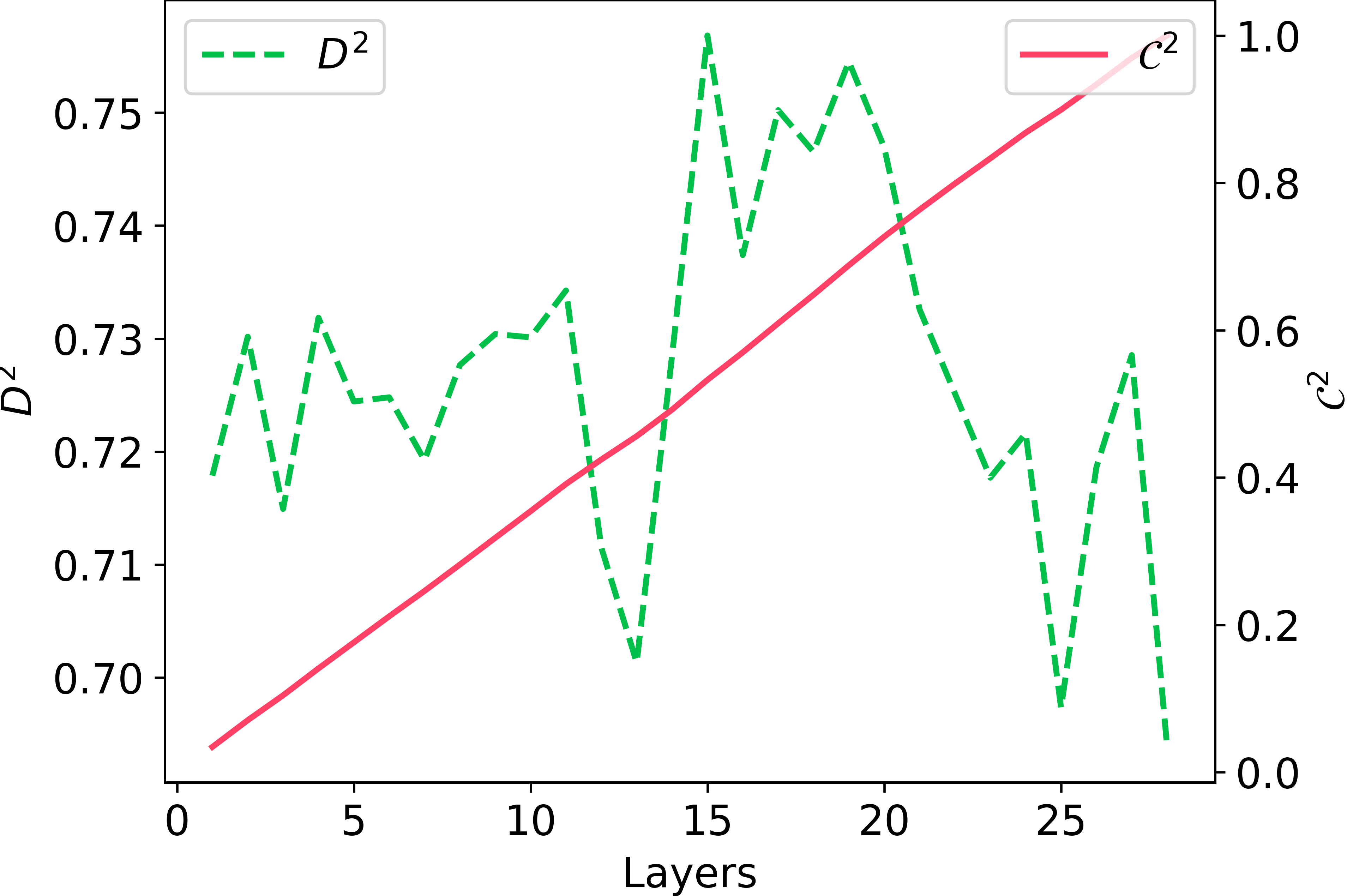}%
    \label{n2r2}}
    \hfil
    \subfloat[]{\includegraphics[width=1.5in]{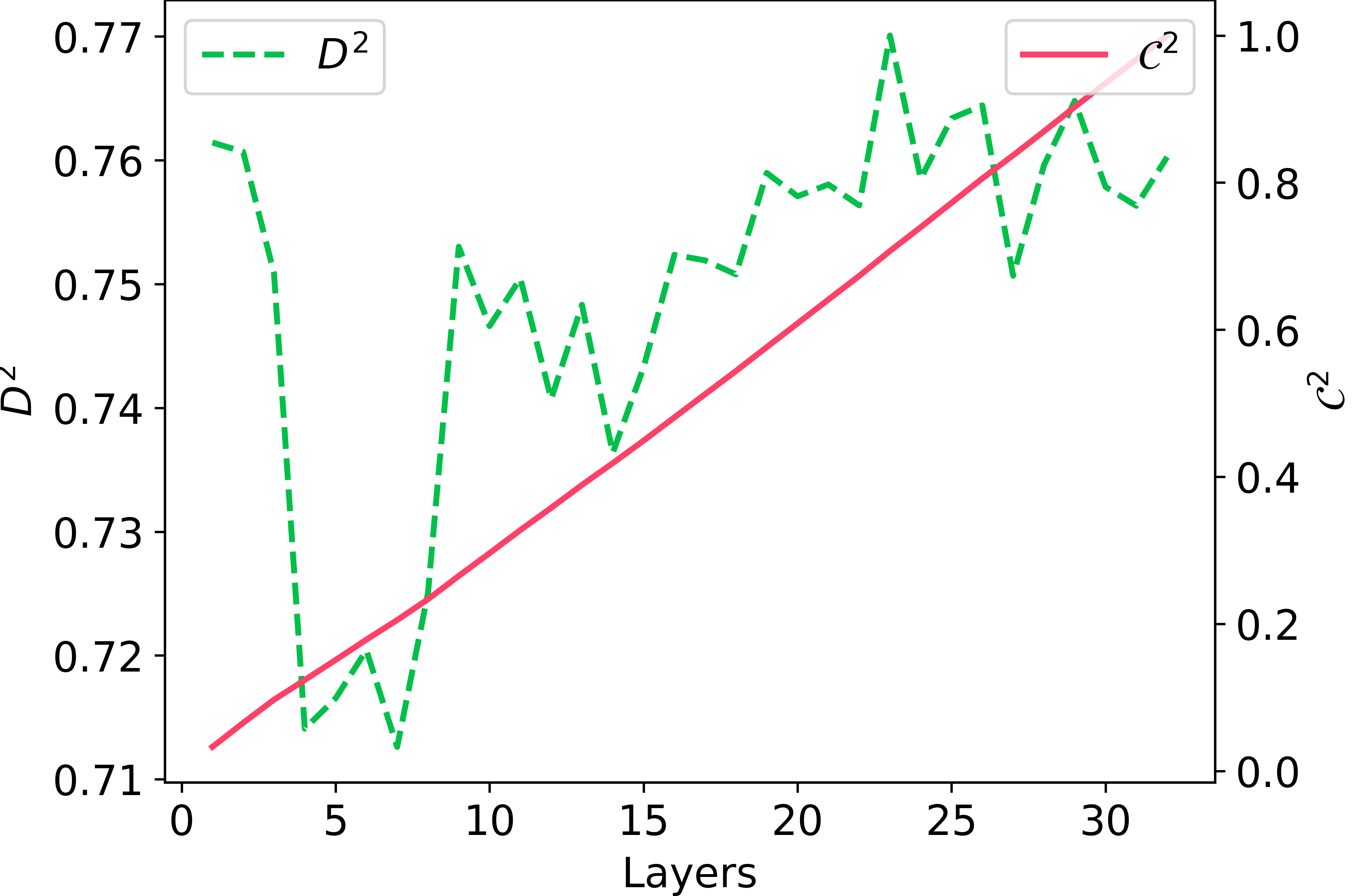}%
    \label{n2r4}}
        \hfil
        \subfloat[]{\includegraphics[width=1.5in]{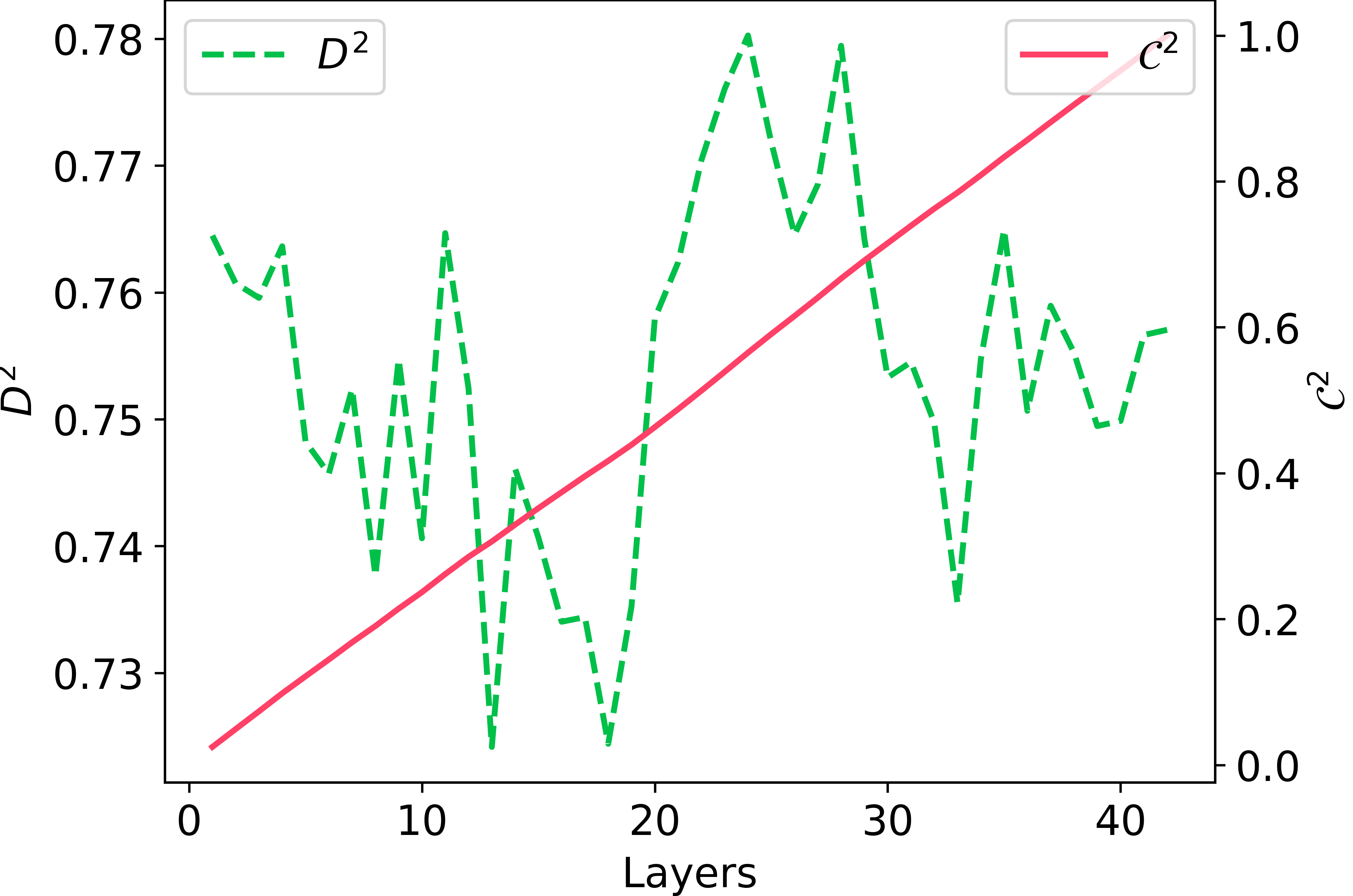}%
    \label{n2r3}}
        \hfil
        \subfloat[]{\includegraphics[width=1.5in]{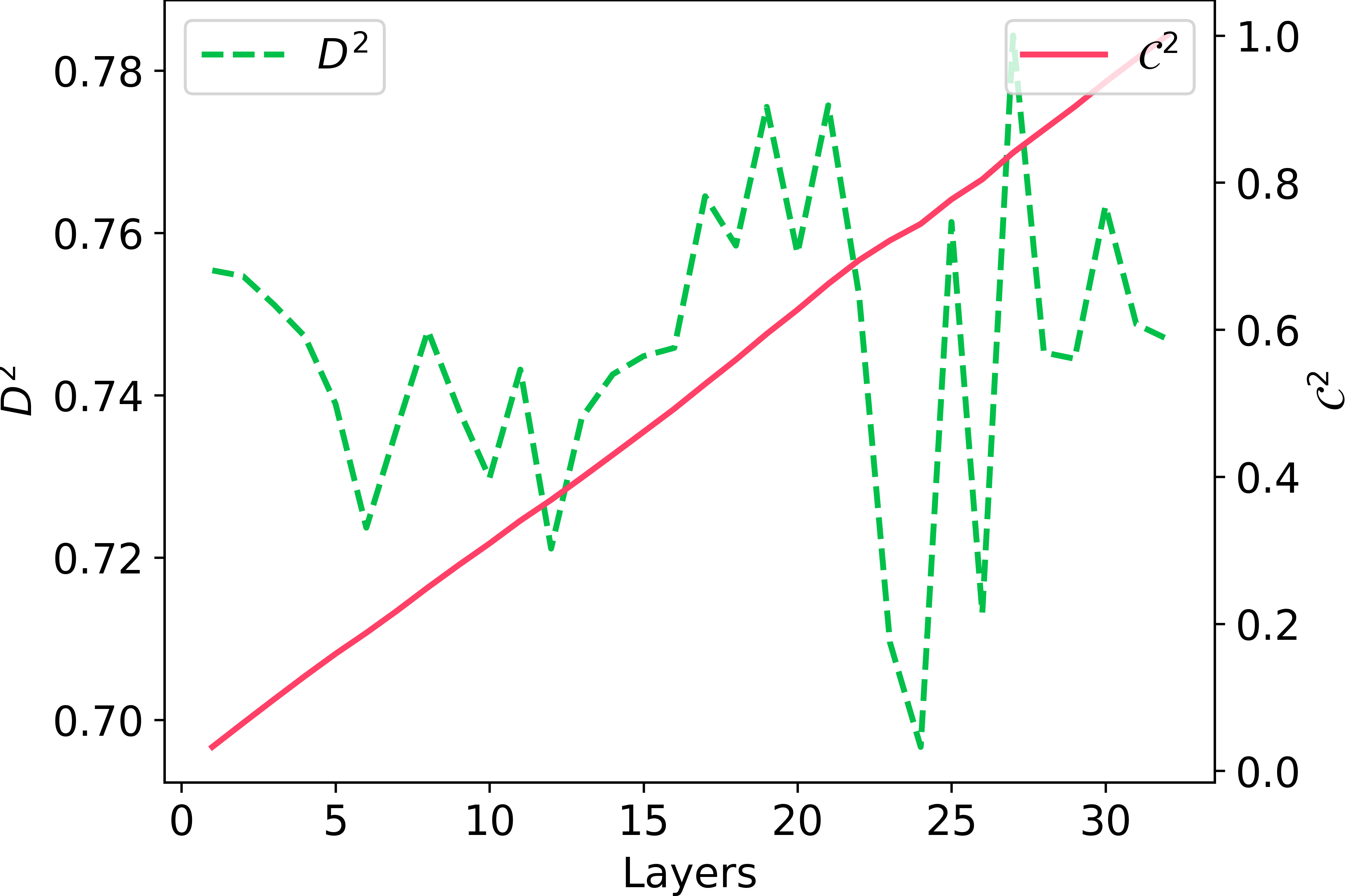}%
    \label{n2r5}}
    \caption{$\mathcal{C}^2_{l}$ and $D^2_l$ of different layers in LLMs. (a) GPT-J. (b) Baichuan2-7b-chat. (c) Gemma2-9b-it. (d) Intern2.5-7b-chat.}
    \label{appendix_key_factor_final_n}
\end{figure*}

See Figure \ref{appendix_key_factor_final_n} for the $\mathcal{C}^2$ and $D^2$ results of the remaining LLMs.

\section{Prompt Templates and Data Examples}
\subsection{System Prompt Templates for Different LLMs}
\label{system_templates}
For the system prompt templates of all LLMs, we use the templates that come with 
LLaMA-Factory. Specifically, Llama2-7b-chat uses \textit{llama2}, GPT-J uses \textit{default}, Baichuan2-7b-chat uses \textit{baichuan}, Gemma2-9b-it uses \textit{gemma}, and Intern2.5-7b-chat uses \textit{intern2}. For details, please refer to: \url{https://github.com/hiyouga/LLaMA-Factory/blob/main/src/llamafactory/data/template.py}

\subsection{Templates}
\label{templates_icl_cot}
The instruction templates for LoRA-Final, ICL, CoT and each task used in this study are displayed in Table \ref{template_method}.

\begin{table*}
\caption{Instruction templates for different methods.}
\centering
\renewcommand{\arraystretch}{1.2}
\begin{tabular}{>{\centering\arraybackslash}p{2cm} p{13cm}}
\toprule  
\textbf{Method/Task} & \textbf{Instruction Template} \\
\midrule
CoT & Please perform Scene Safety Level Recognition task. Given a text paragraph describing the scene in detail, assign a safety level label from [No risk, Low risk, Medium risk, High risk, Extremely high risk]. Let's think step by step. <scene description> \\
\midrule
ICL & Scene: <scene 1> Final risk level: No risk; \newline
Scene: <scene 2> Final risk level: Low risk; \newline
Scene: <scene 3> Final risk level: Medium risk; \newline
Scene: <scene 4> Final risk level: High risk; \newline
Scene: <scene 5> Final risk level: Extremely high risk; \newline
Scene: <test scene description> Final risk level: \\
\midrule
LoRA-Final & Please perform Scene Safety Level Recognition task. Given a text paragraph describing the scene in detail, assign a safety level label from [No risk, Low risk, Medium risk, High risk, Extremely high risk]. Return label only without any other text. <scene description> \\
\midrule
Perception task & Please execute the clause partitioning task. Given a scene description, divide it into multiple clauses, each clause describing an aspect of the paragraph. <scene description> \\
\midrule
Judging task & Please perform the impact judgment task. Given a scene description and a clause within the paragraph that describes an aspect of the paragraph. Determine the impact of this aspect on the overall safety of the scene and choose from [negative, positive, neutral]. Among them, negative represents weakening the safety of the scene, positive represents enhancing the safety of the scene, and neutral represents having no impact on safety or being difficult to judge. <scene description> <aspect> \\
\midrule
Cognitive task & Please perform the scene safety assessment task. Given a scene description paragraph and an analysis of the impact of various aspects in the paragraph on scene safety, directly provide the safety level of the scene and select from [No risk, Low risk, Medium risk, High risk, Extremely high risk]. <scene description> <aspect-impact pairs> \\
\bottomrule
\end{tabular}
\label{template_method}
\end{table*}

\subsection{Data Examples}
\label{data_examples}
For some examples of data, please refer to Table \ref{data_example}.

\section{Reward Model Details}
\label{reward_methods_detail}
 The training of reward models requires a large amount of data and building a process-supervisied dataset similar to PRM800K requires a significant amount of manpower. Thanks to the characteristics of the task labels in this study, the labeling of both the process labels and the final safety level labels contain the implicit meaning of manual sorting, e.g., the preference data pairs of $(x_i,\mathbb{P}_1)\succ (x_i,\mathbb{P}_2)\succ (x_i,\mathbb{P}_3)$ can be constructed for $x_i$ that is manually labeled as \textit{negative}, and $(C_i,\mathbb{D}_1)\succ (C_i,\mathbb{D}_2)\succ (C_i,\mathbb{D}_3)\succ (C_i,\mathbb{D}_4)\succ (C_i,\mathbb{D}_5)$ can be constructed for scene that is manually labeled as \textit{no risk}. Based on this feature, we use only a portion of the original training data to supervise the fine-tuning of the model(for standardizing the format of response), and the remaining portion is used to construct the human preference dataset to train the reward model. The LoRA technique is also used to train the reward model, and its base model is the same as the base model of the SFT model.

\clearpage
\newpage
\begin{sidewaystable*}
\caption{Examples of data in the SSUPL dataset and used by the LoRA-Final approach.}
\label{data_example}
\centering
\begin{tabular*}{\textheight}{@{\extracolsep\fill}m{3cm}ccc}
\hline
\toprule%
\centering \multirow{2}{*}{\textbf{Scene Image}} & \multirow{2}{*}{\textbf{Scene Description}}&\textbf{Response} & \textbf{Response} \\
& & \textbf{(LoRA-Final)} & \textbf{(SSUPL)}\\
\midrule
\centering \includegraphics[width=2cm,height=2cm]{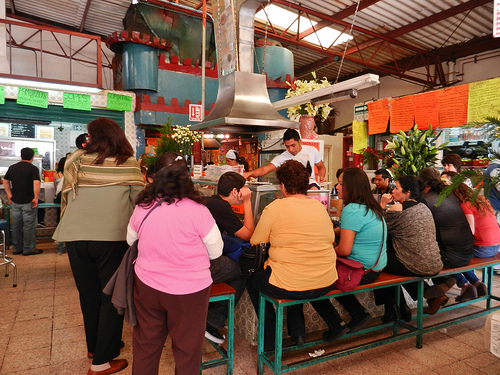}& 

 \makecell{\parbox{8cm}{\raggedright A group of people sitting in a cafeteria style restaurant. There's plenty of light here. A woman in a bright pink shirt is standing by the table with a purse on her shoulder. The legs of the chairs at the table are bright blue.}}
 & No risk & \makecell{\parbox{10cm}{\raggedright Aspect: A group of people sitting in a cafeteria style restaurant. Impact: Neutral; Aspect: There's plenty of light here. Impact: Positive; Aspect: A woman in a bright pink shirt is standing by the table with a purse on her shoulder. Impact: Neutral; Aspect: The legs of the chairs at the table are bright blue. Impact: Neutral; Final Risk Level: No risk}}\\
\midrule
\centering \includegraphics[width=2cm,height=2cm]{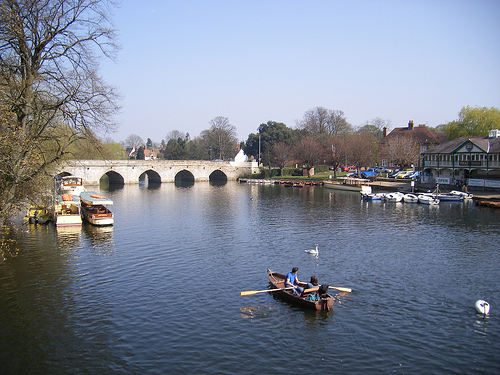}& 
\makecell{\parbox{8cm}{\raggedright People sitting inside a row boat on the water. The water is very calm with very few small waves in it. There are several other boats in the water as well. There is a bridge behind the boats above the water. There is a building next to the water.}}
 & Low risk & \makecell{\parbox{10cm}{\raggedright Aspect: People sitting inside a row boat on the water. Impact: Negative; Aspect: The water is very calm with very few small waves in it. Impact: Positive; Aspect: There are several other boats in the water as well. Impact: Negative; Aspect: There is a bridge behind the boats above the water. Impact: Neutral; Aspect: There is a building next to the water. Impact: Neutral; Final Risk Level: Low risk} }\\
\midrule
\centering \includegraphics[width=2cm,height=2cm]{./pics/medium_risk.jpg}& 
\makecell{\parbox{8cm}{\raggedright A man rides a skateboard with red wheels on the street in front of a public transportation bus. He is wearing a helmet, a striped shirt, short pants knee pads, and shoes and socks. Behind him is a thick row of trees. The bus has blue colored sides with tinted windows and a digital display near the closed doors.}}
 & Medium risk & \makecell{\parbox{10cm}{\raggedright Aspect: A man rides a skateboard with red wheels on the street in front of a public transportation bus. Impact: Negative; Aspect: He is wearing a helmet, a striped shirt, short pants knee pads, and shoes and socks. Impact: Positive; Aspect: Behind him is a thick row of trees. Impact: Neutral; Aspect: The bus has blue colored sides with tinted windows and a digital display near the closed doors. Impact: Neutral; Final Risk Level: Medium risk}}\\
\midrule
\centering \includegraphics[width=2cm,height=2cm]{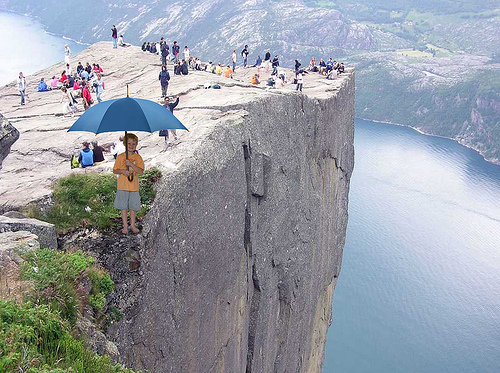}& 
\makecell{\parbox{8cm}{\raggedright The cliff in the scene is really high. In the distance on the other side of the water there is more land that is lower and has sparse vegetation. The water is blue green and has a small ripple from the wind. On the cliff, there are groups of people some of which are standing and others that are sitting. There is one little boy on the edge of the high cliff that is holding a large blue umbrella. He is wearing an orange shirt, gray short, and no shoes.}}
 & High risk & \makecell{\parbox{10cm}{\raggedright Aspect: The cliff in the scene is really high. Impact: Negative; Aspect: In the distance on the other side of the water there is more land that is lower and has sparse vegetation. Impact: Neutral; Aspect: The water is blue green and has a small ripple from the wind. Impact: Neutral; Aspect: On the cliff, there are groups of people some of which are standing and others that are sitting. Impact: Negative; Aspect: There is one little boy on the edge of the high cliff that is holding a large blue umbrella. Impact: Negative; Aspect: He is wearing an orange shirt, gray short, and no shoes. Impact: Neutral; Final Risk Level: High risk}}\\
 \midrule
\centering \includegraphics[width=2cm,height=2cm]{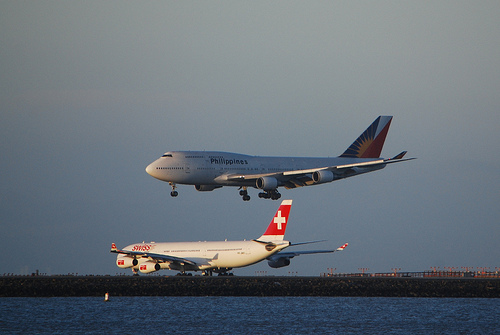}&
\makecell{\parbox{8cm}{\raggedright A large passenger jet prepares to land at an airport on a sunny afternoon. The airport is surrounded by water and a large white jet taxis along a runway. The distance between the two planes is very close and is gradually getting closer.}}
 &\makecell{\parbox{1.5cm}{\centering Extremely high risk}}& \makecell{\parbox{10cm}{\raggedright Aspect: A large passenger jet prepares to land at an airport. Impact: Negative; Aspect: on a sunny afternoon. Impact: Positive; Aspect: The airport is surrounded by water and a large white jet taxis along a runway. Impact: Neutral; Aspect: The distance between the two planes is very close and is gradually getting closer. Impact: Negative; Final Risk Level: Extremely high risk}}\\
\bottomrule
\end{tabular*}
\end{sidewaystable*}


\clearpage
\newpage
\bibliographystyle{cas-model2-names}

\bibliography{cas-refs}



\end{document}